\documentclass{article} 
\usepackage{iclr2026_conference,times}

\usepackage{amsmath,amsfonts,bm}

\def\eqref#1{equation~\ref{#1}}

\def\1{\bm{1}}

\DeclareMathAlphabet{\mathsfit}{\encodingdefault}{\sfdefault}{m}{sl}
\SetMathAlphabet{\mathsfit}{bold}{\encodingdefault}{\sfdefault}{bx}{n}

\usepackage{hyperref}
\usepackage{url}

\usepackage{amsmath}
\usepackage{graphicx}
\usepackage{multirow}
\usepackage[table]{xcolor}
\usepackage{subcaption}
\usepackage{wrapfig}
\usepackage{makecell}
\usepackage{booktabs}       
\usepackage{amsfonts}       
\usepackage{nicefrac}       
\usepackage{microtype}      
\usepackage{xcolor}         

\title{RestoreVAR: Visual Autoregressive Generation for All-in-One Image Restoration}

\author{Sudarshan Rajagopalan, Kartik Narayan \& Vishal M. Patel\\
Johns Hopkins University\\
Baltimore, MD 21218, USA \\
\texttt{\{sambasa2,knaraya4,vpatel36\}@jhu.edu} \\
}

\iclrfinalcopy 
\begin{document}

\maketitle

\begin{abstract}
The use of latent diffusion models (LDMs) such as Stable Diffusion has significantly improved the perceptual quality of All-in-One image Restoration (AiOR) methods, while also enhancing their generalization capabilities. However, these LDM-based frameworks suffer from slow inference due to their iterative denoising process, rendering them impractical for time-sensitive applications. Visual autoregressive modeling (VAR), a recently introduced approach for image generation, performs scale-space autoregression and achieves comparable performance to that of state-of-the-art diffusion transformers with drastically reduced computational costs. Moreover, our analysis reveals that coarse scales in VAR primarily capture degradations while finer scales encode scene detail, simplifying the restoration process. Motivated by this, we propose RestoreVAR, a novel VAR-based generative approach for AiOR that significantly outperforms LDM-based models in restoration performance while achieving over $\mathbf{10\times}$ faster inference. To optimally exploit the advantages of VAR for AiOR, we propose architectural modifications and improvements, including intricately designed cross-attention mechanisms and a latent-space refinement module, tailored for the AiOR task. Extensive experiments show that RestoreVAR achieves state-of-the-art performance among generative AiOR methods, while also exhibiting strong generalization capabilities. Project Page: \url{https://sudraj2002.github.io/restorevarpage/}
\end{abstract}

\section{Introduction}
\label{sec: intro}
Image restoration is a complex inverse problem that aims to recover clean images from degradations, such as haze, rain, snow, blur, and low-light conditions. Recently, the paradigm of All-in-One image Restoration (AiOR) has emerged, where a single network is trained to handle multiple degradation types. Existing AiOR methods can be broadly categorized into non-generative and generative approaches. Non-generative models such as AirNet~\citep{airnet}, PromptIR~\citep{promptir}, InstructIR~\citep{instructir}, AWRaCLe~\citep{awracle}, and AdaIR~\citep{adair}, deterministically map degraded images to their clean counterparts. While these methods offer fast inference and reliable pixel-level restoration performance, they often fail to generalize to diverse degradations encountered in real-world scenarios. To overcome this challenge, recent works have adopted generative models that aim to capture the distribution of clean images and produce more perceptually realistic outputs. Early works~\citep{gan1rain,gan3blur} based on GANs~\citep{gan} attempted this through adversarial learning, but suffered from mode collapse and unstable training. To improve fidelity and training stability, DiffUIR~\citep{diffuir} and DA-CLIP~\citep{daclip} employed pixel-space diffusion models~\citep{ddpm}. However, their high computational cost makes large-scale pretraining infeasible, limiting their ability to learn strong generative priors. In contrast, recent methods such as Diff-Plugin\citep{diffplugin}, AutoDIR\citep{autodir}, and PixWizard~\citep{pixwizard} leverage latent diffusion models (LDMs), such as Stable Diffusion~\citep{stablediff}. By operating in a latent space, LDMs significantly reduce computational costs, enabling large-scale pretraining which equips them with strong generative priors of natural images. These priors allow LDM-based AiOR methods to deliver perceptually realistic restoration and improved generalization to real-world degradations.

Despite their advantages, LDM-based AiOR methods have some shortcomings. (1) LDMs require multiple denoising steps during inference, resulting in significantly longer runtimes compared to non-generative models. Their slow inference speeds pose challenges for applications that demand real-time processing, such as video surveillance or autonomous navigation. (2) LDMs rely on variational autoencoders (VAEs)~\citep{vae} which are primarily trained for generative diversity, rather than accurate pixel-level reconstruction. Consequently, the restored images obtained from LDM-based AiOR methods exhibit loss of fine structural details, hindering their performance.

Autoregressive models have driven rapid advances in natural language processing through large language models (LLMs) such as GPT-3~\citep{radford2019language} and LLaMA~\citep{touvron2023llama}. These models generate outputs by predicting the next token, conditioned on previously generated tokens. Recently, Visual AutoRegressive (VAR) Modeling~\citep{var} introduced scale-space autoregression for image generation, performing next-scale prediction in the latent space of a multi-scale vector-quantized VAE (VQVAE). VAR achieves performance comparable to state-of-the-art diffusion models such as DiT-XL/2~\citep{dit}, while operating $45\times$ faster. Despite its success in generative tasks, the application of VAR to low-level vision tasks such as image restoration remains largely unexplored. To the best of our knowledge, only two prior works: VarSR~\citep{varsr} and Varformer~\citep{varformer}, have used VAR for image restoration. VarSR focused exclusively on the super-resolution task, while Varformer utilized intermediate VAR features to guide a separate non-generative network for AiOR. In contrast, our approach is generative and fully exploits the strong priors of the pretrained VAR model by training it directly for the AiOR task. Our analysis in Sec.~\ref{subsec: var_motivation} also reveals that the scale-space decomposition of VAR captures degradations predominantly in coarse scales and scene-level details in fine scales, making it well-suited for AiOR.

\begin{figure*}[t]
  \centering
  \begin{subfigure}{0.115\textwidth}
      \includegraphics[height=3.156\linewidth, width=\linewidth]{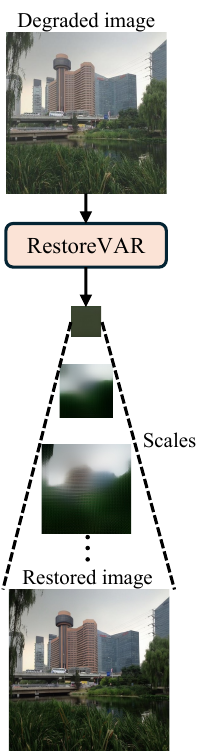}
      \caption{Overview}
  \end{subfigure}\hfill
  \begin{subfigure}{0.363\textwidth}
      \includegraphics[height=\linewidth, width=\linewidth]{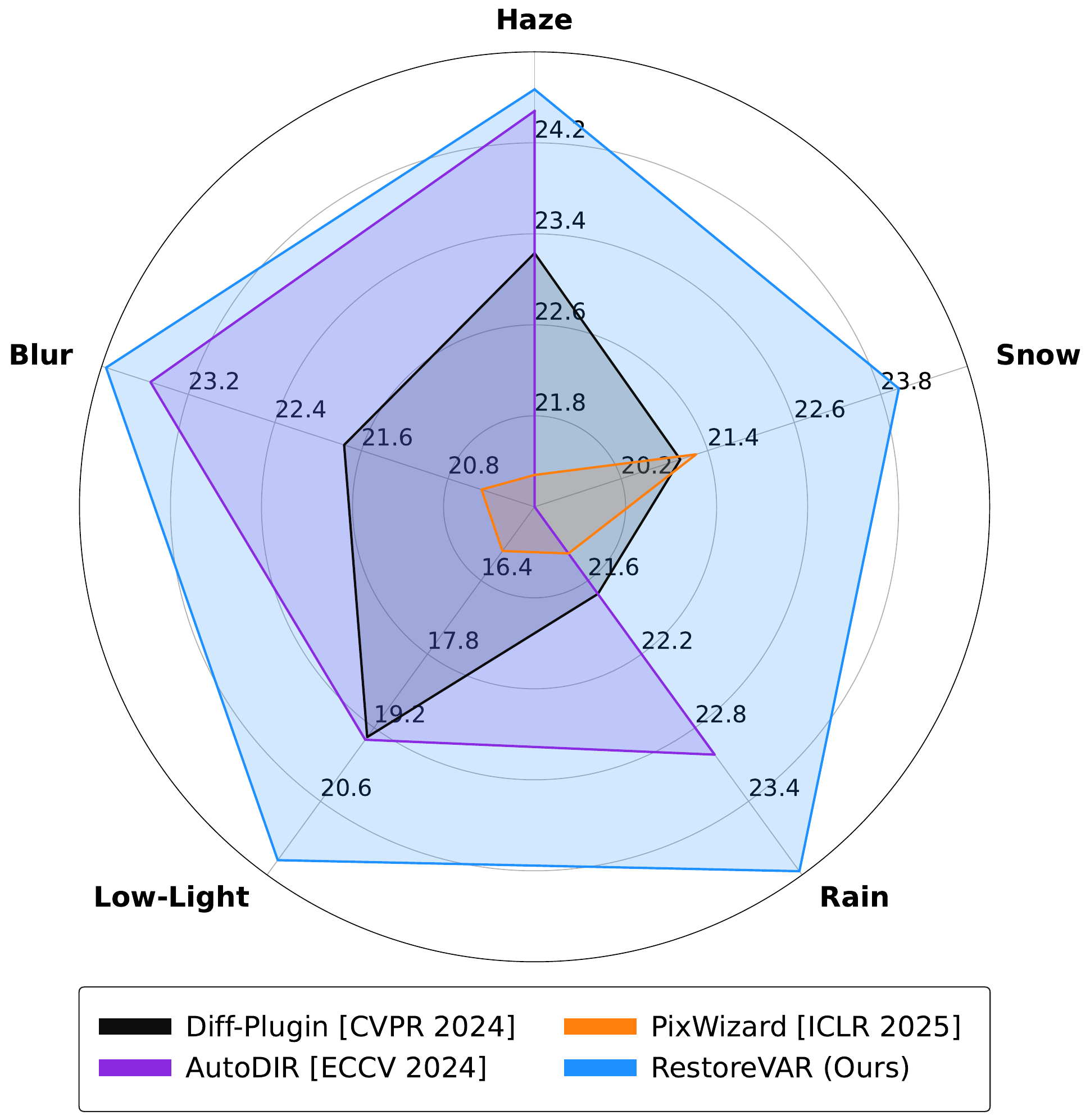}
      \caption{PSNR (dB)$\uparrow$}
  \end{subfigure}\hfill
  \begin{subfigure}{0.521\textwidth}
      \includegraphics[height=0.696\linewidth, width=\linewidth]{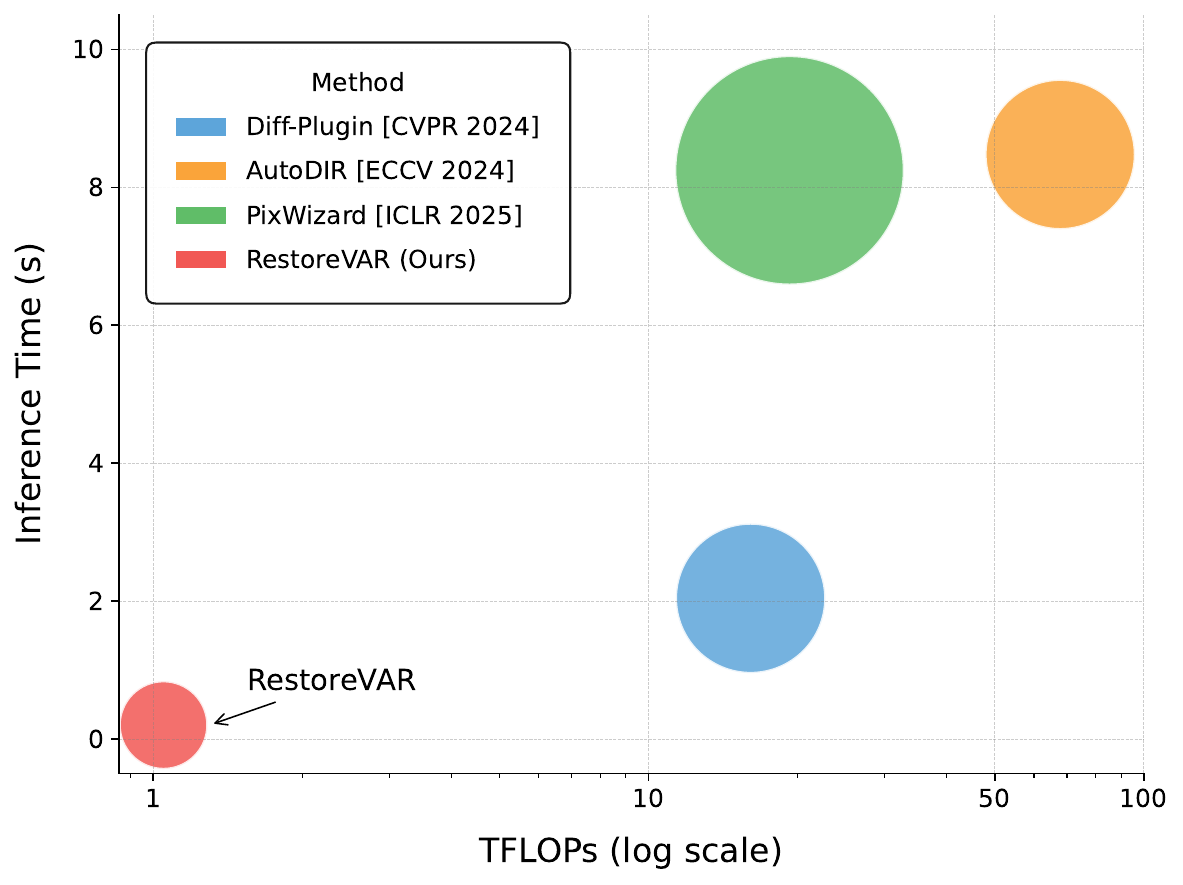}
      \caption{Model complexity (bubble size $\propto$ parameter count)}
  \end{subfigure}
  \vspace{-12pt}
  \caption{RestoreVAR, our proposed VAR-based~\citep{var} scale-space generative AiOR model (a), significantly outperforms LDM-based methods as shown in (b). RestoreVAR also offers drastic reductions in computational complexity as shown in (c).}
  \vspace{-15pt}
  \label{fig: intro}
\end{figure*}

To this end, we introduce RestoreVAR, a novel generative approach for AiOR that addresses some of the key limitations of LDM-based approaches. Firstly, RestoreVAR adopts the autoregressive structure of VAR, achieving state-of-the-art generative AiOR performance with over $\mathbf{10\times}$ faster inference than LDM-based methods (see Fig.~\ref{fig: intro}). Secondly, RestoreVAR employs cross-attention mechanisms conditioned on the degraded image latents, enabling the model to maintain spatial consistency and minimize hallucinations. Thirdly, to mitigate the loss of fine details by the vector quantization and VAE decoding processes, we propose a lightweight (only $\sim\mathbf{3}$\% overhead) non-generative latent refinement transformer which predicts de-quantized latents from the outputs of VAR. Additionally, we fine-tune the VAE decoder to operate on these continuous latents, further enhancing reconstruction quality. Finally, through extensive experiments, we demonstrate that RestoreVAR achieves state-of-the-art performance among generative restoration models, while also exhibiting strong generalization to real-world degradations. To summarize, our key contributions are:

\begin{enumerate}
    \item We propose RestoreVAR, the first VAR-based generative AiOR framework that achieves superior performance and a $\mathbf{10\times}$ faster inference than LDM-based methods. 
    \item To achieve semantically coherent restoration, we introduce degraded image conditioning through cross-attention at each block of the VAR transformer. 
    \item To mitigate the loss of fine details in the vector quantization and VAE decoding processes, we introduce a non-generative latent refiner transformer which converts discretized latents into continuous ones, and fine-tune the VAE decoder to operate on continuous latents.
    \item Extensive experiments show that RestoreVAR attains state-of-the-art performance among generative AiOR approaches, with perceptually preferable results and strong generalization.
\end{enumerate}

\section{Related Works}
\label{sec: related}

\subsection{Image restoration}
Early restoration models primarily addressed specific degradations ~\citep{early1,haze1,rain1,rainnew2,snow1,gopro}. Other methods such as Restormer~\citep{restormer}, MPRNet~\citep{mprnet} and SwinIR~\citep{swinir} introduced architectures for any single restoration task. However, they are restricted to handle one degradation at a time, making them ineffective for multiple degradations.
All-in-One image Restoration (AiOR) methods aim to tackle multiple corruptions with a single model. Early approaches include non-generative models such as All-in-one~\citep{nas} and Transweather~\citep{transw}. PromptIR~\citep{promptir} used learnable prompts while AWRaCLe~\citep{awracle} utilized visual in-context learning to extract degradation characteristics. Other approaches such as InstructIR~\citep{instructir} adopted textual guidance, and DCPT~\citep{dcpt} proposed a novel pre-training strategy for AiOR. DFPIR~\citep{dfpir} proposed a feature perturbation strategy for AiOR. Recent AiOR methods have adopted diffusion models. Pixel-space diffusion models (PSDMs) such as DA-CLIP~\citep{daclip} and DiffUIR~\citep{diffuir} demonstrated improved AiOR performance but lacked robust generative priors. Recent methods have utilized the strong priors of LDMs for AiOR. Diff-Plugin~\citep{diffplugin} adopts task plugins to guide an LDM for AiOR. AutoDIR~\citep{autodir} automatically detects and restores degradations using an LDM. PixWizard~\citep{pixwizard} is a multi-task Lumina-next~\citep{lumina} based model capable of performing AiOR among other tasks. However, LDM-based approaches are slow at inference time, a limitation we aim to overcome using visual autoregressive modeling (VAR).


\subsection{Autoregressive models in vision}
Recent works~\citep{van2016pixel,var} have extended autoregressive (AR) models to vision and can be categorized as pixel‐space AR~\citep{van2016pixel,van2016conditional,chen2019pixelsnail}, token‐based AR~\citep{oord2018neural,yu2023video,ramesh2021zeroshot} and scale‐space AR~\citep{var,ren2024mvar,guo2025fastvar}. 
Pixel-space AR predicts raw pixels one by one in raster order, as in PixelRNN~\citep{van2016pixel} and PixelCNN++~\citep{salimans2017pixelcnn++}, but is very slow at high resolutions. Token-based AR compresses images into discrete latent codes via vector quantization (e.g., VQ-VAE~\citep{oord2018neural}, VQGAN~\citep{esser2021taming}) and then models code sequences with transformers (e.g. ImageGPT~\citep{chen2020image}). This trades-off codebook size and transformer capacity against tractability for high-resolution generation. Scale-space AR, as introduced in VAR~\citep{var}, generates latents from coarse to fine scales and matches the quality of Diffusion Transformers~\citep{dit} at a fraction of the inference cost. HART~\citep{hart} scales VAR to higher resolution and uses a MLP-based diffusion refiner to convert discrete VAR latents into continuous representations. Despite VAR's success in generative tasks, it remains underexplored for image restoration with only two prior works-VarSR~\citep{varsr} and Varformer~\citep{varformer}. VarSR addressed super-resolution, while Varformer used VAR's features to guide a non-generative AiOR model. In contrast, RestoreVAR is a generative model which directly trains VAR for AiOR.

\section{Proposed Method}
\label{sec: proposed}

We first explain the working principles behind VAR for image generation. We then describe our scale-space analysis of VAR and detail RestoreVAR, our proposed VAR-based approach for AiOR.

\subsection{Preliminaries: Visual Autoregressive Modelling}
\label{subsec: prelims}

Visual Autoregressive Modelling, or VAR, is a novel autoregressive class-conditioned image generation method which uses a GPT-2~\citep{gpt2} style decoder-only transformer architecture for next-scale prediction. The VAR transformer operates in the latent space of a multi-scale VQVAE which uses $K$ scales. \begin{wrapfigure}{r}{0.52\textwidth}
    \vspace{-10pt}
    \centering
    \small
    \setlength{\tabcolsep}{0.5pt}
    \begin{tabular}{ccccc}
    &Degraded&GT$+$coarse&GT$+$fine&GT\\
        \rotatebox[origin=c]{90}{Haze\hspace{-26pt}}&\includegraphics[height=1.17cm, width=1.68cm]{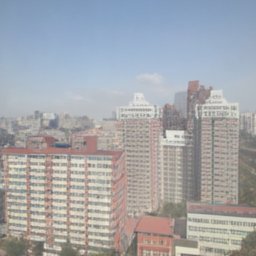}&\includegraphics[height=1.17cm, width=1.68cm]{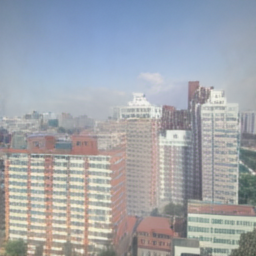}&\includegraphics[height=1.17cm, width=1.68cm]{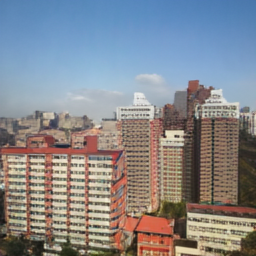}&\includegraphics[height=1.17cm, width=1.68cm]{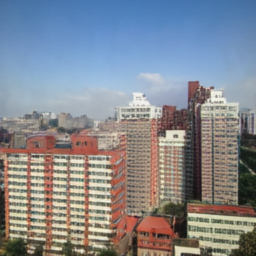}\\

        \rotatebox[origin=c]{90}{Snow\hspace{-26pt}}&\includegraphics[height=1.17cm, width=1.68cm]{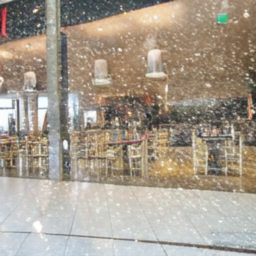}&\includegraphics[height=1.17cm, width=1.68cm]{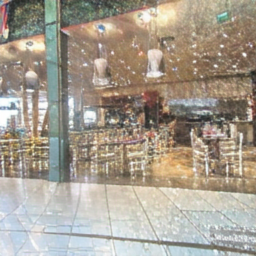}&\includegraphics[height=1.17cm, width=1.68cm]{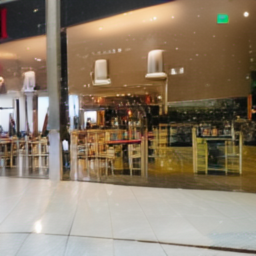}&\includegraphics[height=1.17cm, width=1.68cm]{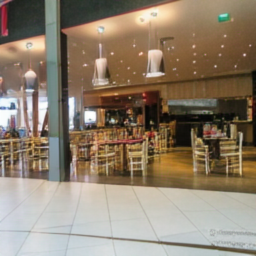}\\

        \rotatebox[origin=c]{90}{Rain\hspace{-26pt}}&\includegraphics[height=1.17cm, width=1.68cm]{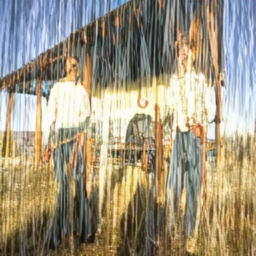}&\includegraphics[height=1.17cm, width=1.68cm]{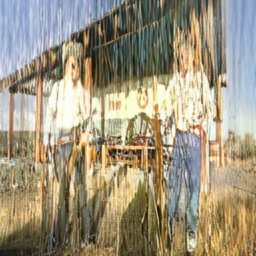}&\includegraphics[height=1.17cm, width=1.68cm]{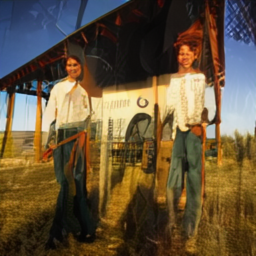}&\includegraphics[height=1.17cm, width=1.68cm]{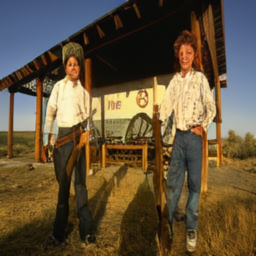}\\

        \rotatebox[origin=c]{90}{Low-light\hspace{-30pt}}&\includegraphics[height=1.17cm, width=1.68cm]{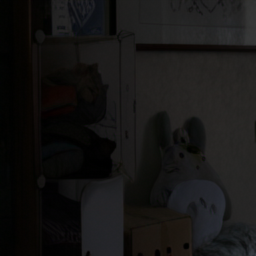}&\includegraphics[height=1.17cm, width=1.68cm]{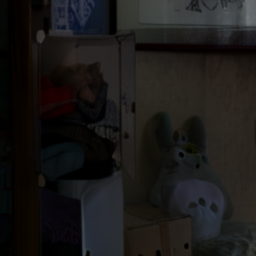}&\includegraphics[height=1.17cm, width=1.68cm]{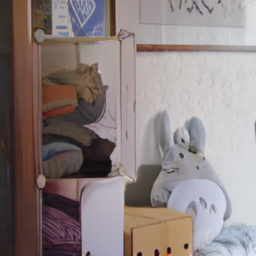}&\includegraphics[height=1.17cm, width=1.68cm]{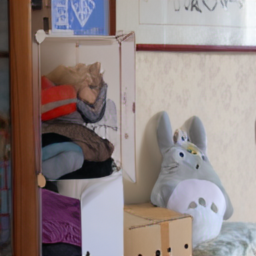}\\

        \rotatebox[origin=c]{90}{Blur\hspace{-26pt}}&\includegraphics[height=1.17cm, width=1.68cm]{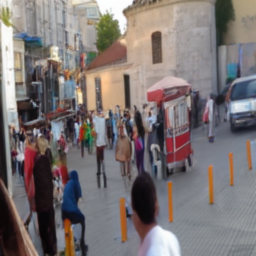}&\includegraphics[height=1.17cm, width=1.68cm]{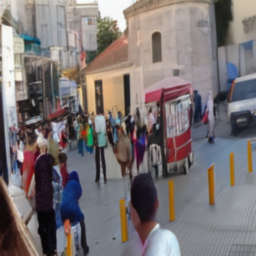}&\includegraphics[height=1.17cm, width=1.68cm]{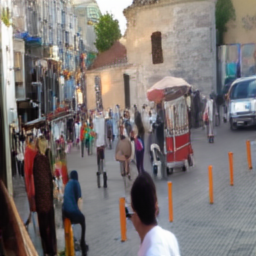}&\includegraphics[height=1.17cm, width=1.68cm]{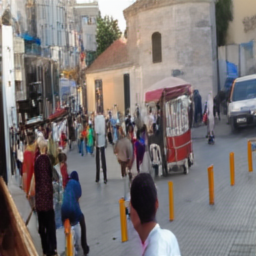}\\

    \end{tabular}
    \vspace{-8pt}
    \caption{VAR captures degradations in early scales (coarse) and scene-level details in later scales (fine). Degraded and GT are VQVAE reconstructions of the degraded and ground truth images. GT$+$coarse replaces early GT scales with degraded ones, while GT$+$fine replaces the late GT scales.}
    \label{fig: new_motivation}
    \vspace{5pt}
\end{wrapfigure} Given an image $I \in \mathbb{R}^{H \times W \times 3}$, the VQVAE encoder outputs a latent representation $f_{\text{cont}} \in \mathbb{R}^{H_K \times W_K \times C}$. Hereafter, we will refer to $f_{\text{cont}}$ as the \textit{continuous latent}, and the latent obtained after quantization as \textit{discrete latent}. Instead of directly quantizing $f_{\text{cont}}$, a multi-scale residual quantization using a shared codebook across $K$ spatial scales is performed. First, the residual and accumulated quantized (or discrete) reconstruction of $f_\text{cont}$ are initialized as $f_{\text{res}}^{(0)} := f_{\text{cont}}$ and $f_{\text{quant}}^{(0)} := 0$, respectively. At each scale $k = 1, \ldots, K$, an index map $r_k \in \mathbb{Z}^{H_k \times W_k}$ is obtained by quantizing the downsampled residual feature:\vspace{-3pt}
\[
r_k := \text{quantize}\left( \text{downsample}\left( f_{\text{res}}^{(k-1)} \right) \right). \vspace{-3pt}
\]
The indices $r_k$ are then decoded using the codebook embeddings $e(\cdot)$, upsampled to match the full resolution, and refined using a convolutional module $\phi_k(\cdot)$ to obtain \vspace{-3pt}
\[
h_k := \phi_k\left( \text{upsample}\left( e(r_k) \right) \right), \in \mathbb{R}^{H_K \times W_K \times C}. \vspace{-3pt}
\]
This is done to approximate the information captured at the current scale which is used to update the residual continuous features to be modelled by subsequent scales as \vspace{-4pt}
\[
f_{\text{quant}}^{(k)} := f_{\text{quant}}^{(k-1)} + h_k, \quad f_{\text{res}}^{(k)} := f_{\text{cont}} - f_{\text{quant}}^{(k)}. \vspace{-3pt}
\]
This process is repeated for all scales and yields a set of index maps $\{r_1, r_2, \ldots, r_K\}$, each consisting of the code-book indices of residual information at an increasingly finer scale. 


For training, VAR uses teacher-forcing, where the ground-truth index maps $\{r_1, r_2, \ldots, r_K\}$ are used to autoregressively predict the next scale. For each scale $k$, the accumulated reconstruction $f_{\text{quant}}^{(k-1)}  = \sum_{i=1}^{k-1} \phi_i\left( \text{upsample}(e(r_i)) \right)$ is interpolated to the resolution of scale $k$ to obtain $\hat{f}_{\text{quant}}^{(k)}$, which is then flattened into tokens, and concatenated with the remaining tokens to form the input sequence. A start-of-sequence (SOS) token, derived from the class label embedding, is then prepended to this input sequence. A block-wise causal attention mask is used to ensure that predictions for scale $k$ attend only to the previous scales. VAR is trained to minimize the cross-entropy loss between predicted logits and the ground-truth index maps, modeling the likelihood \vspace{-4pt}
\[
p(r_1, r_2, \ldots, r_K) = \prod_{k=1}^{K} p(r_k \mid r_1, r_2, \ldots, r_{k-1}).\vspace{-4pt} \vspace{-3pt}
\]

During inference, the SOS token is created from the target class label. VAR then autoregressively predicts each index map $r_k$, one scale at a time. After predicting $r_k$, its embedding is upsampled, refined and accumulated to form the input for the next scale, mimicking the same procedure used during training. The VAR model uses only $K = 10$ latent scales with key-value (KV) caching, enabling significantly faster inference compared to latent diffusion models.

\subsection{Scale-Space Analysis of VAR} 

\label{subsec: var_motivation}

In addition to VAR’s competitive performance to LDMs with far superior inference speed, we found that its residual scale-space decomposition focuses on degradations and scene-level details across different scales. To demonstrate this, we consider clean (GT)–degraded image pairs and compute their scale-wise residual indices $\{ r^{\text{GT}}_{k} \}_{k=1}^K$ and $\{ r^{\text{Deg}}_{k} \}_{k=1}^K$, respectively, where $K=10$. We define coarse scales as $k=1,\ldots,5$ (low-resolution index maps) and fine scales as $k=6,\ldots,10$ (higher-resolution index maps).
The first and last columns of Fig.~\ref{fig: new_motivation} show reconstructions from $r^\text{Deg}_k$ and $r^\text{GT}_k$, respectively. In column 2, we replace the coarse scales of $r^\text{GT}_k$ with those from $r^\text{Deg}_k$. This introduces the degradation, although fine scales remain unchanged. Next, in column 3 we replace the fine scales of $r^\text{GT}_k$ with those from $r^\text{Deg}_k$, which yields a clean image with some loss of fine details. These observations indicate that coarse scales in VAR capture degradations, while finer scales encode scene-level detail. Notably, this observation holds across multiple degradations and simplifies restoration for VAR as removing degradations requires correctly predicting only the early scales which contain a small number of tokens, while scene details can be reconstructed in subsequent scales. 


\subsection{RestoreVAR}
\label{subsec: restorevar}


We now describe RestoreVAR, our proposed approach that effectively adapts VAR for AiOR, leveraging its substantial inference speed advantage over LDMs. Given a degraded image $I_{\text{deg}} \in \mathbb{R}^{H \times W \times 3}$, the goal is to predict a clean output $I_{\text{clean}}$, close to the ground-truth $I_{\text{gt}}$. Adapting VAR to AiOR is non-trivial due to the need for high-quality pixel-level reconstruction, which is compromised by two factors: (1) VAR's strong generative priors can cause hallucinations in the restored images without proper conditioning. (2) Vector quantization and VAE decoding introduce artifacts that hinder pixel-level restoration. RestoreVAR addresses these challenges through architectural enhancements, including cross-attention to incorporate semantic guidance from the degraded image, and a novel  non-generative transformer that refines discrete latents into their continuous form to preserve fine details in the restored image. We describe these components below.


\begin{figure*}
    \centering
    \includegraphics[width=1\linewidth]{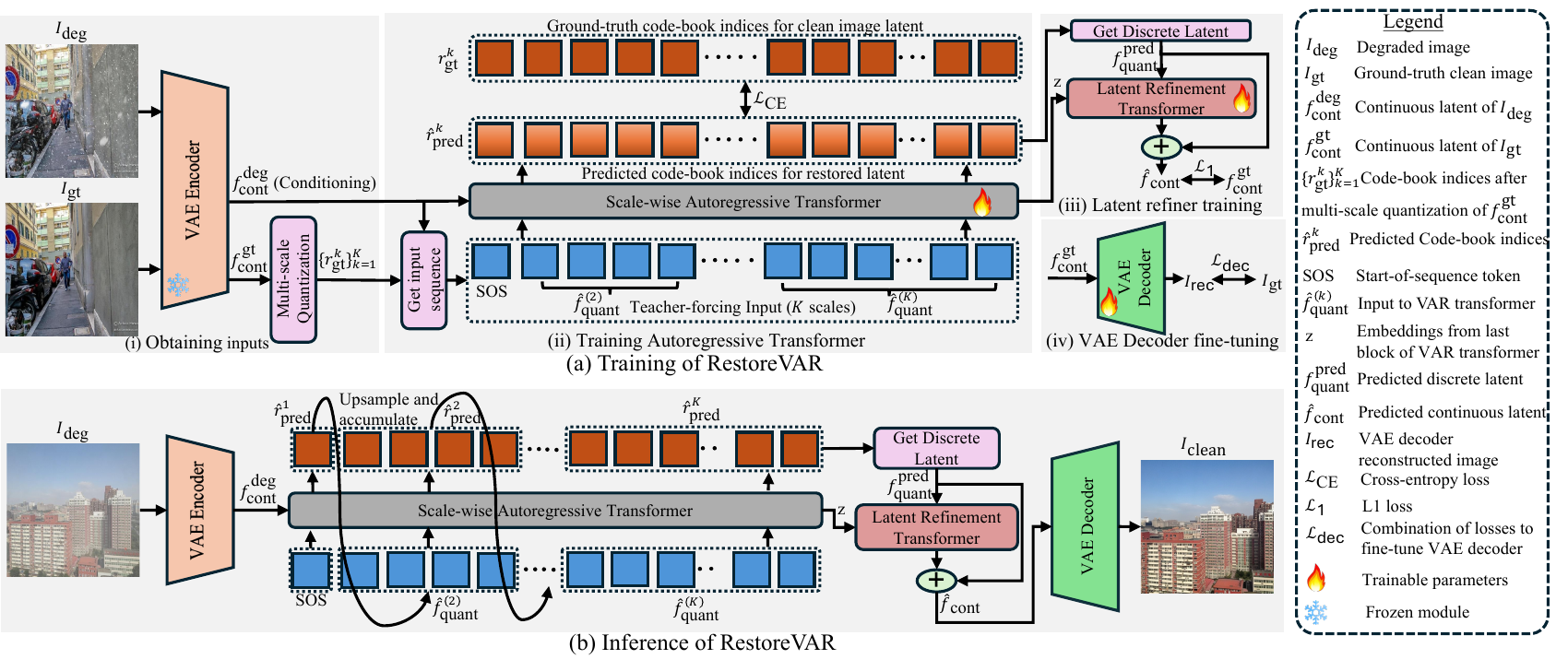}
    \vspace{-16pt}
    \caption{Illustration of RestoreVAR for training and inference. (a) Shows the training procedure for each component of RestoreVAR, and (b) shows the overall pipeline during inference.}
    \label{fig: block}
    \vspace{-12pt}
\end{figure*}
\subsubsection{Autoregressive Transformer Architecture}
\label{subsubsec: arch}

For training, the multi-scale teacher-forcing input is constructed from the ground-truth image $I_{\text{gt}}$ (see Sec.~\ref{subsec: prelims}). The start-of-sequence (SOS) token is computed from a fixed label index and augmented with a global context vector derived from the degraded image (see supplementary for details). These features are flattened and concatenated into a token sequence $\hat{f}_\text{quant} \in \mathbb{R}^{L \times C}$, where $L$ is the total number of tokens across all scales (see Fig.~\ref{fig: block}(a)(i)). The VAR transformer is then trained to autoregressively predict the next-scale indices $\{r^k_{\text{gt}}\}_{k=1}^K \in \mathbb{R}^{L}$ of the clean image.

To enable semantically consistent restoration, we inject information from the degraded image through cross-attention at each transformer block. At block $i$, the queries are given by the output of the feed-forward network ($x_{\text{block}_i} \in \mathbb{R}^{L \times D}$, where $D$ is the embedding dimension), while keys and values are derived from the continuous latent of the degraded image, $f_{\text{cont}}^{\text{deg}} \in \mathbb{R}^{H_K \times W_K \times C}$. This latent is reshaped into a sequence of conditioning tokens and is appropriately projected to the embedding dimension of the transformer. As shown in Sec.~\ref{subsec: ablations}, conditioning on continuous latents significantly outperforms conditioning on discrete ones. To summarize, cross-attention ($\text{CA}(\cdot,\cdot)$) is applied as \vspace{-4pt}
\[
x_{\text{block}_\text{CA}} = x_{\text{block}_i} + g_i \times \text{CA}(x_{\text{block}_i}, f_{\text{cont}}^{\text{deg}}).
\]
We initialize $g_i = 0$ to retain VAR’s pretrained behavior and gradually introduce conditioning. Furthermore, we replace absolute positional embeddings in VAR with 2D Rotary Positional Embeddings (RoPE) for scaling resolution from $256 \times 256$ to $512 \times 512$, as RoPE is well-suited for handling varying sequence lengths~\citep{rope}. We also remove AdaLN layers, reducing $\sim100$M parameters with negligible impact on performance. Inference closely follows that of VAR (see Sec.~\ref{subsec: prelims}), except that each scale prediction is now guided by the degraded latent. The output is a sequence of predicted indices $\{\hat{r}^k_{\text{pred}}\}_{k=1}^K$, which is then used to construct the discrete restored latent $f^\text{pred}_{\text{quant}}  \in \mathbb{R}^{H_K \times W_K \times C}$. The above steps are shown in Fig.~\ref{fig: block}(a)(ii). More architectural details are given in the supplementary.


\subsubsection{Detail-Preserving Restoration}
\label{subssubsec: lrm}

The discrete latent ($f^\text{pred}_{\text{quant}}$) predicted by the RestoreVAR transformer is decoded by the VQVAE to produce the restored image. However, vector-quantization and VAE decoding cause a noticeable loss of fine details in the pixel-space, leading to distorted reconstructions. This presents a major challenge for using VAR in AiOR, as the scene semantics may not be accurately preserved. To address this, we introduce VAE decoder fine-tuning on continuous latents, and a lightweight latent refinement transformer (LRT) that converts discrete latents to continuous latents for decoding.





\textbf{VAE Decoder Fine-Tuning. }HART~\citep{hart} addressed VAE-induced distortions by fine-tuning the VAE decoder on both discrete and continuous latents. While effective for generative tasks, the VAE decoder of HART produces overly textured outputs, compromising accurate reconstruction (see supplementary). Instead, we fine-tune the decoder only on continuous latents, bypassing the quantizer. The encoder and quantizer are kept frozen, and the decoder is trained on $(f^{\text{gt}}_{\text{cont}}, I_{\text{gt}})$ pairs. To avoid overly smooth outputs, we use a PatchGAN~\citep{patchgan} discriminator (see Sec.~\ref{subsec: ablations}) and optimize the decoder using pixel-wise, perceptual, and adversarial losses as\vspace{-2pt}
\[
\mathcal{L}_{\text{dec}} = \lambda_1 \mathcal{L}_{\text{L1}} + \lambda_2 \mathcal{L}_{\text{SSIM}} + \lambda_3 \mathcal{L}_{\text{percep}} + \lambda_4 \mathcal{L}_{\text{adv}},\vspace{-2pt}
\]
where $\mathcal{L}_{\text{L1}}$ is the L1 loss, $\mathcal{L}_{\text{SSIM}}$ is the SSIM loss, $\mathcal{L}_{\text{percep}}$ is the perceptual loss, $\mathcal{L}_{\text{adv}}$ is the adversarial loss and $\lambda_i$ are their respective weights (see Fig.~\ref{fig: block}(a)(iv)). Our fine-tuning approach yields a decoder that is well-aligned with the objectives of AiOR, achieving mean (over $1000$ samples) reconstruction PSNR/SSIM scores of $28.14$dB/$0.842$, outperforming both the VAR VQVAE ($22.59$dB/$0.679$) and HART decoders ($26.48$dB/$0.804$). Qualitative comparisons are given in the supplementary.

\textbf{Refining Discrete Latents. }Since the VAE decoder is fine-tuned for continuous latents, the predicted discrete latent, $f^\text{pred}_{\text{quant}}$, must be converted into a continuous form for decoding. \begin{wrapfigure}{t}{0.485\textwidth}
        \centering
        \vspace{-10pt}
        \begin{subfigure}[t]{0.24\linewidth}
            \includegraphics[width=\linewidth]{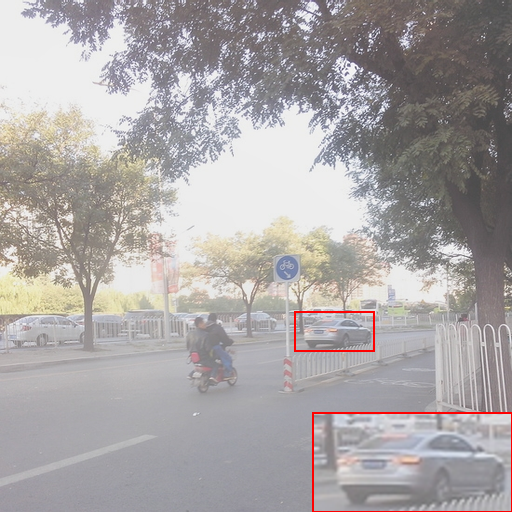}
            \caption*{Input}
        \end{subfigure}\hspace{0.0001\linewidth}
        \begin{subfigure}[t]{0.24\linewidth}
            \includegraphics[width=\linewidth]{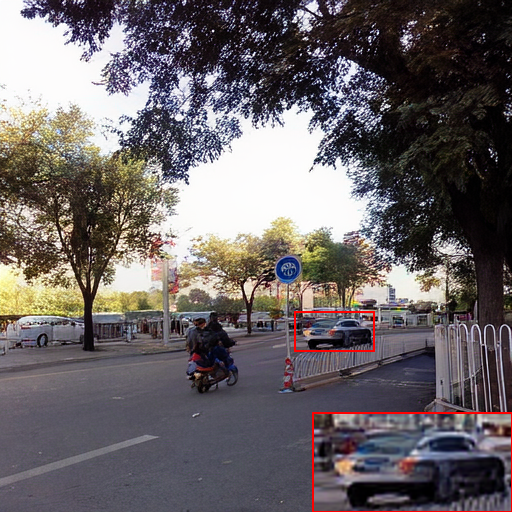}
            \caption*{Discrete}
        \end{subfigure}\hspace{0.0001\linewidth}
        \begin{subfigure}[t]{0.24\linewidth}
            \includegraphics[width=\linewidth]{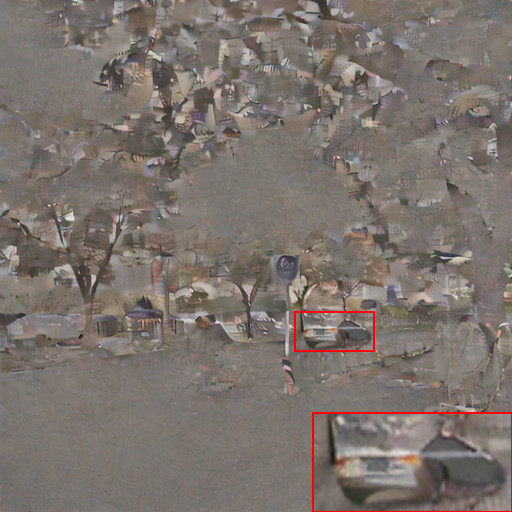}
            \caption*{Refiner}
        \end{subfigure}\hspace{0.0001\linewidth}
        \begin{subfigure}[t]{0.24\linewidth}
            \includegraphics[width=\linewidth]{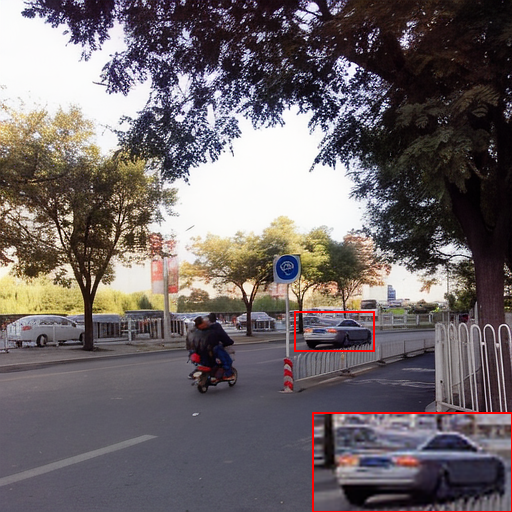}
            \caption*{Continuous}
        \end{subfigure}
        \vspace{-10pt}
        \caption{Illustration of images decoded from discrete and continuous latents, along with the refiner’s predicted residuals.}
        \label{fig:refiner_illustration}
        \vspace{-4pt}
    \end{wrapfigure} While HART uses a $37$M parameter diffusion-based MLP for this, it incurs a $\sim20\%$ inference overhead due to iterative denoising. Instead, we propose a lightweight, non-generative latent refinement transformer (LRT) that predicts a residual, which when added to $f^\text{pred}_{\text{quant}}$, produces a continuous latent, $\hat{f}_{\text{cont}} \in \mathbb{R}^{H_K \times W_K \times C}$ as
\[
\hat{f}_{\text{cont}} = f^\text{pred}_{\text{quant}} + \text{LRM}(f^\text{pred}_{\text{quant}}, z),
\]
where $z \in \mathbb{R}^{L \times D}$ is the output from the final RestoreVAR transformer block. $z$ is passed through cross-attention and provides pseudo-continuous guidance to the LRT which is critical for performance (see Sec.~\ref{subsec: ablations}). The LRT is trained using $\mathcal{L}_1$ loss between the predicted and ground-truth continuous latents ($f_{\text{cont}}^{\text{gt}}$) as $\mathcal{L}_{\text{LRT}} = \mathcal{L}_1(\hat{f}_{\text{cont}}, f_{\text{cont}}^{\text{gt}})$.
Our LRT introduces only $3\%$ additional overhead and significantly outperforms HART's refiner in PSNR and SSIM scores (see Sec.~\ref{subsec: ablations}). The training procedure of the LRT is shown in Fig.~\ref{fig: block}(a)(iii) and Fig.~\ref{fig:refiner_illustration} provides a visual example of its predictions.

Thus, RestoreVAR combines the VAR transformer, LRT, and fine-tuned decoder to deliver fast, perceptually realistic, and structurally faithful results. Fig.~\ref{fig: block}(b) depicts inference of RestoreVAR.

\section{Experiments}
\label{sec: expts}
In this section, we provide implementation details, comparisons with existing All-in-One image Restoration (AiOR) approaches, and present ablations on key components of our framework.

\subsection{Implementation Details}
\label{subsec: impl}
Each component of RestoreVAR was trained independently to disentangle learning objectives. We used the VAR model of depth $16$ as the transformer backbone and trained it with the AdamW optimizer~\citep{adamw}, a learning rate (LR) of $10^{-4}$, batch size of $48$, for $100$ epochs. The latent refiner was trained for $100$ epochs with the AdamW optimizer, LR$=10^{-4}$ and a batch size of $96$. The VAE decoder was fine-tuned using a weighted loss combination (see Sec.~\ref{subssubsec: lrm}) with empirically chosen weights: $\lambda_1 = 2.0$, $\lambda_2 = 0.4$, $\lambda_3 = 0.2$, and $\lambda_4 = 0.01$. Fine-tuning was performed for $5$ epochs with a learning rate of $3 \times 10^{-4}$ and a batch size of $12$, using AdamW. Training was conducted on $8$ RTX A6000 GPUs, while inference was done on an RTX 4090 GPU.

\subsection{Datasets}
\label{subsec: datasets}
We trained RestoreVAR for five tasks: dehazing, desnowing, deraining, low-light enhancement and deblurring. For dehazing, we used the RESIDE~\citep{reside} dataset comprising $72135$ training and $500$ test images. The Snow100k dataset~\citep{snow100k} was used for desnowing, with $50000$ training and $16801$ test images (heavy subset). For deraining, we used Rain13K~\citep{mprnet} consisting of $13711$ training and $4298$ test images. The LOLv1~\citep{lolv1} dataset was used for low-light enhancement, consisting of $485$ training and $15$ test images. For deblurring, we used the GoPro~\citep{gopro} dataset comprising $2103$ training and $1111$ test images. We also assess generalization performance on real-world, unseen and mixed degradation datasets, namely, LHP~\citep{lhprain} ($1000$ images), REVIDE~\citep{revide} ($284$ images), TOLED~\citep{toledpoled} ($30$ images), POLED~\citep{toledpoled} ($30$ images), CDD~\citep{cdd} ($200$ images, mix of haze and rain), and LOLBlur~\citep{lolblur} ($482$ images, mix of low-light and blur). TOLED and POLED datasets contain unseen degradation of under-display camera restoration.

\begin{table*}[t]
  \centering
  \small
  \caption{Quantitative comparisons of RestoreVAR with the state-of-the-art LDM-based generative AiOR approaches, and non-generative methods. RestoreVAR significantly outperforms generative methods on PSNR, SSIM and LPIPS scores. The best generative approach is indicated in \textbf{bold}.}
  \vspace{-8pt}
  \setlength{\tabcolsep}{2pt}
  \resizebox{\textwidth}{!}{%
    \begin{tabular}{l l
      *{3}{c}  
      *{3}{c}  
      *{3}{c}  
      *{3}{c}  
      *{3}{c}  
    }
      \toprule
      \multirow{2}{*}{\textbf{Method}} 
        & \multirow{2}{*}{\textbf{Venue}}
        & \multicolumn{3}{c}{\textbf{RESIDE}} 
        & \multicolumn{3}{c}{\textbf{Snow100k}} 
        & \multicolumn{3}{c}{\textbf{Rain13K}} 
        & \multicolumn{3}{c}{\textbf{LOLv1}} 
        & \multicolumn{3}{c}{\textbf{GoPro}} \\
      \cmidrule(lr){3-5}
      \cmidrule(lr){6-8}
      \cmidrule(lr){9-11}
      \cmidrule(lr){12-14}
      \cmidrule(lr){15-17}
        & 
        & PSNR$\uparrow$ & SSIM$\uparrow$ & LPIPS$\downarrow$
        & PSNR$\uparrow$ & SSIM$\uparrow$ & LPIPS$\downarrow$
        & PSNR$\uparrow$ & SSIM$\uparrow$ & LPIPS$\downarrow$
        & PSNR$\uparrow$ & SSIM$\uparrow$ & LPIPS$\downarrow$
        & PSNR$\uparrow$ & SSIM$\uparrow$ & LPIPS$\downarrow$ \\
      \midrule
      \rowcolor[HTML]{F4CCCC}
      \multicolumn{17}{c}{\textit{\textbf{Non‐generative methods}}} \\
      PromptIR  
        & NeurIPS'23
        & 32.02 & 0.952 & 0.013
        & 31.98 & 0.924 & 0.115
        & 29.56 & 0.888 & 0.087
        & 22.89 & 0.847 & 0.296
        & 27.21 & 0.817 & 0.250 \\
      InstructIR 
        & ECCV'24
        & 26.90 & 0.952 & 0.017
        &   --  &   --  &  --
        & 29.56 & 0.885 & 0.088
        & 22.81 & 0.836 & 0.132
        & 28.26 & 0.870 & 0.146 \\
      AWRaCLe    
        & AAAI'25
        & 30.81 & 0.979 & 0.013
        & 30.56 & 0.904 & 0.088
        & 31.26 & 0.908 & 0.068
        & 21.04 & 0.818 & 0.146
        & 26.78 & 0.820 & 0.248 \\
      DCPT    
        & ICLR'25
        & 29.10 & 0.968 & 0.017
        &   --  &   --  &  --
        & 24.11 & 0.766 & 0.203
        & 23.67 & 0.863 & 0.106
        & 27.92 & 0.877 & 0.169 \\
      DFPIR
        & CVPR'25
        & 31.39 & 0.979 & 0.012
        &   --  &   --  &  --
        & 24.87 & 0.794 & 0.171
        & 23.12 & 0.853 & 0.123
        & 28.66 & 0.884 & 0.158 \\

      \midrule
      \rowcolor[HTML]{D9D2E9}
      \multicolumn{17}{c}{\textit{\textbf{Generative methods}}} \\
      \rowcolor{gray!15}
      Diff-Plugin
        & CVPR'24
        & 23.23 & 0.765 & 0.091
        & 21.02 & 0.611 & 0.196
        & 21.71 & 0.617 & 0.169
        & 19.38 & 0.713 & 0.195
        & 21.76 & 0.633 & 0.217 \\
        \rowcolor{gray!15}
      AutoDIR    
        & ECCV'24
        & 24.48 & 0.780 & 0.081
        & 19.00 & 0.515 & 0.347
        & 23.02 & 0.642 & 0.162
        & 19.43 & 0.766 & 0.135
        & 23.55 & 0.700 & 0.168 \\
        \rowcolor{gray!15}
      PixWizard  
        & ICLR'25
        & 21.28 & 0.738 & 0.142
        & 21.24 & 0.594 & 0.206
        & 21.38 & 0.596 & 0.180
        & 15.84 & 0.629 & 0.305
        & 20.49 & 0.602 & 0.223 \\
        \midrule
        \rowcolor{gray!15}
      \textbf{RestoreVAR (Ours)}  
        & 
        & \textbf{24.67} & \textbf{0.821} & \textbf{0.074}
        & \textbf{24.05} & \textbf{0.713} & \textbf{0.156}
        & \textbf{23.97} & \textbf{0.700} & \textbf{0.153}
        & \textbf{21.72} & \textbf{0.782} & \textbf{0.126}
        & \textbf{23.96} & \textbf{0.737} & \textbf{0.167} \\
      \bottomrule
    \end{tabular}%
  } 
  \label{tab:within_dist}
  \vspace{-5pt}
\end{table*}

\begin{figure*}
    \centering
    \small
    \setlength{\tabcolsep}{1pt}
    \begin{tabular}{ccccccc}
    &Input&Diff-Plugin&AutoDIR&PixWizard&RestoreVAR&GT\\
          \rotatebox[origin=c]{90}{RESIDE\hspace{-32pt}}&\includegraphics[height=1.37cm, width=2.1cm]{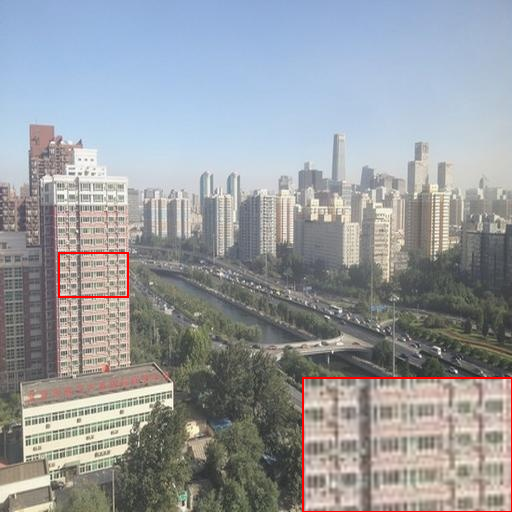}&\includegraphics[height=1.37cm, width=2.1cm]{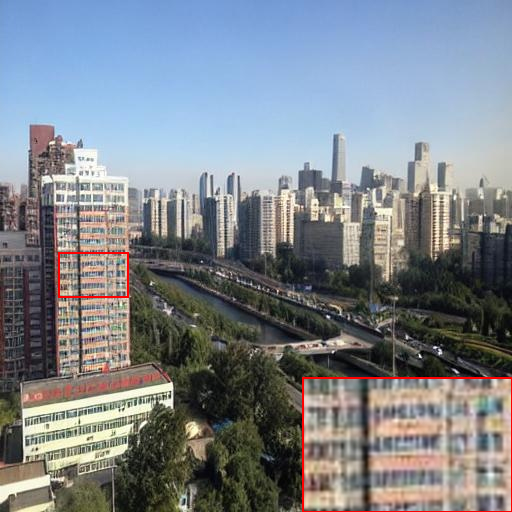}&\includegraphics[height=1.37cm, width=2.1cm]{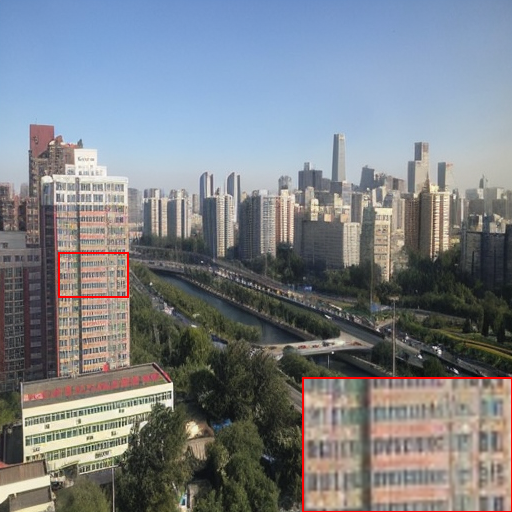}&\includegraphics[height=1.37cm, width=2.1cm]{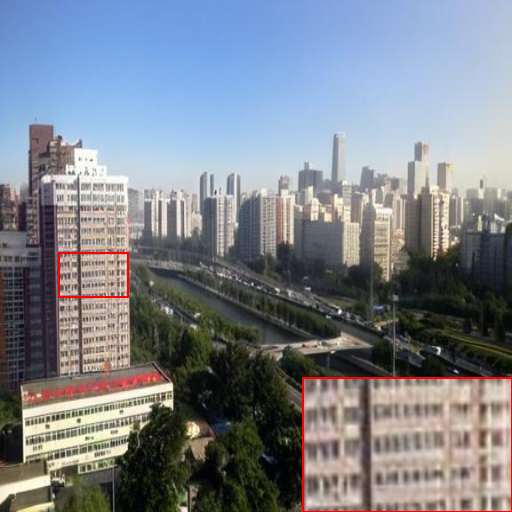}&\includegraphics[height=1.37cm, width=2.1cm]{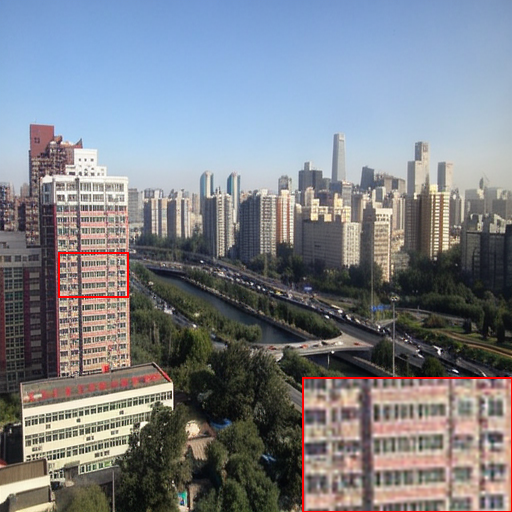}&\includegraphics[height=1.37cm, width=2.1cm]{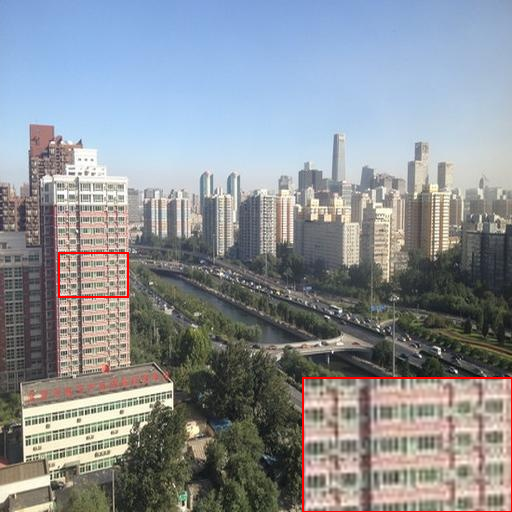}  \\
         

         \rotatebox[origin=c]{90}{Rain13K\hspace{-32pt}}&\includegraphics[height=1.37cm, width=2.1cm]{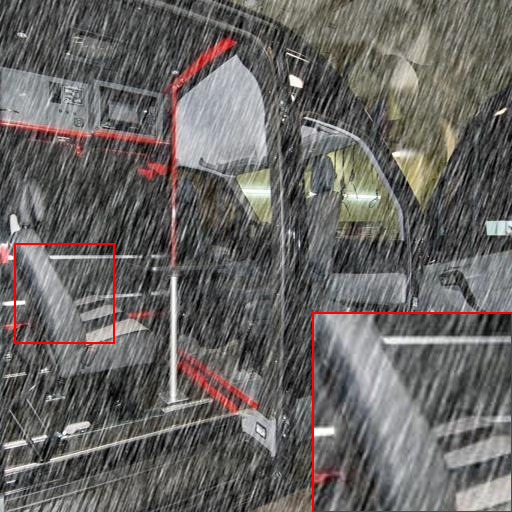}&\includegraphics[height=1.37cm, width=2.1cm]{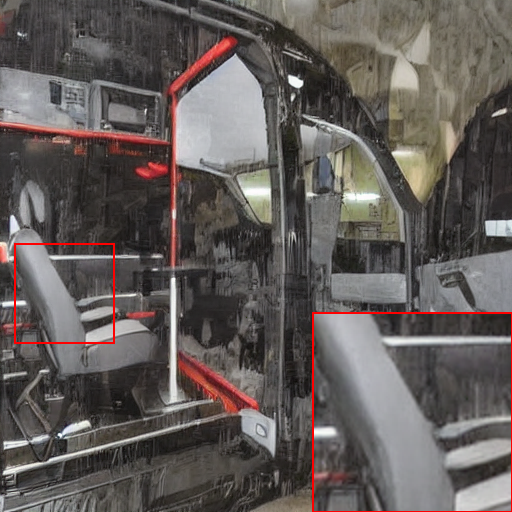}&\includegraphics[height=1.37cm, width=2.1cm]{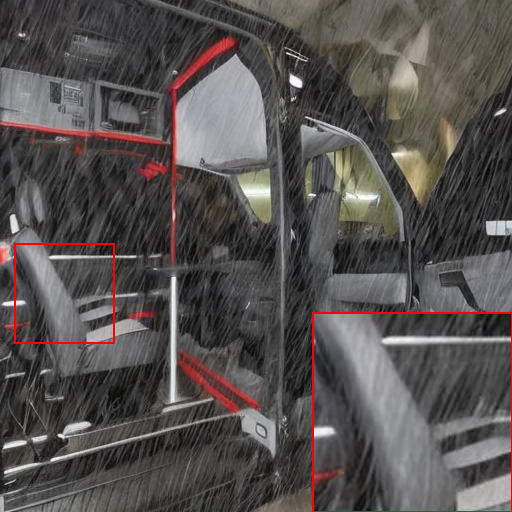}&\includegraphics[height=1.37cm, width=2.1cm]{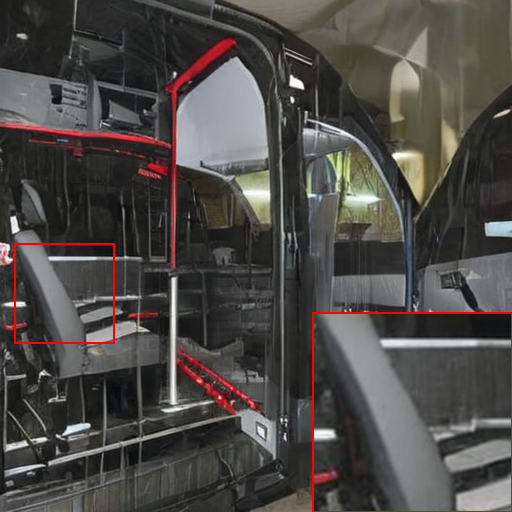}&\includegraphics[height=1.37cm, width=2.1cm]{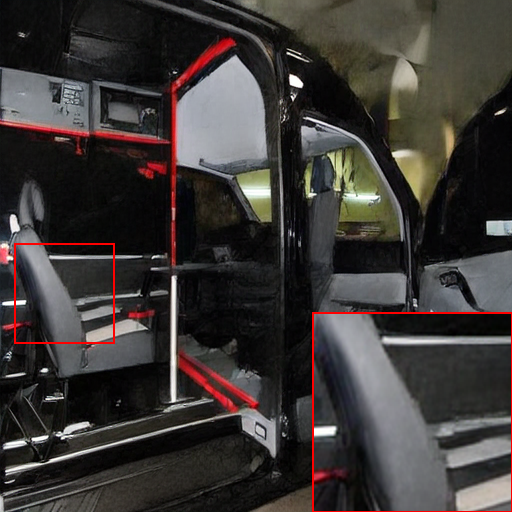}&\includegraphics[height=1.37cm, width=2.1cm]{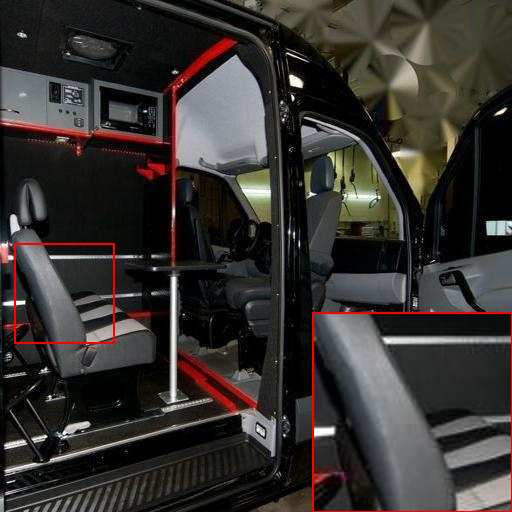}  \\

         \rotatebox[origin=c]{90}{GoPro\hspace{-32pt}}&\includegraphics[height=1.37cm, width=2.1cm]{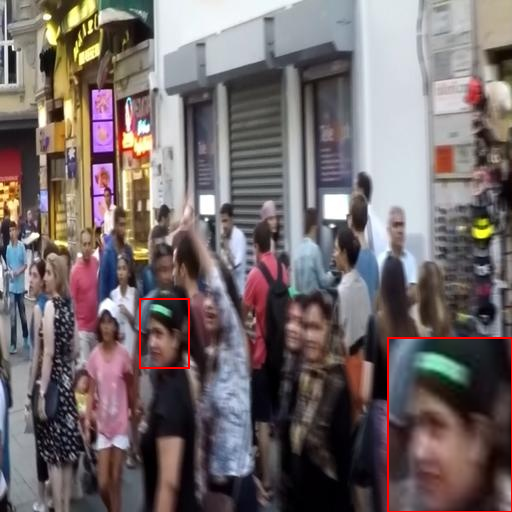}&\includegraphics[height=1.37cm, width=2.1cm]{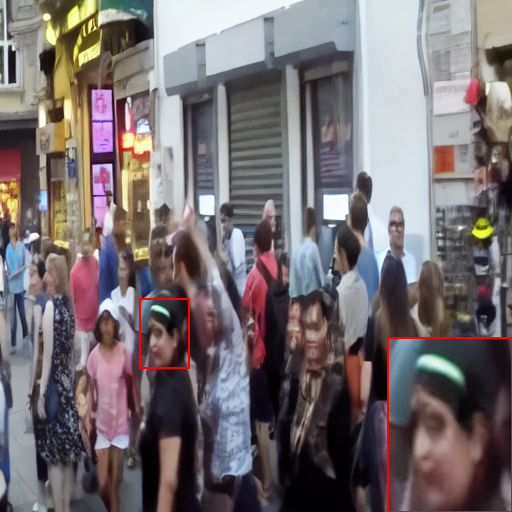}&\includegraphics[height=1.37cm, width=2.1cm]{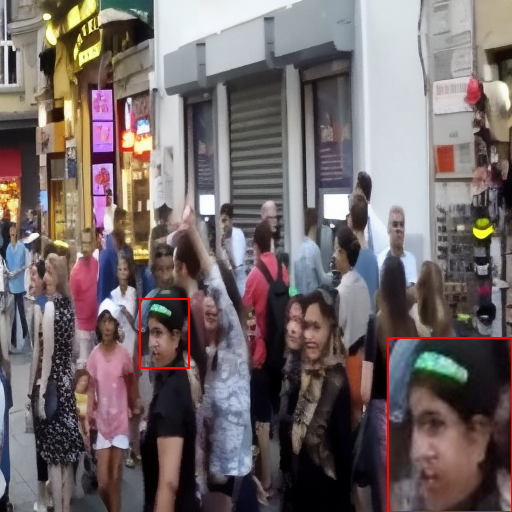}&\includegraphics[height=1.37cm, width=2.1cm]{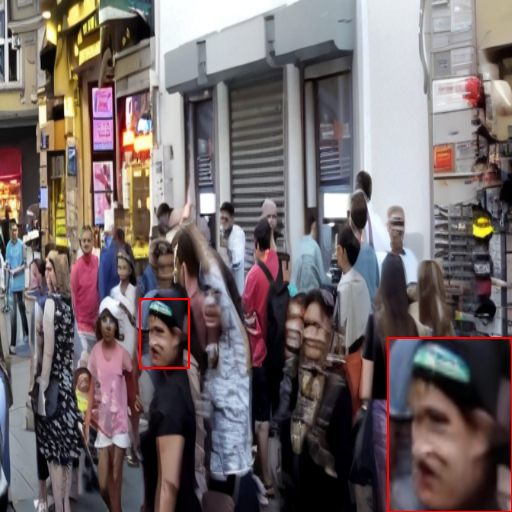}&\includegraphics[height=1.37cm, width=2.1cm]{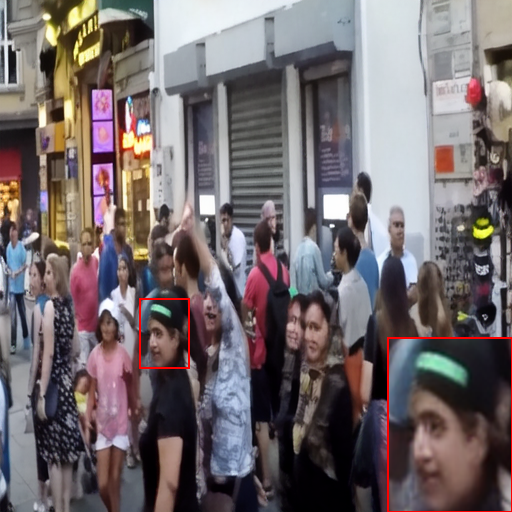}&\includegraphics[height=1.37cm, width=2.1cm]{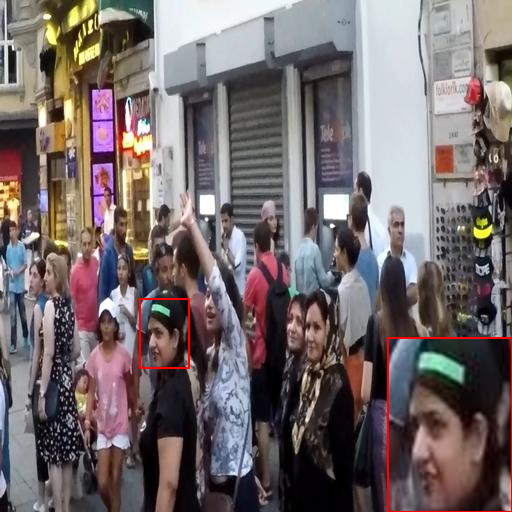} 
    \end{tabular}
    \vspace{-8pt}
    \caption{Qualitative comparisons of RestoreVAR with LDM-based generative AiOR approaches. RestoreVAR achieves consistent restoration with enhanced preservation of fine-details.}
    \label{fig: main_qual}
    \vspace{-12pt}
\end{figure*}
\subsection{Comparisons}
\label{subsec: comparisons}
We compare RestoreVAR with state-of-the-art generative and non-generative methods for AiOR. For non-generative approaches, we include PromptIR~\citep{promptir}, InstructIR~\citep{instructir}, AWRaCLe~\citep{awracle}, DCPT~\citep{dcpt} and DFPIR~\citep{dfpir}. Among generative methods, we compare with the LDM-based approaches Diff-Plugin~\citep{diffplugin}, AutoDIR~\citep{autodir} and PixWizard~\citep{pixwizard}. To ensure a fair comparison, we retrained PromptIR and AWRaCLe, as their official checkpoints were not trained for most of our AiOR tasks. All other methods were evaluated using their publicly released checkpoints. For AutoDIR, we report results without the structure correction module, as this module functions as an independent, non-generative restoration network (more details in supplementary). The results reported for PixWizard were obtained using its publicly released checkpoint. We do not compare with task-specific restoration models, as RestoreVAR is proposed for the AiOR setting.



Table~\ref{tab:within_dist} presents PSNR, SSIM and LPIPS scores on the RESIDE, Snow100k, Rain13K, LOLv1 and GoPro datasets. RestoreVAR surpasses LDM-based AiOR methods at a fraction of their computational cost (inference time (s) per image)---Diff-Plugin:$2.04$s, AutoDIR: $8.477$s, PixWizard: $8.247$s and RestoreVAR: $0.201$s, highlighting the efficacy of our framework. More detailed complexity comparisons are given in the supplementary along with a derivation showing that the time complexity of VAR with maximum latent resolution $n\times n$ is $\mathcal{O}(\log n)$ lower than an LDM operating at the same latent resolution. Qualitative comparisons with LDM-based methods in Fig.~\ref{fig: main_qual} further illustrate that RestoreVAR produces restored images of high quality while better preserving fine details. Visual results for the Snow100k and LOLv1 datasets are provided in the supplementary. While non-generative methods achieve better scores, it is important to recognize that the performance of RestoreVAR is inherently influenced by the quality of the VAE decoder; a limitation shared by all latent generative approaches. Despite this constraint, RestoreVAR narrows the gap with non-generative methods while maintaining the benefits of a generative framework, i.e., perceptually realistic results and strong generalization capabilities. To demonstrate these strengths, we evaluate generalization using no-reference image quality metrics (following prior works~\citep{diffplugin,autodir,awracle}), and assess perceptual realism through a user study.

\begin{table*}[t]
  \centering
  \small
  \caption{Quantitative comparisons of RestoreVAR against state-of-the-art non-generative and generative approaches on real-world, unseen and mixed degradations. The best result among RestoreVAR and non-generative approaches is indicated in bold.}
  \vspace{-7pt}
  \setlength{\tabcolsep}{2pt}
  \resizebox{\textwidth}{!}{%
    \begin{tabular}{l cc cc cc cc cc cc cc}
      \toprule
      \multirow{2}{*}{\textbf{Method}}
        & \multicolumn{2}{c}{\textbf{LHP}}
        & \multicolumn{2}{c}{\textbf{REVIDE}}
        & \multicolumn{2}{c}{\textbf{TOLED}}
        & \multicolumn{2}{c}{\textbf{POLED}}
        & \multicolumn{2}{c}{\textbf{LOLBlur (L $+$ B)}}
        & \multicolumn{2}{c}{\textbf{CDD (H $+$ R)}}
        & \multicolumn{2}{c}{\textbf{Average}} \\
      \cmidrule(lr){2-3}
      \cmidrule(lr){4-5}
      \cmidrule(lr){6-7}
      \cmidrule(lr){8-9}
      \cmidrule(lr){10-11}
      \cmidrule(lr){12-13}
      \cmidrule(lr){14-15}
        & MUSIQ$\uparrow$ & CLIPIQA$\uparrow$
        & MUSIQ$\uparrow$ & CLIPIQA$\uparrow$
        & MUSIQ$\uparrow$ & CLIPIQA$\uparrow$
        & MUSIQ$\uparrow$ & CLIPIQA$\uparrow$
        & MUSIQ$\uparrow$ & CLIPIQA$\uparrow$
        & MUSIQ$\uparrow$ & CLIPIQA$\uparrow$
        & MUSIQ$\uparrow$ & CLIPIQA$\uparrow$ \\
      \midrule
      PromptIR
        & 56.780 & 0.366
        & 61.191 & 0.459
        & 43.218 & 0.281
        & 34.536 & 0.303
        & 33.693 & 0.166
        & 65.895 & 0.483
        & 49.219 & 0.343 \\
      InstructIR
        & \textbf{58.269} & 0.359
        & 63.116 & 0.416
        & 44.985 & 0.298
        & 23.317 & 0.241
        & 40.221 & 0.202
        & 65.491 & 0.482
        & 49.900 & 0.333 \\
      AWRaCLe
        & 57.889 & 0.333
        & 59.287 & 0.368
        & 44.670 & 0.285
        & 40.533 & 0.332
        & 38.186 & 0.171
        & 66.253 & 0.484
        & 51.470 & 0.329 \\
      DCPT
        & 58.044 & 0.372
        & 60.011 & 0.446
        & 44.062 & 0.314
        & 38.138 & \textbf{0.345}
        & 37.393 & 0.175
        & 68.440 & 0.544
        & 51.681 & 0.366 \\
      DFPIR
        & 56.483 & 0.330
        & 61.009 & 0.450
        & 43.820 & 0.276
        & 35.668 & 0.289
        & 36.277 & 0.163
        & 54.408 & 0.349
        & 47.611 & 0.310 \\
        \hline
        \rowcolor{gray!10}
        DiffPlugin
        & 57.351 & 0.420
        & 63.483 & 0.406 
        & 46.219 & 0.311
        & 35.086 & 0.407
        & 40.054 & 0.212
        & 68.791 & 0.578
        & 51.831 & 0.389 \\
        \rowcolor{gray!10}
      AutoDir
        & 58.085 & 0.380
        & 63.918 & 0.416
        & 54.585 & 0.341
        & 48.796 & 0.380
        & 46.642 & 0.225
        & 68.575 & 0.571
        & 56.767 & 0.386 \\
    \rowcolor{gray!10}
      PixWizard
        & 58.600 & 0.411
        & 68.487 & 0.420
        & 45.804 & 0.329
        & 44.305 & 0.348
        & 50.708 & 0.268
        & 68.409 & 0.589
        & 56.052 & 0.394 \\
        \hline
      \rowcolor{gray!25}
      \textbf{RestoreVAR}
        & 57.662 & \textbf{0.414}
        & \textbf{63.562} & \textbf{0.483}
        & \textbf{52.374} & \textbf{0.338}
        & \textbf{48.118} & 0.276
        & \textbf{46.644} & \textbf{0.214}
        & \textbf{68.941} & \textbf{0.572}
        & \textbf{56.217} & \textbf{0.383} \\
      \bottomrule
    \end{tabular}%
  }
  \label{tab:generalization_metrics}
  \vspace{-5pt}
\end{table*}

\begin{figure*}[t]
    \centering
    \small
    \setlength{\tabcolsep}{1pt}
    \begin{tabular}{cccccccc}
    &Input&InstructIR&AWRaCLe&DCPT&DFPIR&RestoreVAR&GT\\
         \rotatebox[origin=c]{90}{LHP\hspace{-33pt}}&\includegraphics[height=1.37cm, width=1.88cm]{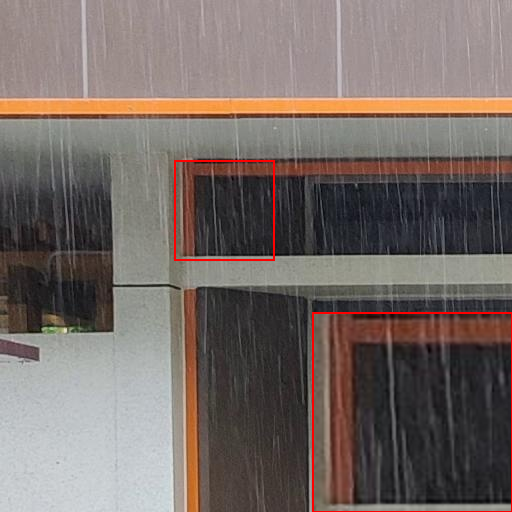}&\includegraphics[height=1.37cm, width=1.88cm]{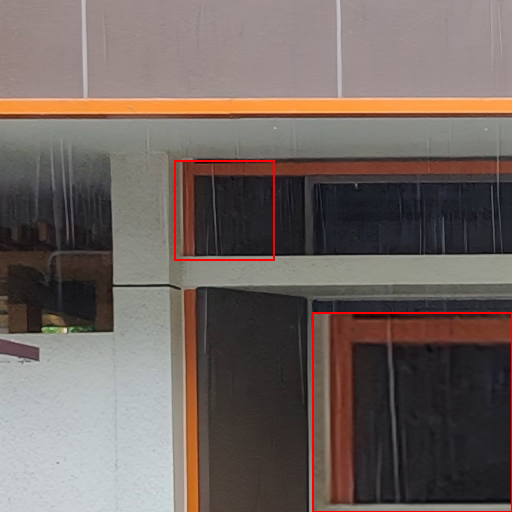}&\includegraphics[height=1.37cm, width=1.88cm]{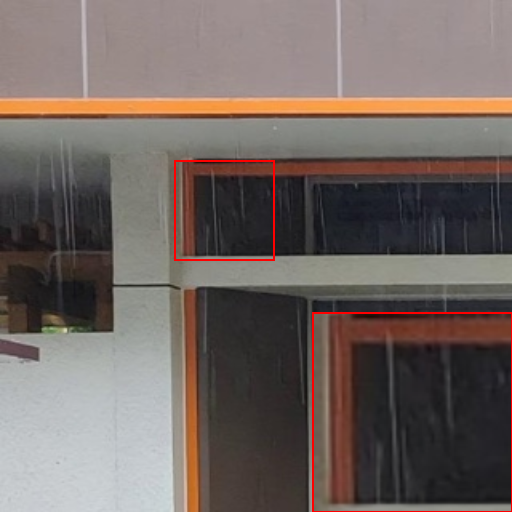}&\includegraphics[height=1.37cm, width=1.88cm]{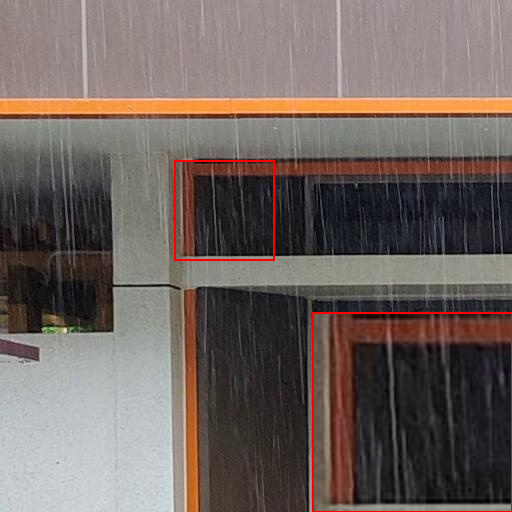}&\includegraphics[height=1.37cm, width=1.88cm]{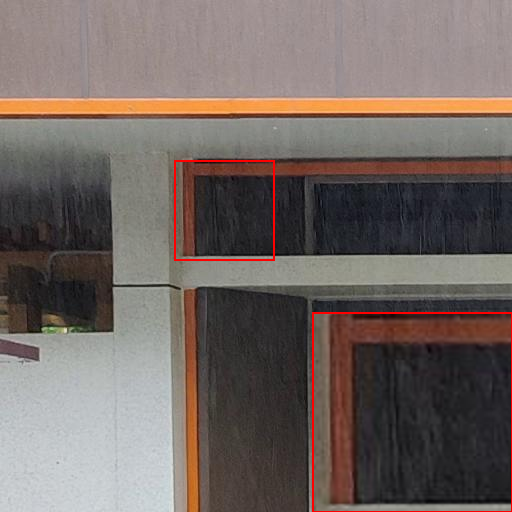}&\includegraphics[height=1.37cm, width=1.88cm]{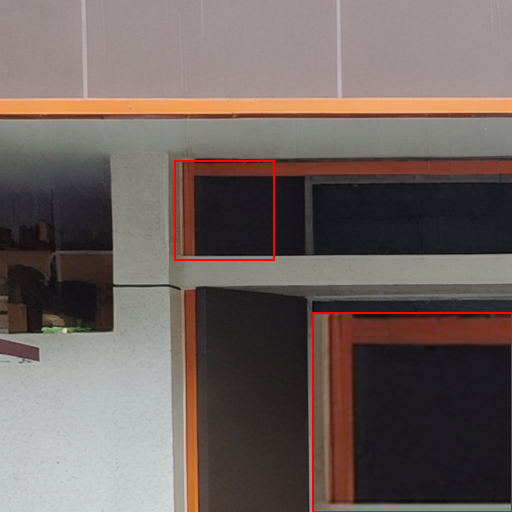}&\includegraphics[height=1.37cm, width=1.88cm]{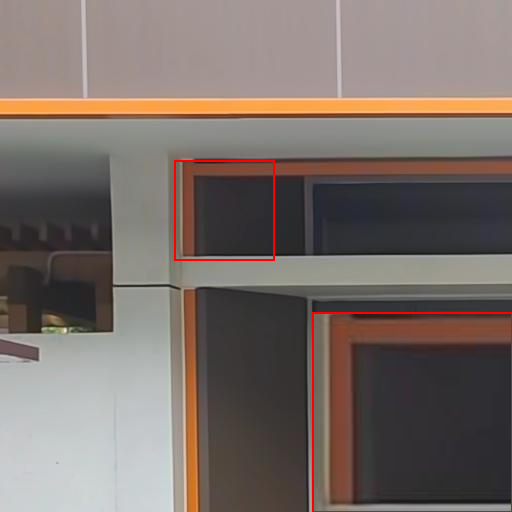}\\

         \rotatebox[origin=c]{90}{REVIDE\hspace{-33.5pt}}&\includegraphics[height=1.37cm, width=1.88cm]{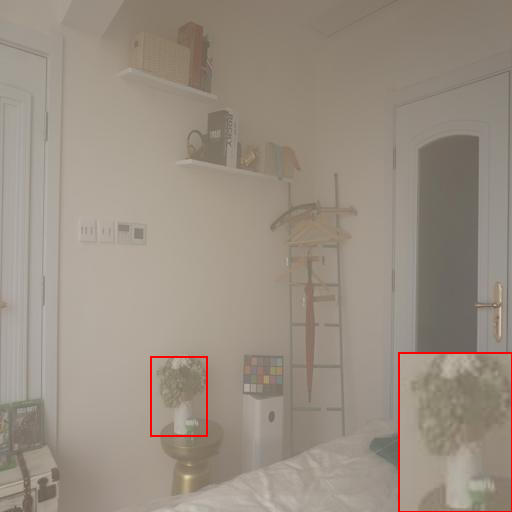}&\includegraphics[height=1.37cm, width=1.88cm]{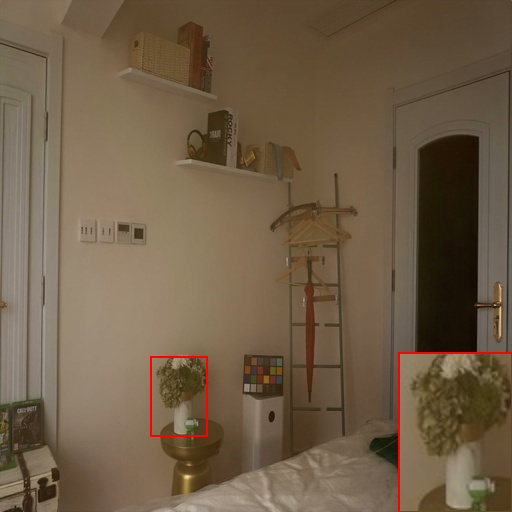}&\includegraphics[height=1.37cm, width=1.88cm]{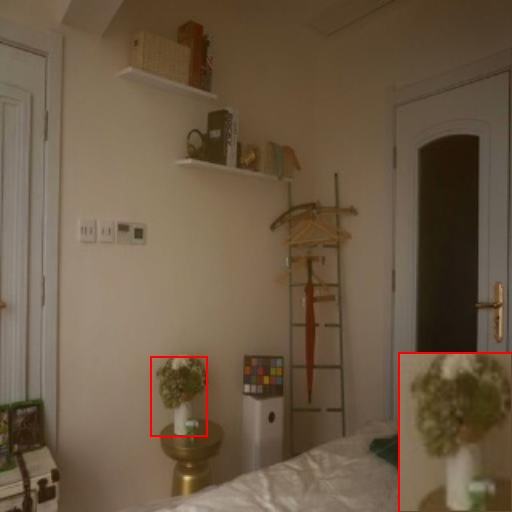}&\includegraphics[height=1.37cm, width=1.88cm]{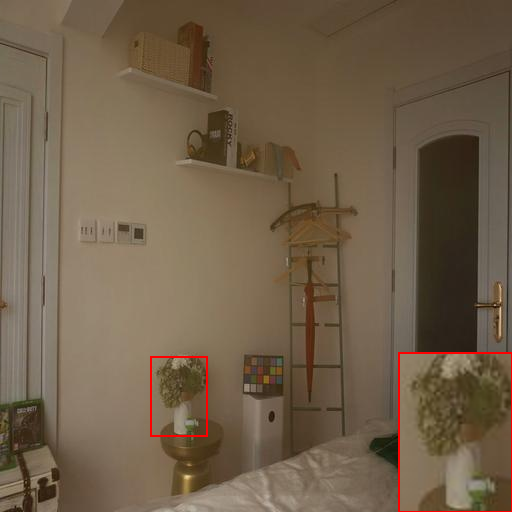}&\includegraphics[height=1.37cm, width=1.88cm]{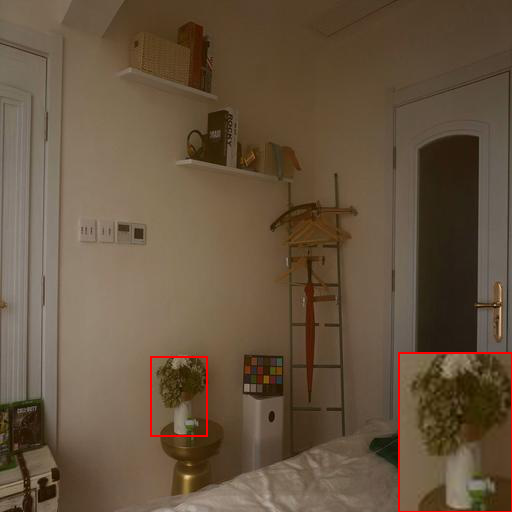}
         &\includegraphics[height=1.37cm, width=1.88cm]{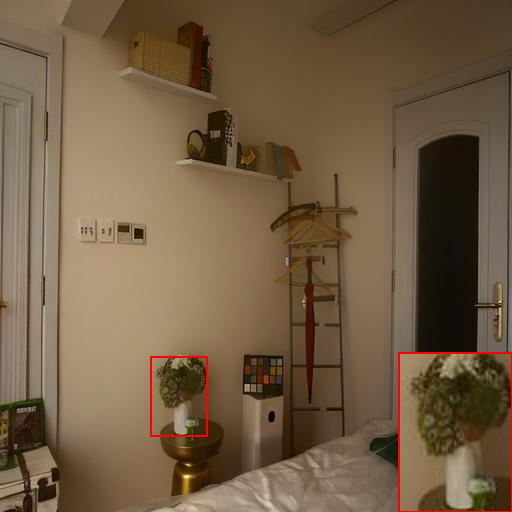}&\includegraphics[height=1.37cm, width=1.88cm]{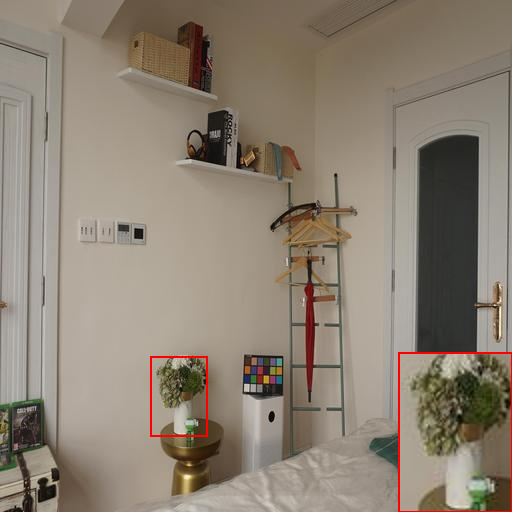}\\


         \rotatebox[origin=c]{90}{TOLED\hspace{-33.5pt}}&\includegraphics[height=1.37cm, width=1.88cm]{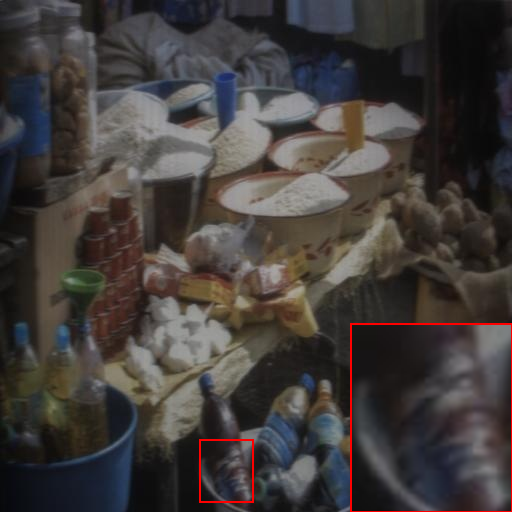}&\includegraphics[height=1.37cm, width=1.88cm]{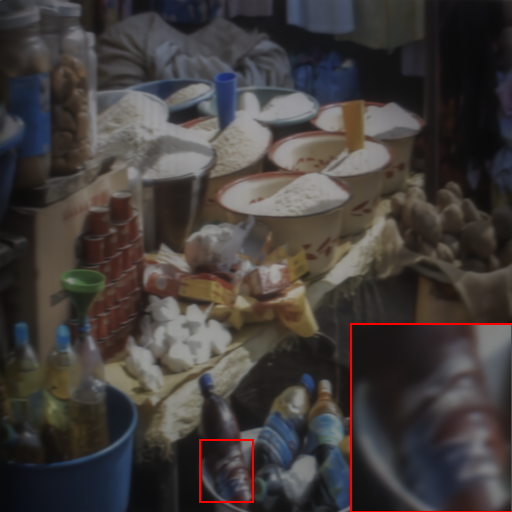}&\includegraphics[height=1.37cm, width=1.88cm]{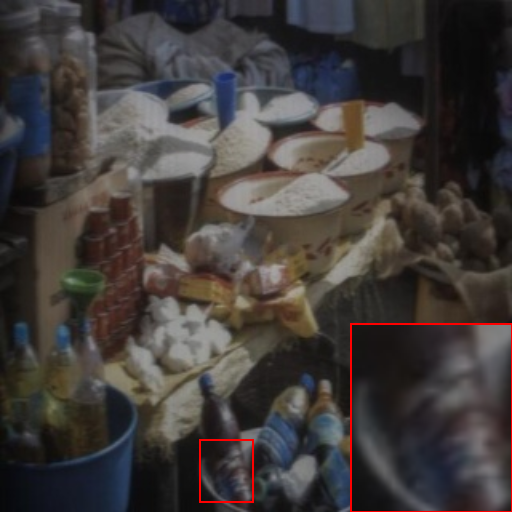}&\includegraphics[height=1.37cm, width=1.88cm]{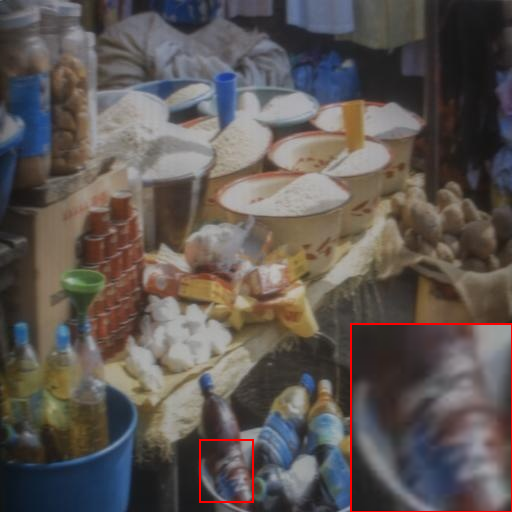}&\includegraphics[height=1.37cm, width=1.88cm]{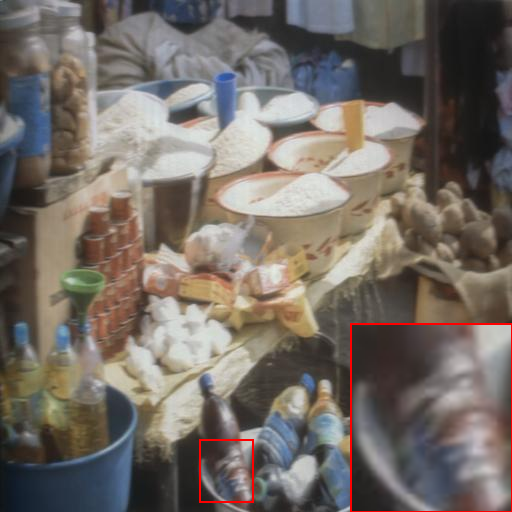}
         &\includegraphics[height=1.37cm, width=1.88cm]{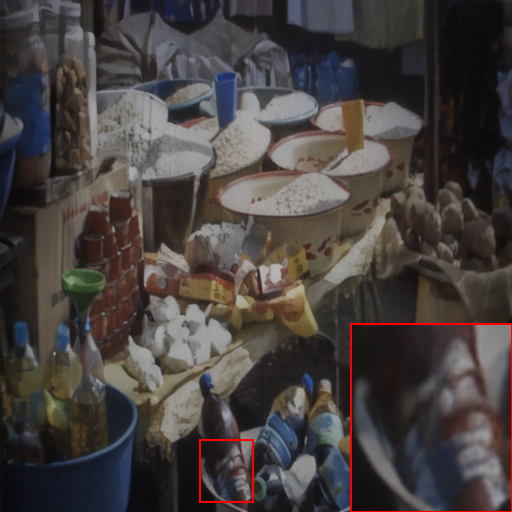}&\includegraphics[height=1.37cm, width=1.88cm]{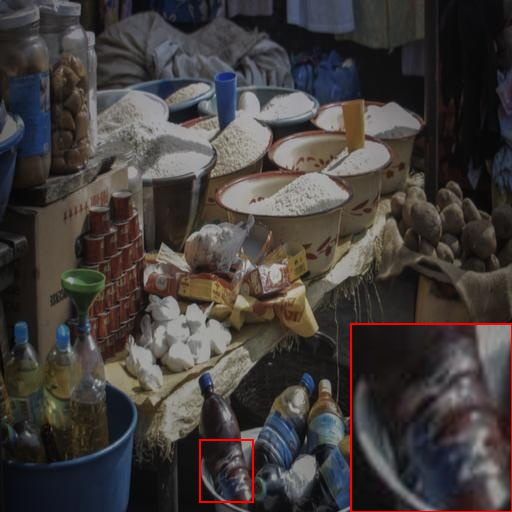}\\

         \rotatebox[origin=c]{90}{POLED\hspace{-33.5pt}}&\includegraphics[height=1.37cm, width=1.88cm]{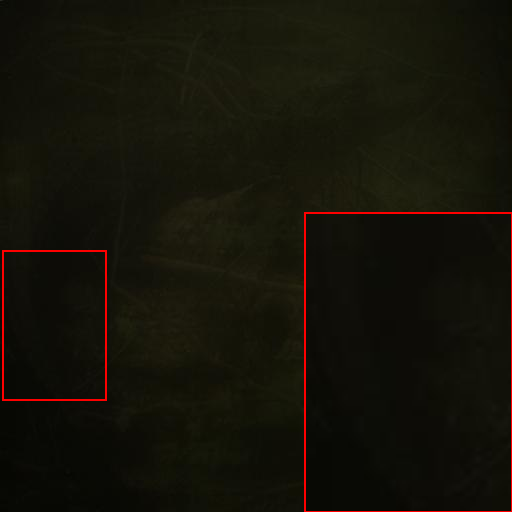}&\includegraphics[height=1.37cm, width=1.88cm]{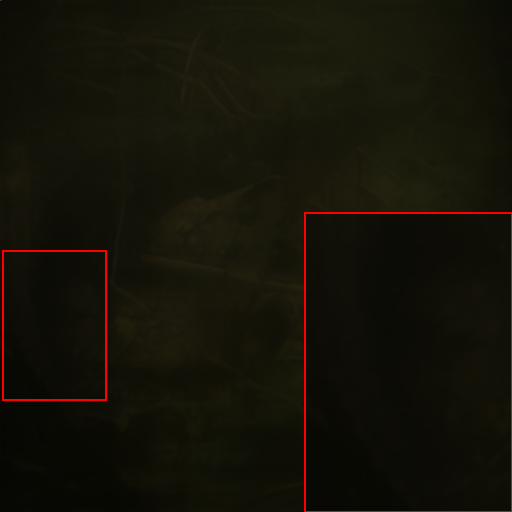}&\includegraphics[height=1.37cm, width=1.88cm]{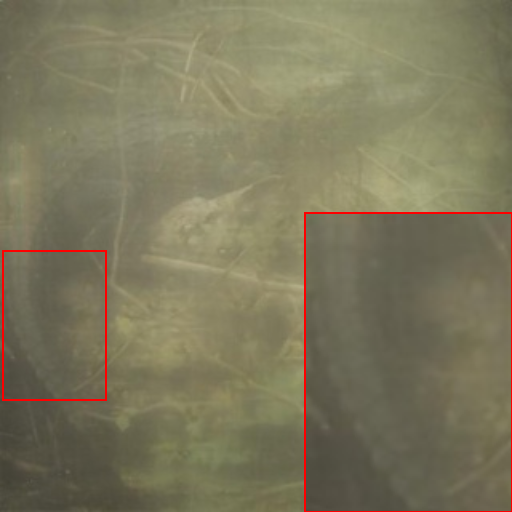}&\includegraphics[height=1.37cm, width=1.88cm]{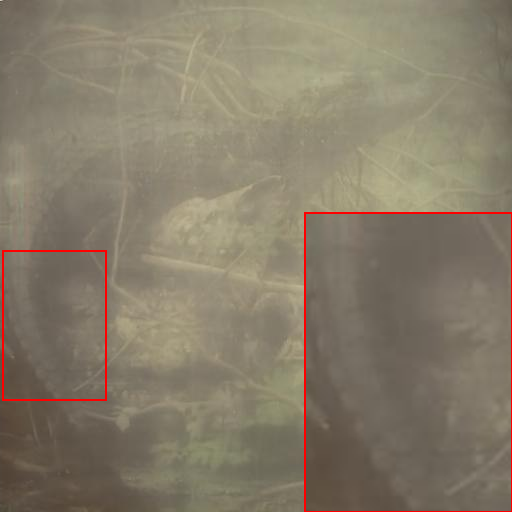}&\includegraphics[height=1.37cm, width=1.88cm]{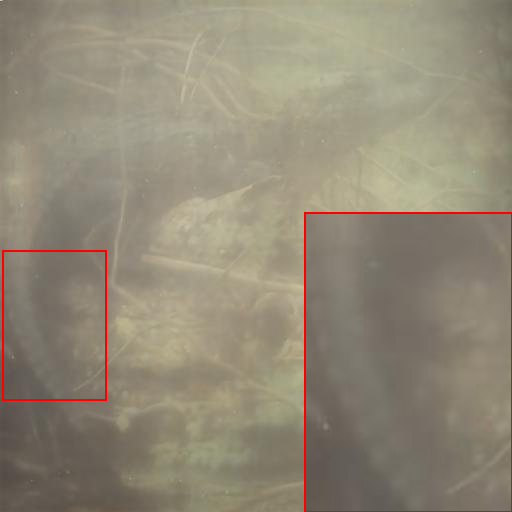}
         &\includegraphics[height=1.37cm, width=1.88cm]{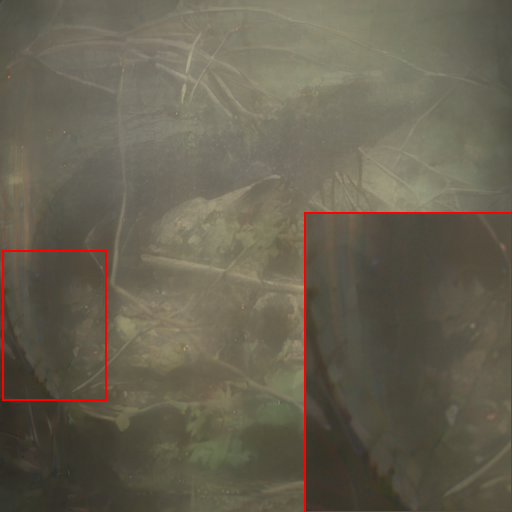}&\includegraphics[height=1.37cm, width=1.88cm]{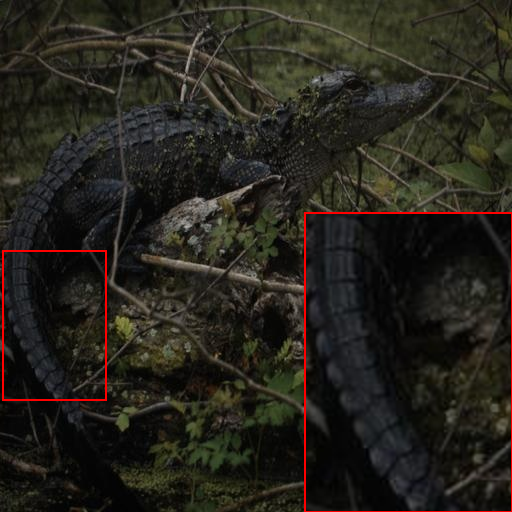}\\

        \rotatebox[origin=c]{90}{CDD\hspace{-33pt}}&\includegraphics[height=1.37cm, width=1.88cm]{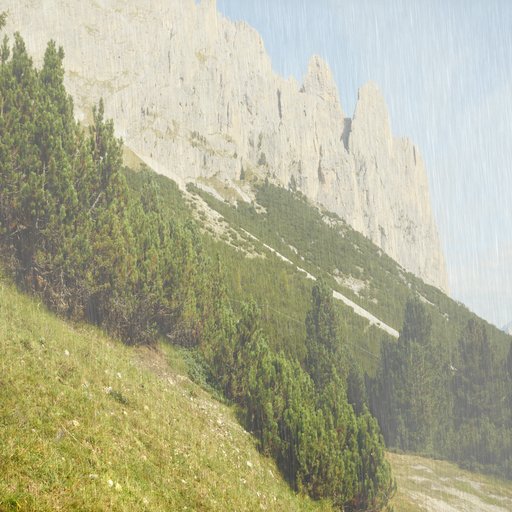}&\includegraphics[height=1.37cm, width=1.88cm]{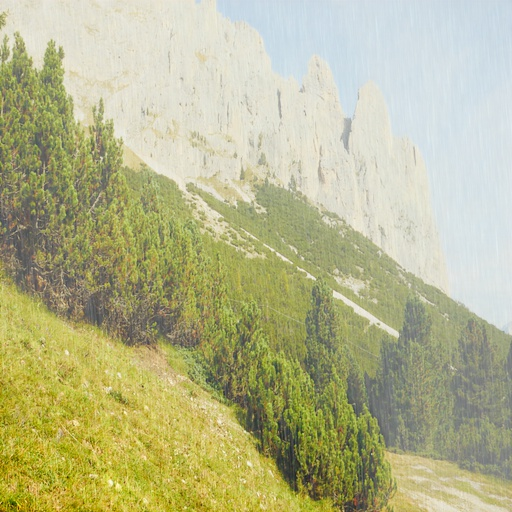}&\includegraphics[height=1.37cm, width=1.88cm]{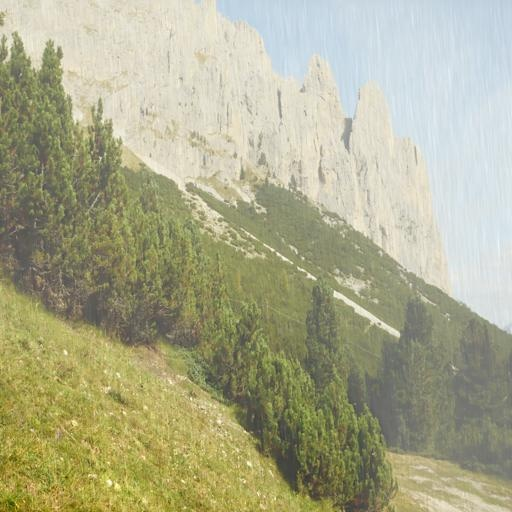}&\includegraphics[height=1.37cm, width=1.88cm]{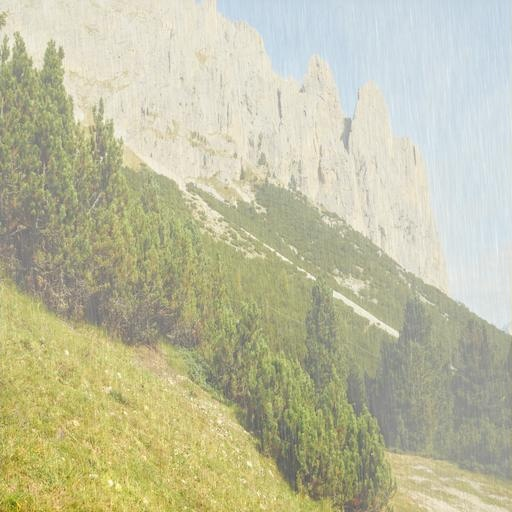}&\includegraphics[height=1.37cm, width=1.88cm]{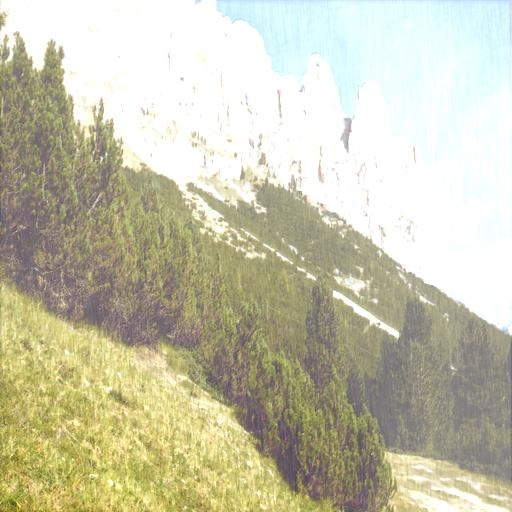}&\includegraphics[height=1.37cm, width=1.88cm]{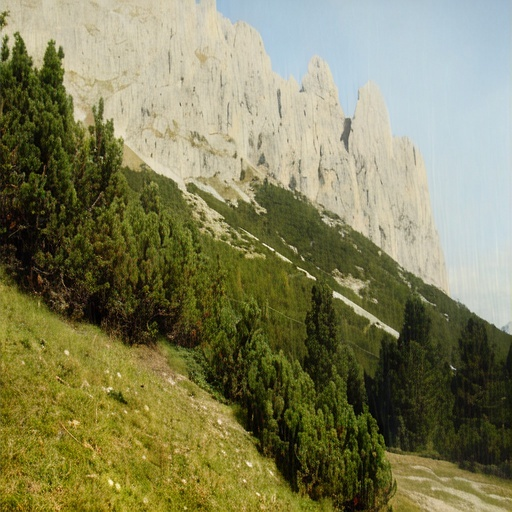}&\includegraphics[height=1.37cm, width=1.88cm]{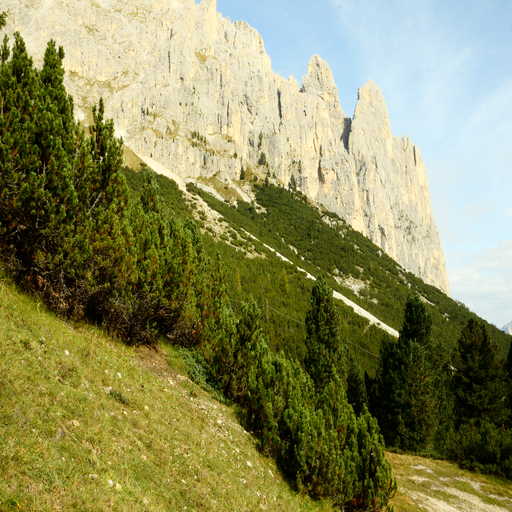}\\

         
    \end{tabular}
    \vspace{-10pt}
    \caption{Qualitative comparisons of RestoreVAR with non-generative methods on real, unseen and mixed degradations. RestoreVAR consistently achieves better results.}
    \label{fig: ood_qual}
    \vspace{-15pt}
\end{figure*}

 \begin{wraptable}{r}{0.21\textwidth}
  \vspace{-
  15pt}                          
  \centering
  \scriptsize
  \renewcommand{\arraystretch}{0.9}
  \setlength{\tabcolsep}{3pt}
  \caption{Mean scores from user study.}
  \vspace{-3pt}
  \rowcolors{2}{gray!15}{white}
  \begin{tabular}{l c}
    \toprule
    \textbf{Method} & \textbf{Score}\,$\uparrow$ \\
    \midrule
    PromptIR     & 2.11 \\
    InstructIR & 2.93 \\
    AWRaCLe      & 2.33 \\
    DCPT             & 2.42 \\
    DFPIR & 2.35\\
    AutoDIR       & 3.68 \\
    \rowcolor{gray!25}
    \textbf{RestoreVAR}          & \textbf{4.36} \\
    \bottomrule
  \end{tabular}
  \label{tab:user_study}
  \vspace{-10pt}
\end{wraptable}
For testing generalization, we report MUSIQ~\citep{musiq} and CLIPIQA~\citep{clipiqa} scores in Table~\ref{tab:generalization_metrics} on the real-world, unseen and mixed degradation datasets discussed in Sec.~\ref{subsec: datasets}. RestoreVAR achieves higher scores than non-generative models (on average), indicating better robustness under these degradations. Qualitative results for this experiment are shown in Fig.~\ref{fig: ood_qual}, where RestoreVAR consistently outperforms non-generative approaches. Due to space constraints, qualitative comparisons with PromptIR and visual results for LOLBlur are given in the supplementary. Furthermore, RestoreVAR achieves performance competitive to those of LDM-based methods on these datasets. The slightly lower metrics for RestoreVAR can be explained by the fact that the latent diffusion backbones (Stable Diffusion~\citep{stablediff}) used by the competing generative models are trained on substantially larger datasets, typically hundreds of millions of images, whereas the VAR backbone is trained on only $\sim 1$M ImageNet-1K~\citep{imagenet} images. Despite this, RestoreVAR demonstrates comparable generalization performance while offering much faster inference and improved pixel-level fidelity. To further evaluate perceptual quality, we conducted a user study in which participants rated outputs from non-generative models, AutoDIR (LDM-based)  and RestoreVAR, for $50$ real-world scenes. We received $36$ responses with each participant scoring outputs  based on scene consistency, restoration quality, and overall appeal on a 5-point scale. Table~\ref{tab:user_study} shows that RestoreVAR received the highest average ratings (across all three criteria), highlighting its ability to produce images that align closely with human preferences. More quantitative results are provided in the supplementary.

To summarize, although non-generative AiOR models offer fast inference and good pixel-level fidelity, their generalization to real-world, unseen, and mixed degradations is limited. They work well only when test samples come from a distribution close to their training data, which hinders real-world applicability. In contrast, LDM-based generative approaches exhibit substantially stronger perceptual quality and generalization, but they are computationally expensive and suffer from low pixel-level fidelity.
RestoreVAR addresses the limitations of LDM-based approaches while retaining their strengths. It achieves far superior pixel-level fidelity as established by the multi-task restoration performance on datasets in Table~\ref{tab:within_dist}. Furthermore, on the real-world and unseen degradation datasets (Table~\ref{tab:generalization_metrics}), RestoreVAR achieves perceptual quality on par with diffusion-based methods while significantly outperforming non-generative models, demonstrating strong generalization. Moreover, RestoreVAR is substantially faster than LDM-based approaches and even nears the efficiency of some non-generative methods (see supplementary for details), despite requiring multiple steps for inference. Together, these results demonstrate that RestoreVAR closes the fidelity gap between LDM-based generative models and non-generative approaches while retaining strong perceptual quality and generalization, and offering enhanced computational efficiency.

We discuss limitations of RestoreVAR in the supplementary.

\subsection{Ablations and Analysis}
\label{subsec: ablations}

\begin{wraptable}{t}{0.45\textwidth}
    \centering
    \vspace{-13pt}
    \small
    \setlength{\tabcolsep}{2pt}
    \caption{Ablations on the types of latent refiners. Our proposed latent refiner transformer (LRT) performs best, with minimal overhead.}
    \vspace{-6pt}
    \resizebox{0.45\textwidth}{!}{
    \begin{tabular}{l c c c}
      \toprule
      \textbf{Refiner Type} & \textbf{Time (s)} & \textbf{Params (M)} & \textbf{PSNR / SSIM} \\
      \midrule
      No Refiner            & –      & –     & 21.71 / 0.690 \\
      HART Refiner          & 0.0455 & 36.06 & 23.48 / 0.777 \\
      LRT w/o Last-Block    & 0.0036 & 14.61 & 21.23 / 0.660 \\
      Proposed LRT          & 0.0061 & 22.97 & 24.67 / 0.821 \\
      \bottomrule
      \vspace{-10pt}
    \end{tabular}}
    \label{tab:refiner_ablation}
\end{wraptable}

\begin{figure}[t]
  \centering
  \begin{minipage}{0.46\linewidth}
    \centering
    \includegraphics[height=0.52\linewidth, width=\linewidth]{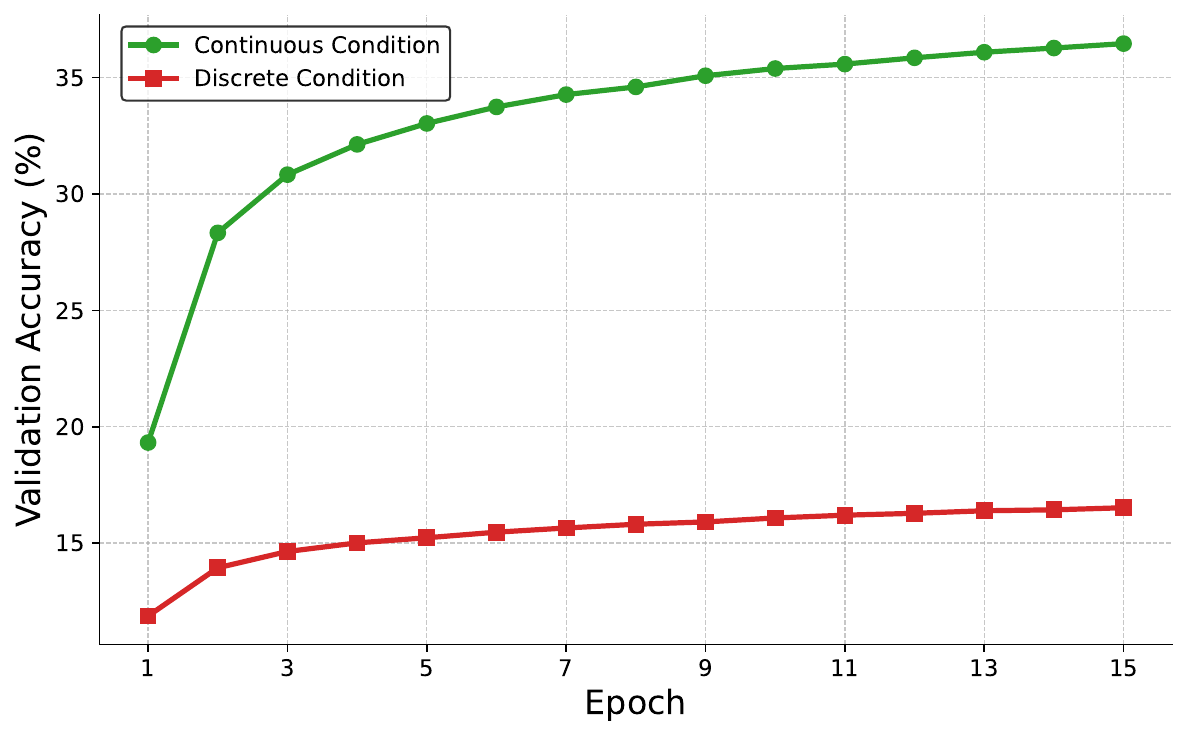}
    \vspace{-20pt}
    \caption{Validation accuracy of RestoreVAR under discrete vs. continuous conditioning.}
    \label{fig:contvsdisc}
  \end{minipage}\hfill
  \begin{minipage}{0.5\linewidth}
    \centering
    \begin{subfigure}[t]{0.32\linewidth}
      \includegraphics[width=\linewidth]{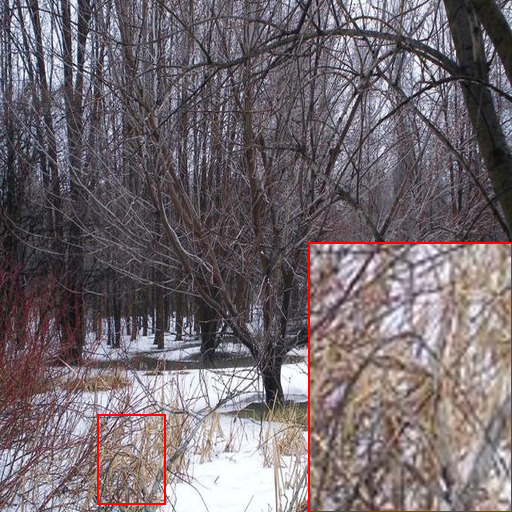}
      \caption*{Input image}
    \end{subfigure}\hfill
    \begin{subfigure}[t]{0.32\linewidth}
      \includegraphics[width=\linewidth]{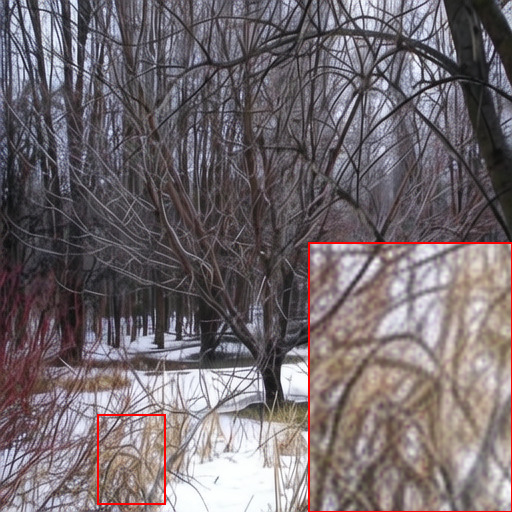}
      \caption*{w/o Disc}
    \end{subfigure}\hfill
    \begin{subfigure}[t]{0.32\linewidth}
      \includegraphics[width=\linewidth]{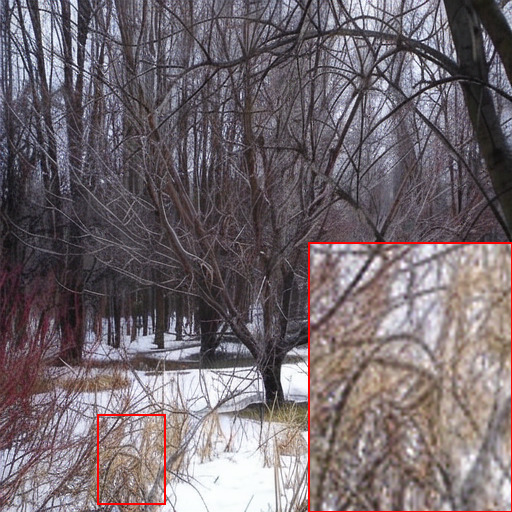}
      \caption*{w Disc}
    \end{subfigure}
    \vspace{-5pt}
    \caption{Image reconstructed by VAE decoders fine-tuned on continuous latents with (w) and without (w/o) a discriminator (Disc).}
    \label{fig:discriminator_comp}
  \end{minipage}
  \vspace{-12pt}
\end{figure}

\textbf{Continuous vs. Discrete Conditioning. }The RestoreVAR transformer conditions on the continuous latent of the degraded image ($f_{\text{cont}}^{\text{deg}}$). While conditioning with discrete multi-scale latents appears more aligned with VAR’s multi-scale prediction objective, it results in significantly worse performance. To demonstrate this, we train RestoreVAR with discrete and continuous conditioning for $15$ epochs each. As shown in Fig.~\ref{fig:contvsdisc}, RestoreVAR with discrete conditioning exhibits much lower validation accuracy. 




\textbf{Discriminator for VAE fine-tuning. }As described in Sec.~\ref{subssubsec: lrm}, we fine-tune the VAE decoder on continuous latents using a combination of pixel-level loss and an adversarial loss. To analyze the impact of the discriminator, we compare the reconstructions of VAE decoders fine-tuned with and without the adversarial loss. As shown in Fig.~\ref{fig:discriminator_comp}, removing the discriminator leads to blurrier reconstruction while including it yields sharper and perceptually better looking outputs. Ablations on SSIM and perceptual losses are provided in the supplementary.

\textbf{Latent Refiner Transformer. }The Latent Refiner Transformer (LRT) is critical for preserving pixel-level detail in restored images. To analyze its impact, we compare four RestoreVAR variants: (i) No refiner, (ii) HART's diffusion refiner, (iii) LRT without final block outputs, and (iv) our proposed LRT. As shown in Table~\ref{tab:refiner_ablation}, our LRT achieves the best PSNR and SSIM, while maintaining low inference time and a low parameter count. Using no refiner yields poor PSNR/SSIM scores. \begin{wrapfigure}{r}{0.526\textwidth}
    \vspace{-10pt}
    \centering
    \small
    \setlength{\tabcolsep}{1pt}
    \begin{tabular}{ccccc}
    &Degraded&Res.$+$coarse&Res.$+$fine&Res.\\
        \rotatebox[origin=c]{90}{Haze\hspace{-26pt}}&\includegraphics[height=1.17cm, width=1.68cm]{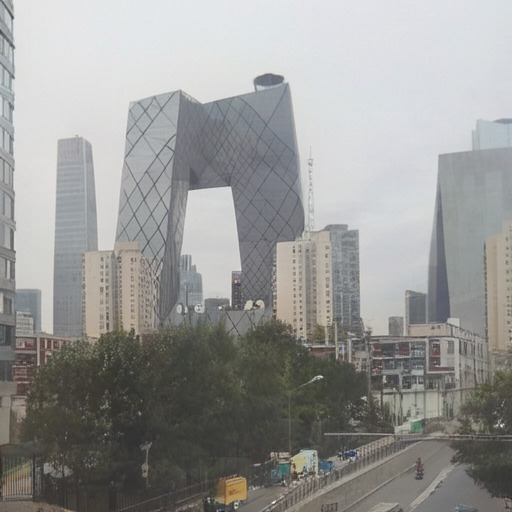}&\includegraphics[height=1.17cm, width=1.68cm]{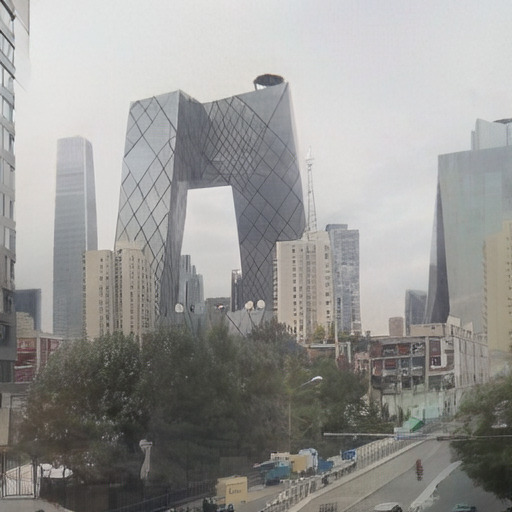}&\includegraphics[height=1.17cm, width=1.68cm]{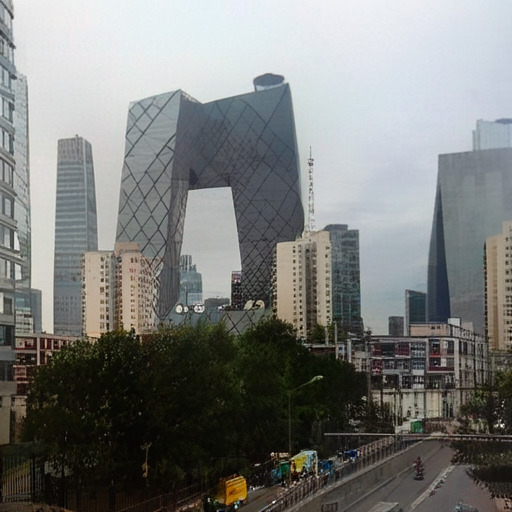}&\includegraphics[height=1.17cm, width=1.68cm]{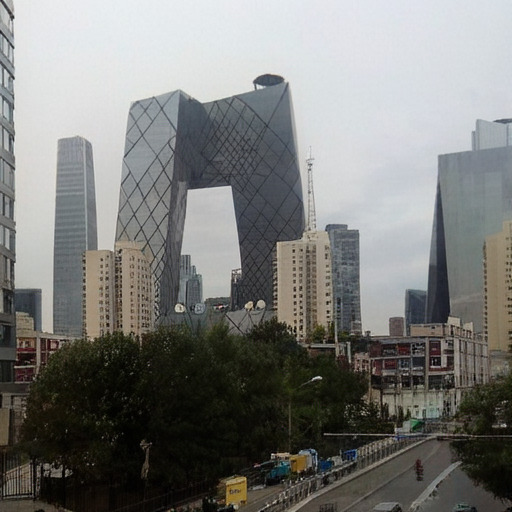}\\

        \rotatebox[origin=c]{90}{Snow\hspace{-26pt}}&\includegraphics[height=1.17cm, width=1.68cm]{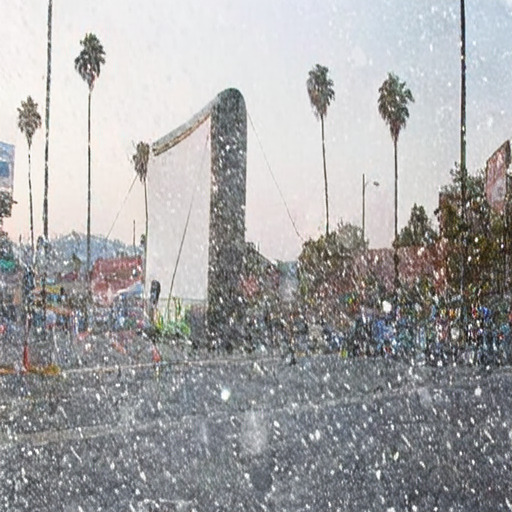}&\includegraphics[height=1.17cm, width=1.68cm]{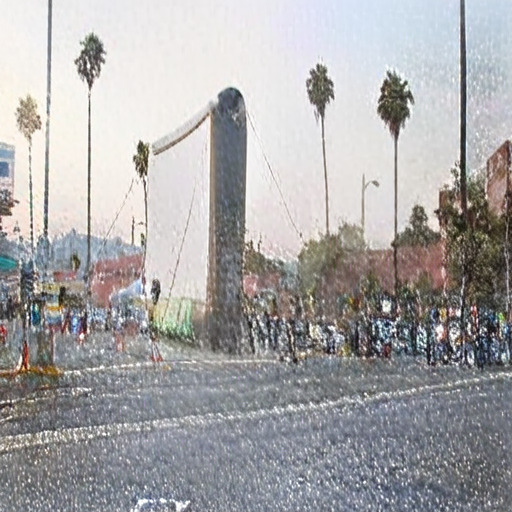}&\includegraphics[height=1.17cm, width=1.68cm]{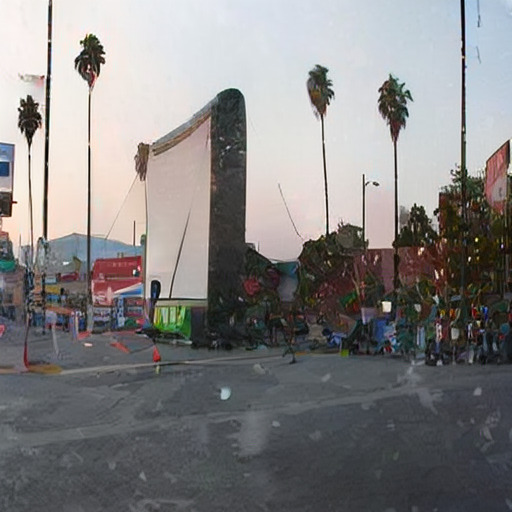}&\includegraphics[height=1.17cm, width=1.68cm]{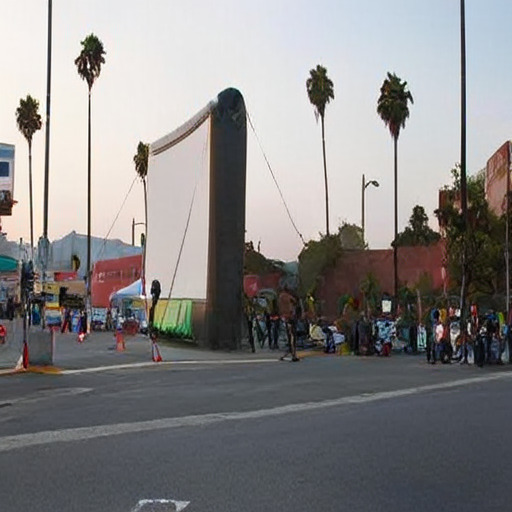}\\

        \rotatebox[origin=c]{90}{Rain\hspace{-26pt}}&\includegraphics[height=1.17cm, width=1.68cm]{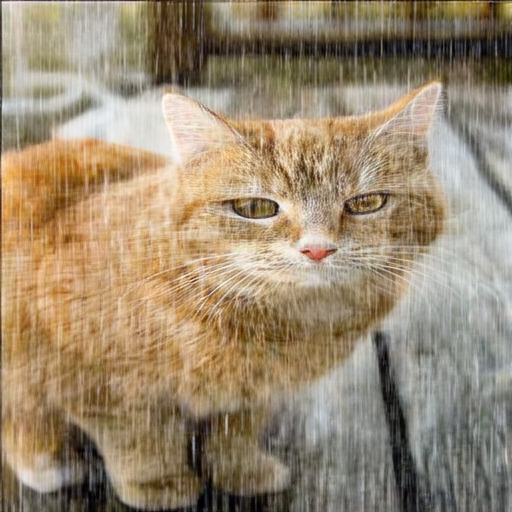}&\includegraphics[height=1.17cm, width=1.68cm]{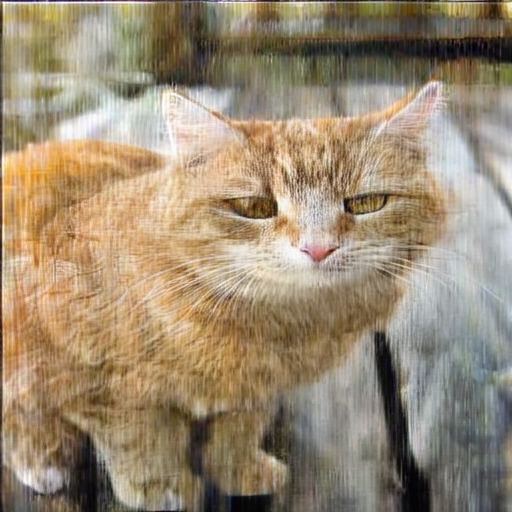}&\includegraphics[height=1.17cm, width=1.68cm]{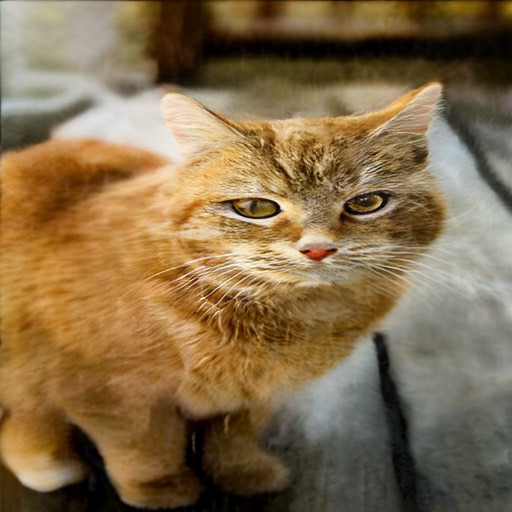}&\includegraphics[height=1.17cm, width=1.68cm]{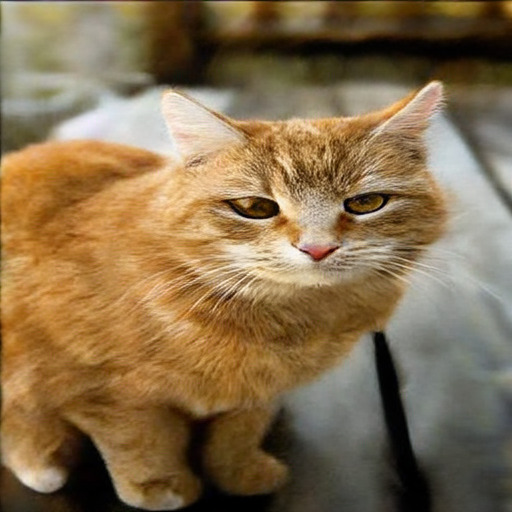}\\

        \rotatebox[origin=c]{90}{Low-light\hspace{-30pt}}&\includegraphics[height=1.17cm, width=1.68cm]{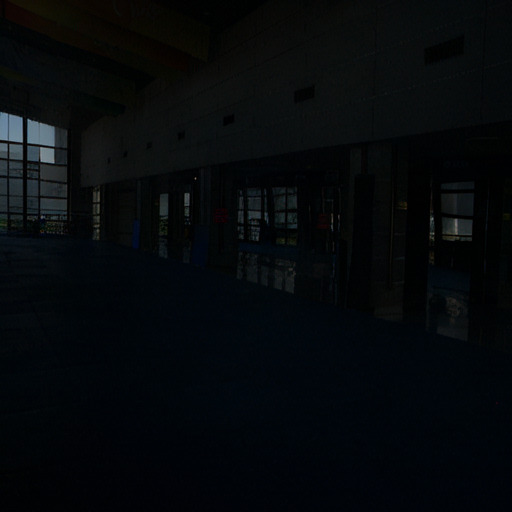}&\includegraphics[height=1.17cm, width=1.68cm]{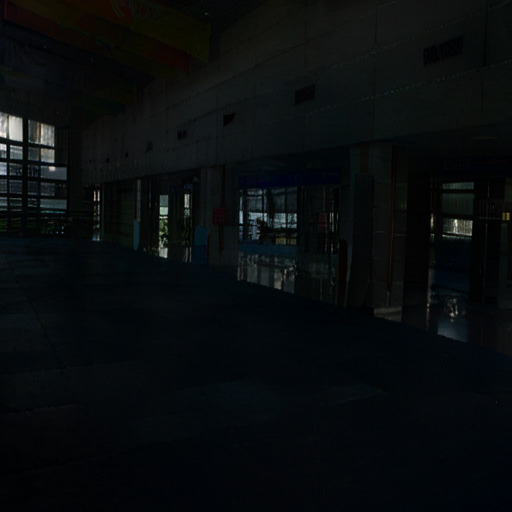}&\includegraphics[height=1.17cm, width=1.68cm]{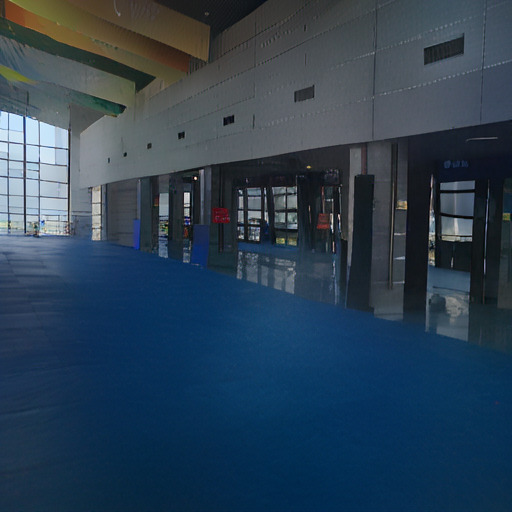}&\includegraphics[height=1.17cm, width=1.68cm]{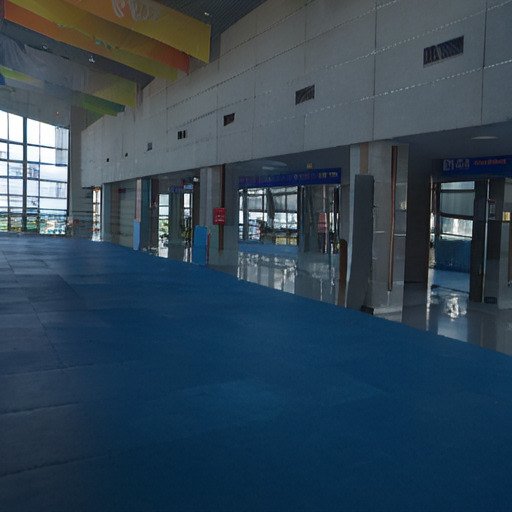}\\

        \rotatebox[origin=c]{90}{Blur\hspace{-26pt}}&\includegraphics[height=1.17cm, width=1.68cm]{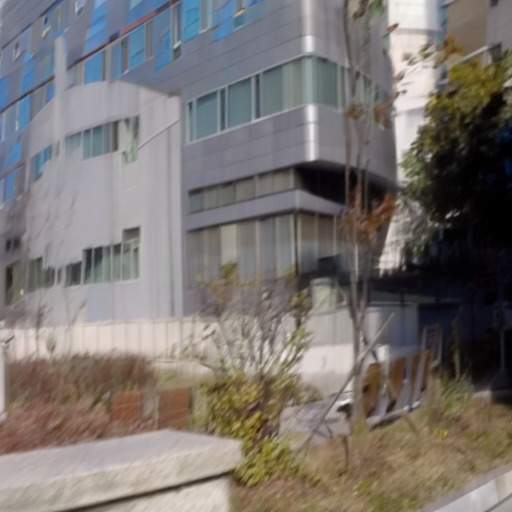}&\includegraphics[height=1.17cm, width=1.68cm]{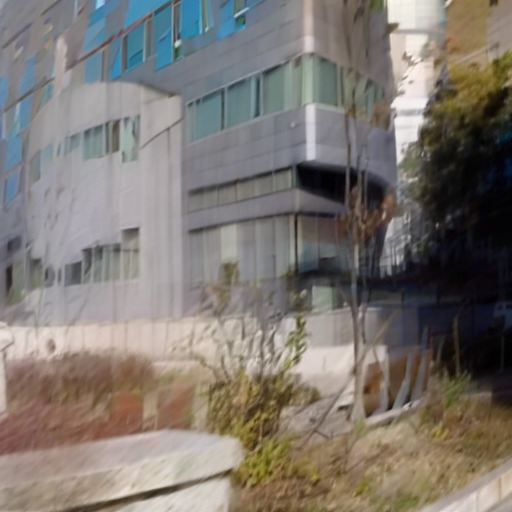}&\includegraphics[height=1.17cm, width=1.68cm]{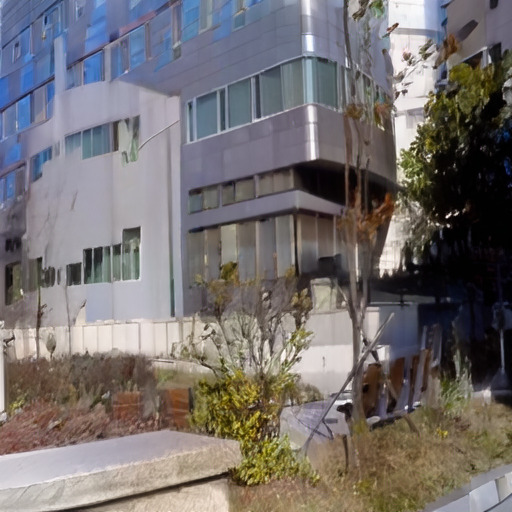}&\includegraphics[height=1.17cm, width=1.68cm]{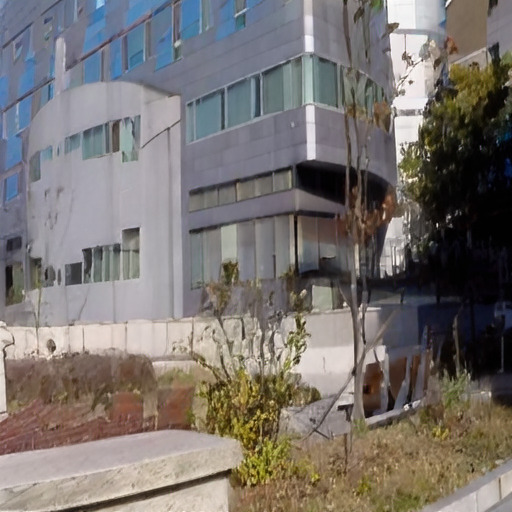}\\

    \end{tabular}
    \vspace{-10pt}
    \caption{Scale-space analysis of outputs from RestoreVAR. Res. denotes the Restored output. Res.$+$coarse replaces early scales of restored output with degraded ones, while Res.$+$fine replaces the late scales. It can be observed that the coarse scales contribute most to overall restoration.}
    \label{fig: scale_space_var}
    \vspace{-15pt}
\end{wrapfigure}Removing the last block outputs significantly reduces performance, indicating its importance as pseudo-continuous guidance for refinement. HART's MLP diffusion-based refiner performs worse than our LRT while having a much higher parameter count and runs $\sim\mathbf{7\times}$ slower. To assess if the LRT causes loss of perceptual quality, we calculated MUSIQ and CLIPIQA scores (higher is better) with and without the LRT. Using no LRT yields a MUSIQ/CLIPIQA score of $66.56/0.472$ while incorporating the LRT results in a score of $66.45/0.481$. The scores are nearly identical indicating negligible impact on perceptual quality.

\textbf{Scale-space analysis of outputs. }A core motivation behind our approach is that the scale-space residual quantization of VAR captures degradations at coarse scales and scene level details at finer scales (see Sec.~\ref{subsec: var_motivation}). Fig. ~\ref{fig: scale_space_var} visualizes this for the outputs of RestoreVAR by comparing (i) the degraded image, (ii) the restored image with its coarse scales replaced by those of the degraded input, (iii) the restored image with only fine scales replaced, and (iv) the restored image. Similar to our observations in Sec.~\ref{subsec: var_motivation}, replacing the coarse-scale components of the restored output reintroduces the degradation, whereas replacing only the finer scales does not. This once again demonstrates that degradations predominantly reside in the coarse-scale indices, and that RestoreVAR predicts coarse-scale tokens that are mostly free of degradation while reconstructing scene-level details at the finer scales.

More ablations are provided in the supplementary.

\section{Conclusions}
We proposed RestoreVAR, a fast and effective generative approach for AiOR. Built on the VAR backbone, RestoreVAR benefits from VAR's strong generative priors and significantly faster inference compared to LDMs. To tailor VAR for AiOR, we introduced cross-attention mechanisms that inject semantic information from the degraded image into the generation process. Additionally, we proposed a non-generative latent refiner transformer to convert discrete latents to continuous ones, along with a VAE decoder fine-tuned on continuous latents, which together improve reconstruction fidelity. RestoreVAR achieves state-of-the-art performance among generative AiOR models, outperforming LDM-based methods while delivering over $10\times$ faster inference and strong generalization.

\section*{Acknowledgments}
This work is supported by the Intelligence Advanced Research Projects Activity (IARPA) via Department of Interior/ Interior Business Center (DOI/IBC) contract number 140D0423C0076. The U.S. Government is authorized to reproduce and distribute reprints for Governmental purposes notwithstanding any copyright annotation thereon. Disclaimer: The views and conclusions contained herein are those of the authors and should not be interpreted as necessarily representing the official policies or endorsements, either expressed or implied, of IARPA, DOI/IBC, or the U.S. Government.

\section*{Ethics Statement}
We acknowledge that we have read and adhered to the ICLR Code of Ethics. For all our experiments, we used publicly available open-source datasets.

\section*{Reproducibility Statement}
Our code will be made publicly available after the review process. Details to reproduce the work have been provided in Secs.~\ref{sec: proposed} and ~\ref{subsec: impl} of the main paper and Sec.~\ref{supsec: arch_details} of the supplementary.

\bibliography{iclr2026_conference}

@String(CVPR= {IEEE Conf. Comput. Vis. Pattern Recog.})

@String(ICCV= {Int. Conf. Comput. Vis.})

@String(AAAI = {AAAI})

@String(CVPR  = {CVPR})

@String(ICCV  = {ICCV})

@inproceedings{gopro,
  title={Deep multi-scale convolutional neural network for dynamic scene deblurring},
  author={Nah, Seungjun and Hyun Kim, Tae and Mu Lee, Kyoung},
  booktitle={Proceedings of the IEEE conference on computer vision and pattern recognition},
  pages={3883--3891},
  year={2017}
}

@article{daclip,
  title={Controlling vision-language models for universal image restoration},
  author={Luo, Ziwei and Gustafsson, Fredrik K and Zhao, Zheng and Sj{\"o}lund, Jens and Sch{\"o}n, Thomas B},
  journal={arXiv preprint arXiv:2310.01018},
  volume={3},
  number={8},
  year={2023}
}

@article{lolv1,
  title={Deep retinex decomposition for low-light enhancement},
  author={Wei, Chen and Wang, Wenjing and Yang, Wenhan and Liu, Jiaying},
  journal={arXiv preprint arXiv:1808.04560},
  year={2018}
}

@inproceedings{sid,
  title={Learning to see in the dark},
  author={Chen, Chen and Chen, Qifeng and Xu, Jia and Koltun, Vladlen},
  booktitle={Proceedings of the IEEE conference on computer vision and pattern recognition},
  pages={3291--3300},
  year={2018}
}

@inproceedings{musiq,
  title={Musiq: Multi-scale image quality transformer},
  author={Ke, Junjie and Wang, Qifei and Wang, Yilin and Milanfar, Peyman and Yang, Feng},
  booktitle={Proceedings of the IEEE/CVF international conference on computer vision},
  pages={5148--5157},
  year={2021}
}

@inproceedings{clipiqa,
  title={Exploring clip for assessing the look and feel of images},
  author={Wang, Jianyi and Chan, Kelvin CK and Loy, Chen Change},
  booktitle={Proceedings of the AAAI conference on artificial intelligence},
  volume={37},
  number={2},
  pages={2555--2563},
  year={2023}
}

@article{gpt2,
  title={Language models are unsupervised multitask learners},
  author={Radford, Alec and Wu, Jeffrey and Child, Rewon and Luan, David and Amodei, Dario and Sutskever, Ilya and others},
  journal={OpenAI blog},
  volume={1},
  number={8},
  pages={9},
  year={2019}
}

@inproceedings{toledpoled,
  title={Image restoration for under-display camera},
  author={Zhou, Yuqian and Ren, David and Emerton, Neil and Lim, Sehoon and Large, Timothy},
  booktitle={Proceedings of the ieee/cvf conference on computer vision and pattern recognition},
  pages={9179--9188},
  year={2021}
}

@article{adamw,
  title={Decoupled weight decay regularization},
  author={Loshchilov, Ilya and Hutter, Frank},
  journal={arXiv preprint arXiv:1711.05101},
  year={2017}
}

@inproceedings{instructir,
  title={InstructIR: High-Quality Image Restoration Following Human Instructions},
  author={Conde, Marcos V and Geigle, Gregor and Timofte, Radu},
  booktitle={European Conference on Computer Vision},
  pages={1--21},
  year={2025},
  organization={Springer}
}

@article{autodir,
  title={Autodir: Automatic all-in-one image restoration with latent diffusion},
  author={Jiang, Yitong and Zhang, Zhaoyang and Xue, Tianfan and Gu, Jinwei},
  journal={arXiv preprint arXiv:2310.10123},
  year={2023}
}

@INPROCEEDINGS{early1,
  author={Kaiming He and Jian Sun and Xiaoou Tang},
  booktitle={2009 IEEE Conference on Computer Vision and Pattern Recognition}, 
  title={Single image haze removal using dark channel prior}, 
  year={2009},
  volume={},
  number={},
  pages={1956-1963},
  doi={10.1109/CVPR.2009.5206515}}

@InProceedings{lhprain,
    author    = {Guo, Yun and Xiao, Xueyao and Chang, Yi and Deng, Shumin and Yan, Luxin},
    title     = {From Sky to the Ground: A Large-scale Benchmark and Simple Baseline Towards Real Rain Removal},
    booktitle = {Proceedings of the IEEE/CVF International Conference on Computer Vision (ICCV)},
    month     = {October},
    year      = {2023},
    pages     = {12097-12107}
}

@INPROCEEDINGS{rain1,
  author={Wang, Tianyu and Yang, Xin and Xu, Ke and Chen, Shaozhe and Zhang, Qiang and Lau, Rynson W.H.},
  booktitle={2019 IEEE/CVF Conference on Computer Vision and Pattern Recognition (CVPR)}, 
  title={Spatial Attentive Single-Image Deraining With a High Quality Real Rain Dataset}, 
  year={2019},
  volume={},
  number={},
  pages={12262-12271},
  doi={10.1109/CVPR.2019.01255}}

@inproceedings{imagenet,
  title={Imagenet: A large-scale hierarchical image database},
  author={Deng, Jia and Dong, Wei and Socher, Richard and Li, Li-Jia and Li, Kai and Fei-Fei, Li},
  booktitle={2009 IEEE conference on computer vision and pattern recognition},
  pages={248--255},
  year={2009},
  organization={Ieee}
}

@article{vae,
  title={Auto-encoding variational bayes},
  author={Kingma, Diederik P},
  journal={arXiv preprint arXiv:1312.6114},
  year={2013}
}

@inproceedings{stablediff,
  title={High-resolution image synthesis with latent diffusion models},
  author={Rombach, Robin and Blattmann, Andreas and Lorenz, Dominik and Esser, Patrick and Ommer, Bj{\"o}rn},
  booktitle={Proceedings of the IEEE/CVF conference on computer vision and pattern recognition},
  pages={10684--10695},
  year={2022}
}

@ARTICLE{haze1,
  author={Zhang, He and Sindagi, Vishwanath and Patel, Vishal M.},
  journal={IEEE Transactions on Circuits and Systems for Video Technology}, 
  title={Joint Transmission Map Estimation and Dehazing Using Deep Networks}, 
  year={2020},
  volume={30},
  number={7},
  pages={1975-1986},
  doi={10.1109/TCSVT.2019.2912145}}

@ARTICLE{snow1,
  author={Zhang, Kaihao and Li, Rongqing and Yu, Yanjiang and Luo, Wenhan and Li, Changsheng},
  journal={IEEE Transactions on Image Processing}, 
  title={Deep Dense Multi-Scale Network for Snow Removal Using Semantic and Depth Priors}, 
  year={2021},
  volume={30},
  number={},
  pages={7419-7431},
  doi={10.1109/TIP.2021.3104166}}

@ARTICLE{snow100k,
  author={Liu, Yun-Fu and Jaw, Da-Wei and Huang, Shih-Chia and Hwang, Jenq-Neng},
  journal={IEEE Transactions on Image Processing}, 
  title={DesnowNet: Context-Aware Deep Network for Snow Removal}, 
  year={2018},
  volume={27},
  number={6},
  pages={3064-3073},
  doi={10.1109/TIP.2018.2806202}}

@inproceedings{mprnet,
    title={Multi-Stage Progressive Image Restoration},
    author={Syed Waqas Zamir and Aditya Arora and Salman Khan and Munawar Hayat
            and Fahad Shahbaz Khan and Ming-Hsuan Yang and Ling Shao},
    booktitle={CVPR},
    year={2021}
}

@inproceedings{restormer,
    title={Restormer: Efficient Transformer for High-Resolution Image Restoration}, 
    author={Syed Waqas Zamir and Aditya Arora and Salman Khan and Munawar Hayat 
            and Fahad Shahbaz Khan and Ming-Hsuan Yang},
    booktitle={CVPR},
    year={2022}
}

@INPROCEEDINGS{swinir,
  author={Liang, Jingyun and Cao, Jiezhang and Sun, Guolei and Zhang, Kai and Van Gool, Luc and Timofte, Radu},
  booktitle={2021 IEEE/CVF International Conference on Computer Vision Workshops (ICCVW)}, 
  title={SwinIR: Image Restoration Using Swin Transformer}, 
  year={2021},
  volume={},
  number={},
  pages={1833-1844},
  doi={10.1109/ICCVW54120.2021.00210}}

@INPROCEEDINGS{airnet,
  author={Li, Boyun and Liu, Xiao and Hu, Peng and Wu, Zhongqin and Lv, Jiancheng and Peng, Xi},
  booktitle={2022 IEEE/CVF Conference on Computer Vision and Pattern Recognition (CVPR)}, 
  title={All-In-One Image Restoration for Unknown Corruption}, 
  year={2022},
  volume={},
  number={},
  pages={17431-17441},
  doi={10.1109/CVPR52688.2022.01693}}

@INPROCEEDINGS{nas,
  author={Li, Ruoteng and Tan, Robby T. and Cheong, Loong-Fah},
  booktitle={2020 IEEE/CVF Conference on Computer Vision and Pattern Recognition (CVPR)}, 
  title={All in One Bad Weather Removal Using Architectural Search}, 
  year={2020},
  volume={},
  number={},
  pages={3172-3182},
  doi={10.1109/CVPR42600.2020.00324}}

@INPROCEEDINGS {transw,
author = {J. Jose Valanarasu and R. Yasarla and V. M. Patel},
booktitle = {2022 IEEE/CVF Conference on Computer Vision and Pattern Recognition (CVPR)},
title = {TransWeather: Transformer-based Restoration of Images Degraded by Adverse Weather Conditions},
year = {2022},
volume = {},
issn = {},
pages = {2343-2353}}

@article{promptir,
  title={PromptIR: Prompting for All-in-One Image Restoration},
  author={Potlapalli, Vaishnav and Zamir, Syed Waqas and Khan, Salman H and Shahbaz Khan, Fahad},
  journal={Advances in Neural Information Processing Systems},
  volume={36},
  year={2024}
}

@inproceedings{gan1rain,
  title={Unpaired deep image deraining using dual contrastive learning},
  author={Chen, Xiang and Pan, Jinshan and Jiang, Kui and Li, Yufeng and Huang, Yufeng and Kong, Caihua and Dai, Longgang and Fan, Zhentao},
  booktitle={Proceedings of the IEEE/CVF conference on computer vision and pattern recognition},
  pages={2017--2026},
  year={2022}
}

@inproceedings{gan3blur,
  title={Deblurgan: Blind motion deblurring using conditional adversarial networks},
  author={Kupyn, Orest and Budzan, Volodymyr and Mykhailych, Mykola and Mishkin, Dmytro and Matas, Ji{\v{r}}{\'\i}},
  booktitle={Proceedings of the IEEE conference on computer vision and pattern recognition},
  pages={8183--8192},
  year={2018}
}

@ARTICLE{reside,
  author={Li, Boyi and Ren, Wenqi and Fu, Dengpan and Tao, Dacheng and Feng, Dan and Zeng, Wenjun and Wang, Zhangyang},
  journal={IEEE Transactions on Image Processing}, 
  title={Benchmarking Single-Image Dehazing and Beyond}, 
  year={2019},
  volume={28},
  number={1},
  pages={492-505},
  doi={10.1109/TIP.2018.2867951}}

@inproceedings{revide,
	author={Zhang, Xinyi and Dong, Hang and Pan, Jinshan and Zhu, Chao and Tai, Ying and Wang, Chengjie and Li, Jilin and Huang, Feiyue and Wang, Fei},
	title= {Learning To Restore Hazy Video: A New Real-World Dataset and a New Method},
	booktitle = {CVPR},
	pages={9239--9248},
	year = {2021}
}

@inproceedings{diffplugin,
  title={Diff-Plugin: Revitalizing Details for Diffusion-based Low-level Tasks},
  author={Liu, Yuhao and Ke, Zhanghan and Liu, Fang and Zhao, Nanxuan and Lau, Rynson WH},
  booktitle={Proceedings of the IEEE/CVF Conference on Computer Vision and Pattern Recognition},
  pages={4197--4208},
  year={2024}
}

@inproceedings{diffuir,
  title={Selective Hourglass Mapping for Universal Image Restoration Based on Diffusion Model},
  author={Zheng, Dian and Wu, Xiao-Ming and Yang, Shuzhou and Zhang, Jian and Hu, Jian-Fang and Zheng, Wei-shi},
  booktitle={Proceedings of the IEEE/CVF Conference on Computer Vision and Pattern Recognition},
  year={2024}
}

@inproceedings{rainnew2,
  title={Uncertainty guided multi-scale residual learning-using a cycle spinning cnn for single image de-raining},
  author={Yasarla, Rajeev and Patel, Vishal M},
  booktitle={Proceedings of the IEEE/CVF conference on computer vision and pattern recognition},
  pages={8405--8414},
  year={2019}
}

@article{awracle,
  title={AWRaCLe: All-Weather Image Restoration using Visual In-Context Learning},
  author={Rajagopalan, Sudarshan and Patel, Vishal M},
  journal={arXiv preprint arXiv:2409.00263},
  year={2024}
}

@article{gan,
  title={Generative adversarial networks},
  author={Goodfellow, Ian and Pouget-Abadie, Jean and Mirza, Mehdi and Xu, Bing and Warde-Farley, David and Ozair, Sherjil and Courville, Aaron and Bengio, Yoshua},
  journal={Communications of the ACM},
  volume={63},
  number={11},
  pages={139--144},
  year={2020},
  publisher={ACM New York, NY, USA}
}

@article{ddpm,
  title={Denoising diffusion probabilistic models},
  author={Ho, Jonathan and Jain, Ajay and Abbeel, Pieter},
  journal={Advances in neural information processing systems},
  volume={33},
  pages={6840--6851},
  year={2020}
}

@article{pixwizard,
  title={Pixwizard: Versatile image-to-image visual assistant with open-language instructions},
  author={Lin, Weifeng and Wei, Xinyu and Zhang, Renrui and Zhuo, Le and Zhao, Shitian and Huang, Siyuan and Teng, Huan and Xie, Junlin and Qiao, Yu and Gao, Peng and others},
  journal={arXiv preprint arXiv:2409.15278},
  year={2024}
}

@article{var,
  title={Visual autoregressive modeling: Scalable image generation via next-scale prediction},
  author={Tian, Keyu and Jiang, Yi and Yuan, Zehuan and Peng, Bingyue and Wang, Liwei},
  journal={Advances in neural information processing systems},
  volume={37},
  pages={84839--84865},
  year={2024}
}

@inproceedings{dit,
  title={Scalable diffusion models with transformers},
  author={Peebles, William and Xie, Saining},
  booktitle={Proceedings of the IEEE/CVF international conference on computer vision},
  pages={4195--4205},
  year={2023}
}

@article{varformer,
  title={Varformer: Adapting VAR's Generative Prior for Image Restoration},
  author={Wang, Siyang and Zhao, Feng},
  journal={arXiv preprint arXiv:2412.21063},
  year={2024}
}

@article{varsr,
  title={Visual Autoregressive Modeling for Image Super-Resolution},
  author={Qu, Yunpeng and Yuan, Kun and Hao, Jinhua and Zhao, Kai and Xie, Qizhi and Sun, Ming and Zhou, Chao},
  journal={arXiv preprint arXiv:2501.18993},
  year={2025}
}

@article{hart,
  title={Hart: Efficient visual generation with hybrid autoregressive transformer},
  author={Tang, Haotian and Wu, Yecheng and Yang, Shang and Xie, Enze and Chen, Junsong and Chen, Junyu and Zhang, Zhuoyang and Cai, Han and Lu, Yao and Han, Song},
  journal={arXiv preprint arXiv:2410.10812},
  year={2024}
}

@article{dcpt,
  title={Universal Image Restoration Pre-training via Degradation Classification},
  author={Hu, JiaKui and Jin, Lujia and Yao, Zhengjian and Lu, Yanye},
  journal={arXiv preprint arXiv:2501.15510},
  year={2025}
}

@inproceedings{patchgan,
  title={Image-to-image translation with conditional adversarial networks},
  author={Isola, Phillip and Zhu, Jun-Yan and Zhou, Tinghui and Efros, Alexei A},
  booktitle={Proceedings of the IEEE conference on computer vision and pattern recognition},
  pages={1125--1134},
  year={2017}
}

@article{rope,
  title={Roformer: Enhanced transformer with rotary position embedding},
  author={Su, Jianlin and Ahmed, Murtadha and Lu, Yu and Pan, Shengfeng and Bo, Wen and Liu, Yunfeng},
  journal={Neurocomputing},
  volume={568},
  pages={127063},
  year={2024},
  publisher={Elsevier}
}

@article{adair,
  title={Adair: Adaptive all-in-one image restoration via frequency mining and modulation},
  author={Cui, Yuning and Zamir, Syed Waqas and Khan, Salman and Knoll, Alois and Shah, Mubarak and Khan, Fahad Shahbaz},
  journal={arXiv preprint arXiv:2403.14614},
  year={2024}
}

@article{radford2019language,
  title={Language models are unsupervised multitask learners},
  author={Radford, Alec and Wu, Jeffrey and Child, Rewon and Luan, David and Amodei, Dario and Sutskever, Ilya and others},
  journal={OpenAI blog},
  volume={1},
  number={8},
  pages={9},
  year={2019}
}

@article{touvron2023llama,
  title={Llama: Open and efficient foundation language models},
  author={Touvron, Hugo and Lavril, Thibaut and Izacard, Gautier and Martinet, Xavier and Lachaux, Marie-Anne and Lacroix, Timoth{\'e}e and Rozi{\`e}re, Baptiste and Goyal, Naman and Hambro, Eric and Azhar, Faisal and others},
  journal={arXiv preprint arXiv:2302.13971},
  year={2023}
}

@inproceedings{van2016pixel,
  title={Pixel recurrent neural networks},
  author={Van Den Oord, A{\"a}ron and Kalchbrenner, Nal and Kavukcuoglu, Koray},
  booktitle={International conference on machine learning},
  pages={1747--1756},
  year={2016},
  organization={PMLR}
}

@article{van2016conditional,
  title={Conditional image generation with pixelcnn decoders},
  author={Van den Oord, Aaron and Kalchbrenner, Nal and Espeholt, Lasse and Vinyals, Oriol and Graves, Alex and others},
  journal={Advances in neural information processing systems},
  volume={29},
  year={2016}
}

@article{salimans2017pixelcnn++,
  title={Pixelcnn++: Improving the pixelcnn with discretized logistic mixture likelihood and other modifications},
  author={Salimans, Tim and Karpathy, Andrej and Chen, Xi and Kingma, Diederik P},
  journal={arXiv preprint arXiv:1701.05517},
  year={2017}
}

@inproceedings{chen2019pixelsnail,
  title={Pixelsnail: An improved autoregressive generative model},
  author={Chen, Xi and Mishra, Nikhil and Rohaninejad, Mostafa and Abbeel, Pieter},
  booktitle={International conference on machine learning},
  pages={864--872},
  year={2018},
  organization={PMLR}
}

@article{oord2018neural,
  title={Neural discrete representation learning},
  author={Van Den Oord, Aaron and Vinyals, Oriol and others},
  journal={Advances in neural information processing systems},
  volume={30},
  year={2017}
}

@inproceedings{esser2021taming,
  title={Taming transformers for high-resolution image synthesis},
  author={Esser, Patrick and Rombach, Robin and Ommer, Bjorn},
  booktitle={Proceedings of the IEEE/CVF conference on computer vision and pattern recognition},
  pages={12873--12883},
  year={2021}
}

@inproceedings{ramesh2021zeroshot,
  title={Zero-shot text-to-image generation},
  author={Ramesh, Aditya and Pavlov, Mikhail and Goh, Gabriel and Gray, Scott and Voss, Chelsea and Radford, Alec and Chen, Mark and Sutskever, Ilya},
  booktitle={International conference on machine learning},
  pages={8821--8831},
  year={2021},
  organization={Pmlr}
}

@article{ren2024mvar,
  title={M-VAR: Decoupled Scale-wise Autoregressive Modeling for High-Quality Image Generation},
  author={Ren, Sucheng and Yu, Yaodong and Ruiz, Nataniel and Wang, Feng and Yuille, Alan and Xie, Cihang},
  journal={arXiv preprint arXiv:2411.10433},
  year={2024}
}

@article{guo2025fastvar,
  title={FastVAR: Linear Visual Autoregressive Modeling via Cached Token Pruning},
  author={Guo, Hang and Li, Yawei and Zhang, Taolin and Wang, Jiangshan and Dai, Tao and Xia, Shu-Tao and Benini, Luca},
  journal={arXiv preprint arXiv:2503.23367},
  year={2025}
}

@inproceedings{chen2020image,
  title={Generative pretraining from pixels},
  author={Chen, Mark and Radford, Alec and Child, Rewon and Wu, Jeffrey and Jun, Heewoo and Luan, David and Sutskever, Ilya},
  booktitle={International conference on machine learning},
  pages={1691--1703},
  year={2020},
  organization={PMLR}
}

@inproceedings{yu2023video,
  title={Video probabilistic diffusion models in projected latent space},
  author={Yu, Sihyun and Sohn, Kihyuk and Kim, Subin and Shin, Jinwoo},
  booktitle={Proceedings of the IEEE/CVF conference on computer vision and pattern recognition},
  pages={18456--18466},
  year={2023}
}

@inproceedings{lolblur,
  title={Lednet: Joint low-light enhancement and deblurring in the dark},
  author={Zhou, Shangchen and Li, Chongyi and Change Loy, Chen},
  booktitle={European conference on computer vision},
  pages={573--589},
  year={2022},
  organization={Springer}
}

@inproceedings{dfpir,
  title={Degradation-Aware Feature Perturbation for All-in-One Image Restoration},
  author={Tian, Xiangpeng and Liao, Xiangyu and Liu, Xiao and Li, Meng and Ren, Chao},
  booktitle={Proceedings of the Computer Vision and Pattern Recognition Conference},
  pages={28165--28175},
  year={2025}
}

@inproceedings{cdd,
  title={Onerestore: A universal restoration framework for composite degradation},
  author={Guo, Yu and Gao, Yuan and Lu, Yuxu and Zhu, Huilin and Liu, Ryan Wen and He, Shengfeng},
  booktitle={European conference on computer vision},
  pages={255--272},
  year={2024},
  organization={Springer}
}

@article{ddim,
  title={Denoising diffusion implicit models},
  author={Song, Jiaming and Meng, Chenlin and Ermon, Stefano},
  journal={arXiv preprint arXiv:2010.02502},
  year={2020}
}

@inproceedings{wang2025target,
  title={Target-driven distillation: Consistency distillation with target timestep selection and decoupled guidance},
  author={Wang, Cunzheng and Guo, Ziyuan and Duan, Yuxuan and Li, Huaxia and Chen, Nemo and Tang, Xu and Hu, Yao},
  booktitle={Proceedings of the AAAI Conference on Artificial Intelligence},
  volume={39},
  number={7},
  pages={7619--7627},
  year={2025}
}

@article{lumina,
  title={Lumina-next: Making lumina-t2x stronger and faster with next-dit},
  author={Zhuo, Le and Du, Ruoyi and Xiao, Han and Li, Yangguang and Liu, Dongyang and Huang, Rongjie and Liu, Wenze and Zhu, Xiangyang and Wang, Fu-Yun and Ma, Zhanyu and others},
  journal={Advances in Neural Information Processing Systems},
  volume={37},
  pages={131278--131315},
  year={2024}
}
\bibliographystyle{iclr2026_conference}

\newpage
\appendix

\section{Overview of Supplementary}
\label{supsec: overview}

In this supplementary, we first present detailed computational complexity comparisons of RestoreVAR and LDM-based AiOR approaches. We then provide a theoretical analysis comparing the time complexities of VAR and LDMs. Subsequently, we analyze the effect of using Absolute Positional Embeddings (APE) versus Rotary Positional Embeddings (RoPE)~\citep{rope} when scaling the resolution from $256 \times 256$ to $512 \times 512$. Next, we present more architectural details of RestoreVAR, followed by a breakdown of the runtime and parameter count for each component of RestoreVAR. We then provide additional visual results, which include qualitative results for the continuous vs. discrete conditioning ablation, visual comparisons of VAE decoders, refiner ablation, and more qualitative comparisons with other methods. Subsequently, we provide experiments to show that the structure correction module of AutoDIR~\citep{autodir} behaves as an independent non-generative restoration network. We then discuss the limitations of our approach and scope for future work. Finally, we mention the usage of LLMs in the paper. To summarize, the supplementary discusses the following:
\begin{enumerate}
    \item Detailed computation complexity comparisons   (Sec.~\ref{supsec: computation_full})
    \item Theoretical complexity comparison with LDM (Sec.~\ref{supsec: theoretical})
    
    \item Additional Ablations (Sec.~\ref{supsec: additional_ablt})
    \begin{enumerate}
        \item Performance analysis: APE vs. RoPE (Sec.~\ref{supsubsec: apevsrope})
       
        \item Additional VAE loss ablations (Sec.~\ref{supsubsec: l1_ssim})
        \item Additional scale-space analysis (Sec.~\ref{supsubsec: scale_space_extra})
        \item Additional LRT analysis (Sec.~\ref{supsubsec: lrt_analysis})
    \end{enumerate}
    \item Additional Quantitative Results (Sec.~\ref{supsubsec: addn_quant})
    \begin{enumerate}
        \item User Study (Sec.~\ref{supsubsec: user_study_extra})
        \item Generalization Metrics (Sec.~\ref{supsubsec: generalization_extra})
        \item More Perceptual Quality Evaluations (Sec.~\ref{supsubsec: percep_extra})
    \end{enumerate}
    
    \item Additional Architectural Details (Sec.~\ref{supsec: arch_details})
    \item Runtime and Parameter Breakdown (Sec.~\ref{supsec: runtime})
    \item Additional Visual Results (Sec.~\ref{supsec: visual_results})
    \begin{enumerate}
        \item  Continuous vs. Discrete Conditioning (Sec.~\ref{supsubsec: contvsdisc})
        \item Qualitative comparisons of VAE decoders (Sec.~\ref{supsubsec: vae_comp_qual})
        \item Visual Comparison of Refiner Variants (Sec.~\ref{supsubsec: refiner})
        \item Additional Qualitative Comparisons (Sec.~\ref{supsubsec: add_qual})
    \end{enumerate}
    \item More details about AutoDIR comparison (Sec.~\ref{supsec: autodir_comp})
    \item Limitations and scope for future work
    (Sec.~\ref{supsec: limitations})
    \item LLM Usage
    (Sec.~\ref{supsec: llm_usage})

\end{enumerate}


\section{Detailed computation complexity comparisons}
\label{supsec: computation_full}
\begin{wraptable}{t}{0.5\textwidth}
  \centering
    \small
    \setlength{\tabcolsep}{1pt}
    \caption{Comparison of the computational complexity of RestoreVAR with LDM-based AiOR approaches.}
    \begin{tabular}{l c c c c}
      \toprule
      \textbf{Method} & \textbf{Steps} & \textbf{Time (s)} & \textbf{TFLOPs} & \textbf{Params (M)} \\
      \midrule
      Diff-Plugin   & 20  & 2.04   & 16.08  & 859.50  \\
      AutoDIR       & 100 & 8.477  & 67.80  & 859.50  \\
      PixWizard     & 60  & 8.247  & 19.27  & 2011.40 \\
      RestoreVAR    & 10  & 0.201  & 1.05   & 296.95  \\
      \bottomrule
    \end{tabular}
    \label{suptab:computation}
  \vspace{-8pt}
\end{wraptable}
RestoreVAR achieves substantial performance improvements over LDM-based AiOR approaches at a fraction of their computational cost. To show this, we compared RestoreVAR with Diff-Plugin~\citep{diffplugin}, AutoDIR~\citep{autodir}, and PixWizard~\citep{pixwizard} in terms of inference steps, runtime, TeraFLOPs, and total parameter count. As shown in Table~\ref{suptab:computation}, RestoreVAR achieves a $\mathbf{10\times}$ speed-up over Diff-Plugin and a $\sim\mathbf{16\times}$ reduction in TFLOPs.  Compared to AutoDIR and PixWizard, RestoreVAR is over $\mathbf{40\times}$ faster in inference. 

\begin{figure}[t]
  \centering
  \begin{minipage}[t]{0.48\linewidth}
    \centering
    \includegraphics[width=\linewidth]{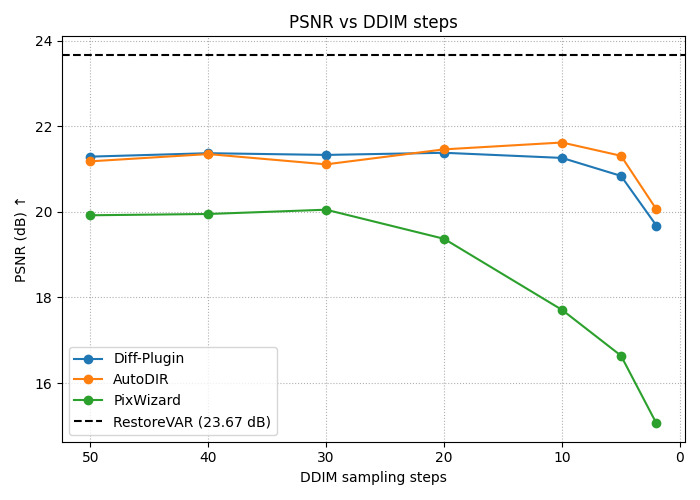}
    \vspace{-6pt}
    \caption*{(a) PSNR vs DDIM steps}
  \end{minipage}
  \hfill
  \begin{minipage}[t]{0.48\linewidth}
    \centering
    \includegraphics[width=\linewidth]{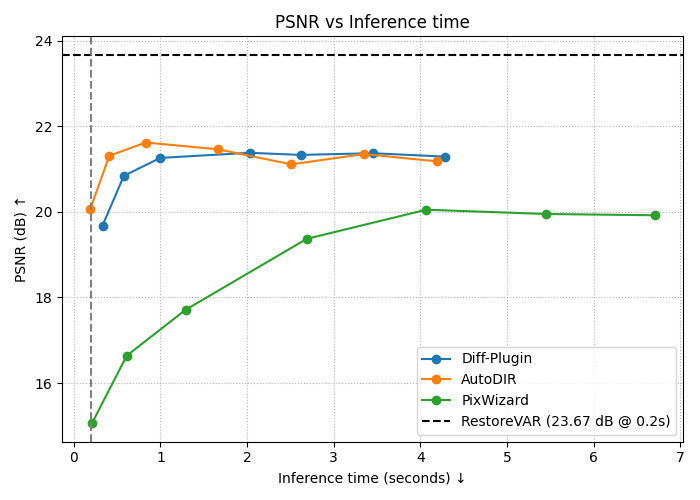}
    \vspace{-6pt}
    \caption*{(b) PSNR vs inference time}
  \end{minipage}
  \vspace{-6pt}
  \caption{Comparison of LDM-based methods accelerated using DDIM sampling and RestoreVAR. RestoreVAR achieves best results.}
  \label{fig:psnr_ddim_runtime}
\end{figure}

Additionally, we conducted an experiment to speed up LDM sampling using DDIM~\cite{ddim} sampling and compared the performance with RestoreVAR. Specifically, we varied the number of DDIM sampling steps as $50, 40, 30, 20, 10, 5 \text{ and } 2$. Figs.~\ref{fig:psnr_ddim_runtime}(a) and (b) shows the variation of mean (across the datasets from Table 1) PSNR (dB) scores with inference time (seconds) and DDIM sampling steps. Diffusion models match the inference time of RestoreVAR only when using around $2$ sampling steps, but show over a $3$dB decrease in PSNR compared to RestoreVAR. Even at higher step counts with DDIM sampling, diffusion models lag by over $1$dB, highlighting RestoreVAR's clear advantages in both speed and performance over LDM-based methods.


Consistency models and rectified-flow approaches are promising recent developments in accelerating diffusion models. However, both require substantial modifications to the training pipeline. Consistency models demand fine-tuning pre-trained diffusion models with a new consistency objective that enables few-step inference. However, using very few steps can cause over-smoothed outputs that are undesirable for image restoration~\citep{wang2025target}. Rectified-flow methods learn a deterministic velocity field along a straight-line path between noise and data, using a flow-matching objective. \begin{wraptable}{t}{0.3\textwidth}
   \small
        \setlength{\tabcolsep}{1pt}
        \caption{Comparison of the computational complexity of RestoreVAR with non-generative approaches.}
        \begin{tabular}{l c c}
          \toprule
          \textbf{Method} & \textbf{Time (s)} & \textbf{Params (M)} \\
          \midrule
          PromptIR   & 0.162& 35.59  \\
          InstructIR & 0.023 & 15.84 \\
          AWRaCLe  &  0.188   & 186.68  \\
          DCPT  & 0.024 & 67.88  \\
          DFPIR  & 0.117 & 182.38 \\
          RestoreVAR  &  0.201 & 296.95  \\
          \bottomrule
        \end{tabular}
        \label{suptab:computation_nongen}
      \vspace{-10pt}
\end{wraptable} Although these frameworks speed up inference, they still lack the scale-space (Sec.~\ref{subsec: var_motivation}) insight offered by VAR, which elegantly fits the objective of all-in-one image restoration.


We also provide computational complexity comparisons with non-generative methods in Table~\ref{suptab:computation_nongen}. 
Non-generative models are generally faster since RestoreVAR, like other generative approaches, requires multiple inference steps. Nevertheless, RestoreVAR maintains competitive inference speed to methods such as PromptIR, AWRaCLe, and DFPIR, even though it contains more parameters and requires a multi-step generation process. While it is slower than methods such as DCPT and InstructIR, the substantially improved generalization and superior perceptual quality offered by RestoreVAR make it a more robust choice.

\section{Theoretical complexity comparison with LDM}
\label{supsec: theoretical}

We now provide a theoretical comparison of the run-time complexities of VAR and diffusion transformers (DiT), offering fundamental insights into their efficiency differences. The VAR time complexity derivation closely follows that in the original VAR paper.

Let $a > 1$ be the geometric factor of the vector quantized (VQ) scale pyramid and let the largest scale have dimensions $h = w = n$.  
Let the number of scales be $K = \log_a n + 1$ so that the side length at scale $i$ is $n_i = a^{i-1}$ and the largest scale is $n_K = n = a^{K-1}$.  
Assume a standard self-attention transformer as in VAR with time complexity $\mathcal{O}(T^2)$ for $T$ tokens.

\paragraph{VAR.}
At the generation of the $k$-th scale the total number of tokens across the current and previous scales $(r_1, \dots, r_k)$ is
\[
\sum_{i=1}^k n_i^2
= \sum_{i=1}^k a^{2(i-1)}
= \frac{a^{2k} - 1}{a^2 - 1}.
\]
Hence the cost of generation of the $k$-th scale is
\[
C_k^{\text{VAR}} \;=\; \left(\frac{a^{2k}-1}{a^2-1}\right)^2.
\]
Summing over all $K$ scales gives
\[
C^{\text{VAR}} \;=\; \sum_{k=1}^K \left(\frac{a^{2k}-1}{(a^2-1)}\right)^2
= \frac{1}{(a^2-1)^2}\sum_{k=1}^K \big(a^{4k}-2a^{2k}+1\big).
\]
Substituting $K = \log_a n + 1$ (so that $a^{2K} = a^2n^2$ and $a^{4K} = a^4 n^4$) yields
\[
C^{\text{VAR}} = \frac{1}{(a^2-1)^2}\left[
\frac{a^4(n^4-1)}{a^4-1}
- \frac{2a^2(a^2n^2-1)}{a^2-1}
+ (\log_a n + 1)
\right].
\]
The asymptotic time complexity is governed by the dominant term:
\[
C^{\text{VAR}} \;\sim\; \frac{a^8}{(a^4-1)(a^2-1)^2}\,n^4 \;=\; \mathcal{O}(n^4).
\]

\paragraph{Diffusion.}
Assume a self-attention DiT where each diffusion step uses the fixed largest resolution $n\times n$, i.e., $n^2$ tokens.  
So, a single step costs
\[
C^{\text{Diff}}_1 = (n^2)^2 = n^4.
\]
With the same number of forward steps as VAR, namely $K = \log_a n + 1$, we get
\[
C^{\text{Diff}} = \sum_{k=1}^K n^4 = (\log_a n + 1)n^4 = \mathcal{O}(n^4 \log n).
\]

\paragraph{Comparison.}
From the above,
\[
\frac{C^{\text{Diff}}}{C^{\text{VAR}}}
\sim \frac{(\log_a n)\,n^4}{\tfrac{(a^4-1)(a^2-1)^2}{a^8}\,n^4}
= \frac{a^8}{(a^4-1)(a^2-1)^2}\,\log_a n = \mathcal{O}(\log n).
\]
That is, with the same number of forward passes, VAR totals $\mathcal{O}(n^4)$ complexity while diffusion totals $\mathcal{O}(n^4 \log n)$, yielding a $\mathcal{O}(\log n)$ speedup for VAR.

\section{Additional Ablations}
\label{supsec: additional_ablt}

In this section, we provide additional ablations and analysis of RestoreVAR and its components.

\subsection{Performance Analysis: APE vs. RoPE}
\label{supsubsec: apevsrope}

As discussed in Sec.~\ref{subsec: restorevar}, we replace the absolute position embeddings (APE) used in VAR~\citep{var} with Rotary Positional Embeddings~\citep{rope} (RoPE). We found that using RoPE yields better performance when scaling the resolution from $256 \times 256$ to $512 \times 512$. \begin{wrapfigure}{r}{0.5\textwidth}
    \centering
    \includegraphics[width=\linewidth]{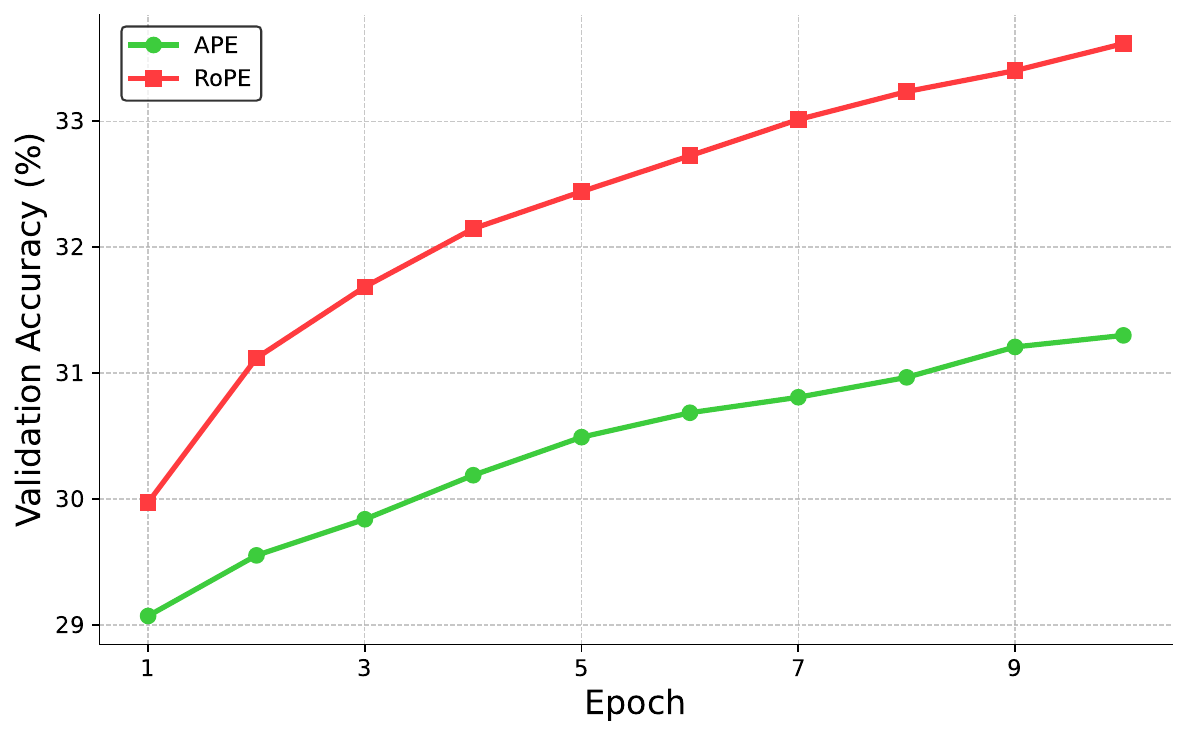}
    \caption{Validation accuracy comparison of APE and RoPE-based fine-tuning at $512 \times 512$ resolution. RoPE demonstrates better performance.}
    \label{supfig:apevsrope}
    \vspace{-10pt}
\end{wrapfigure} To demonstrate this, we conducted an ablation where both APE and RoPE-based variants were fine-tuned at $512 \times 512$ resolution for $10$ epochs. As shown in Fig.~\ref{supfig:apevsrope}, the RoPE-based model achieves higher validation accuracy compared to the APE-based model, indicating its effectiveness.

\subsection{Additional VAE loss ablations}
\label{supsubsec: l1_ssim}

\begin{figure}[t]
  \centering
  
    \begin{subfigure}[t]{0.195\linewidth}
      \includegraphics[width=\linewidth]{figs/proc_vae/Snow100k/GT/winter_tree_12895.png.jpg}
      \caption*{Input image}
    \end{subfigure}\hfill
    \begin{subfigure}[t]{0.195\linewidth}
      \includegraphics[width=\linewidth]{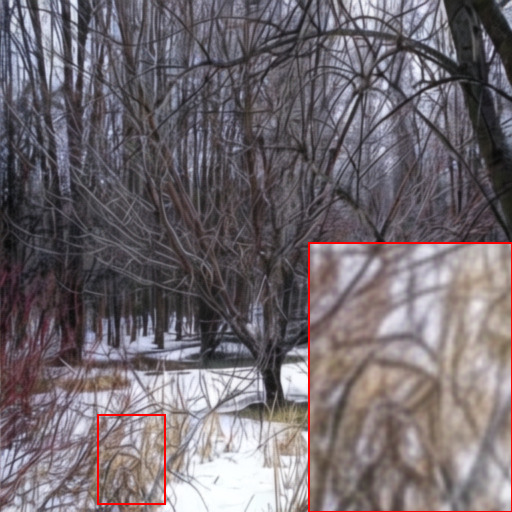}
      \caption*{L1 only}
    \end{subfigure}\hfill
    \begin{subfigure}[t]{0.195\linewidth}
      \includegraphics[width=\linewidth]{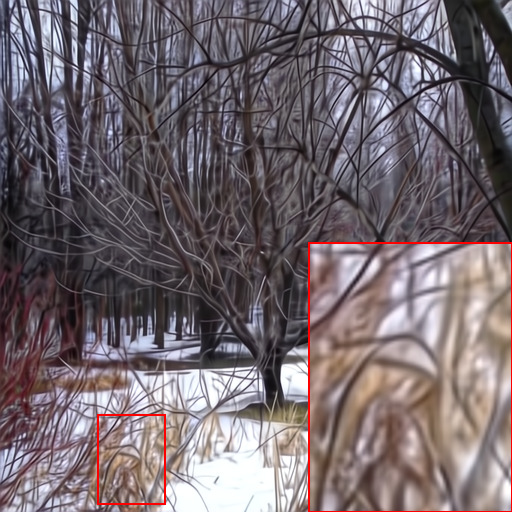}
      \caption*{L1 + SSIM}
    \end{subfigure}\hfill
    \begin{subfigure}[t]{0.195\linewidth}
      \includegraphics[width=\linewidth]{figs/proc_vae/Snow100k/vae_outputs/nodisc/winter_tree_12895.png.jpg}
      \caption*{w/o Disc}
    \end{subfigure}\hfill
    \begin{subfigure}[t]{0.195\linewidth}
      \includegraphics[width=\linewidth]{figs/proc_vae/Snow100k/vae_outputs/disc/winter_tree_12895.png.jpg}
      \caption*{w Disc}
    \end{subfigure}
    \caption{Image reconstructed by VAE decoders fine-tuned on continuous latents with various objective functions.}
    \label{supfig: l1_ssim}
  \vspace{-5pt}
\end{figure}

We now provide ablations on fine-tuning the VAE decoder without SSIM and perceptual losses. Fig.~\ref{supfig: l1_ssim} illustrates that using only L1 loss produces blurry results, adding the SSIM losses enhances structural details and incorporating the perceptual loss improves sharpness while preserving the structure (w/o Disc column). Incorporating the discriminator enhances sharpness even further.

  

\subsection{Additional scale-space analysis}
\label{supsubsec: scale_space_extra}

\begin{wrapfigure}{r}{0.526\textwidth}
    \centering
    \small
    \setlength{\tabcolsep}{1pt}
    \begin{tabular}{ccccc}
    &Degraded&GT$+$coarse&GT$+$fine&GT\\
        \rotatebox[origin=c]{90}{POLED\hspace{-26pt}}&\includegraphics[height=1.17cm, width=1.68cm]{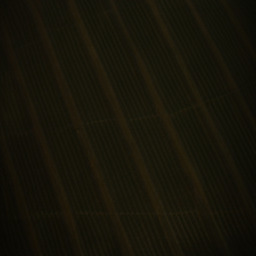}&\includegraphics[height=1.17cm, width=1.68cm]{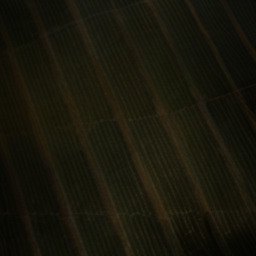}&\includegraphics[height=1.17cm, width=1.68cm]{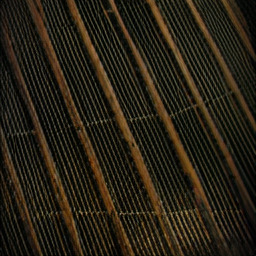}&\includegraphics[height=1.17cm, width=1.68cm]{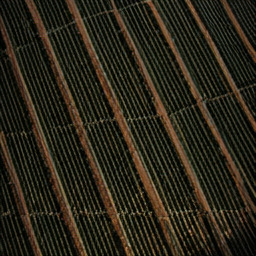}\\

        \rotatebox[origin=c]{90}{REVIDE\hspace{-26pt}}&\includegraphics[height=1.17cm, width=1.68cm]{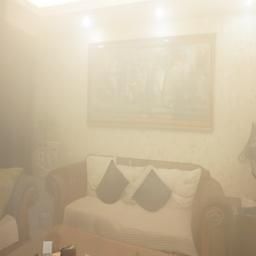}&\includegraphics[height=1.17cm, width=1.68cm]{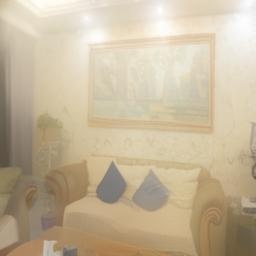}&\includegraphics[height=1.17cm, width=1.68cm]{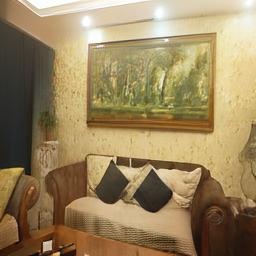}&\includegraphics[height=1.17cm, width=1.68cm]{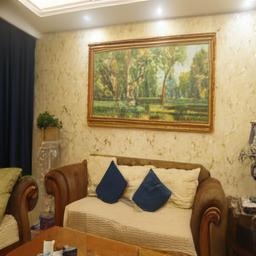}\\
    \end{tabular}
    \vspace{-8pt}
    \caption{Our analysis that VAR captures degradations in early scales (coarse) and scene-level details in later scales (fine) also tends to hold for real degradations from POLED and REVIDE datasets. GT$+$coarse replaces early GT scales with degraded ones, while GT$+$fine replaces the late GT scales.}
    \label{supfig: motivation_real}
    \vspace{-20pt}
\end{wrapfigure}

We provide an analysis similar to Fig.~\ref{fig: new_motivation} for real-world degradations using the POLED and REVIDE datasets. The results in Fig.~\ref{supfig: motivation_real} indicate that the scale-space decomposition of VAR tends to capture real-world degradations in coarse scales while preserving scene-level details in finer scales. This suggests that the scale-space behavior observed in the synthetic examples of Fig.~\ref{fig: new_motivation} also extends to real-world degradations. 

\subsection{Additional LRT analysis}
\label{supsubsec: lrt_analysis}

\begin{figure}[t]
  \centering
  \setlength{\tabcolsep}{2pt}
  \renewcommand{\arraystretch}{0.0}
  \begin{tabular}{cccc}
    Input &
    $f^\text{deg}_\text{quant} + \text{LRT}(f^\text{deg}_\text{quant}, z)$ &
    $f^\text{deg}_\text{cont} + \text{LRT}(f^\text{deg}_\text{cont}, z)$ &
    GT \\
    \includegraphics[width=0.23\textwidth]{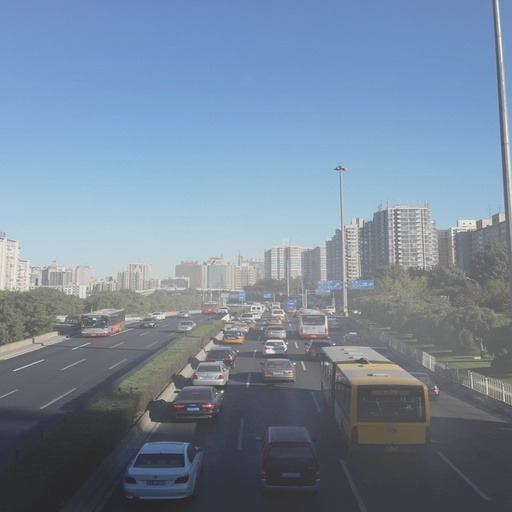} &
    \includegraphics[width=0.23\textwidth]{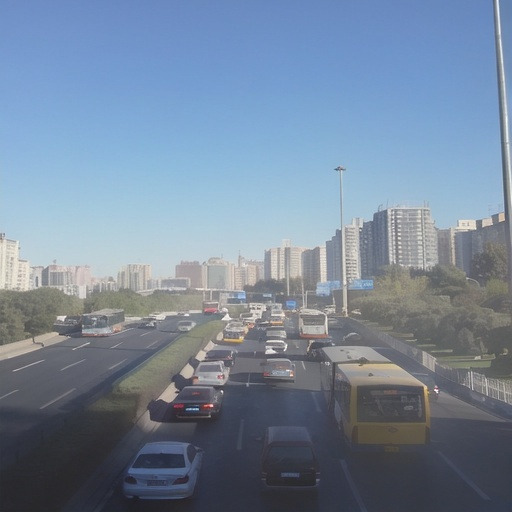} &
    \includegraphics[width=0.23\textwidth]{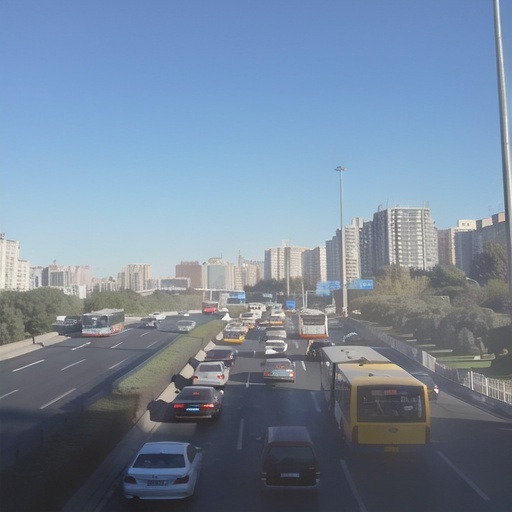} &
    \includegraphics[width=0.23\textwidth]{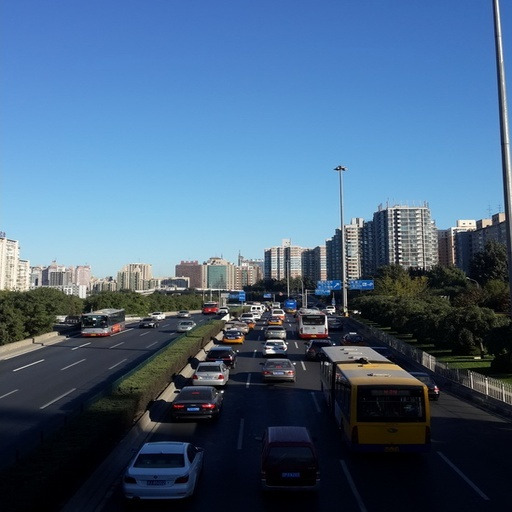} \\
    \includegraphics[width=0.23\textwidth]{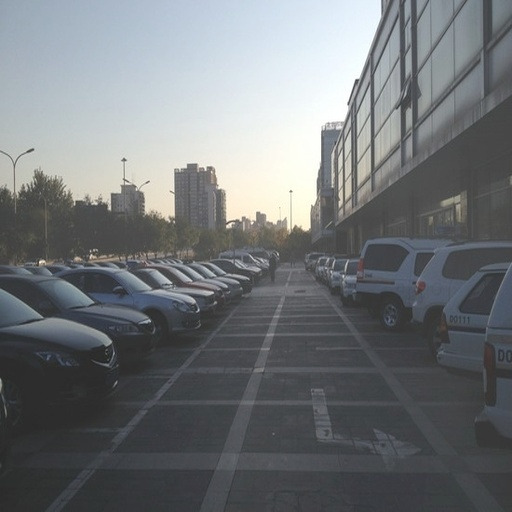} &
    \includegraphics[width=0.23\textwidth]{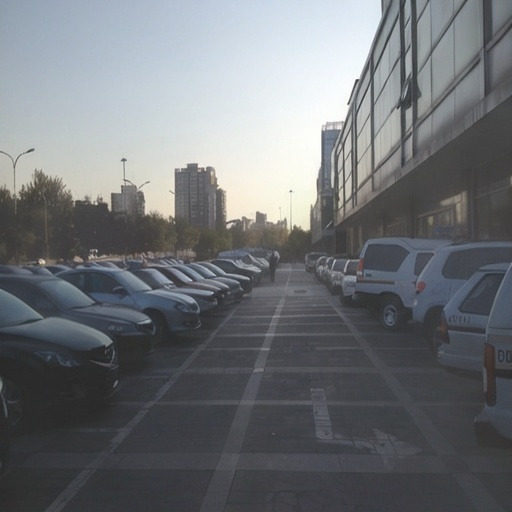} &
    \includegraphics[width=0.23\textwidth]{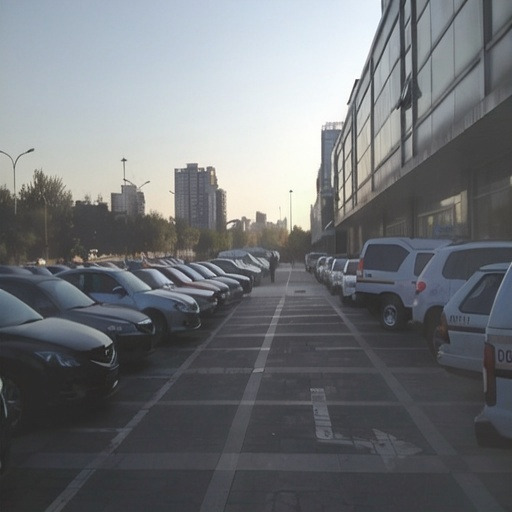} &
    \includegraphics[width=0.23\textwidth]{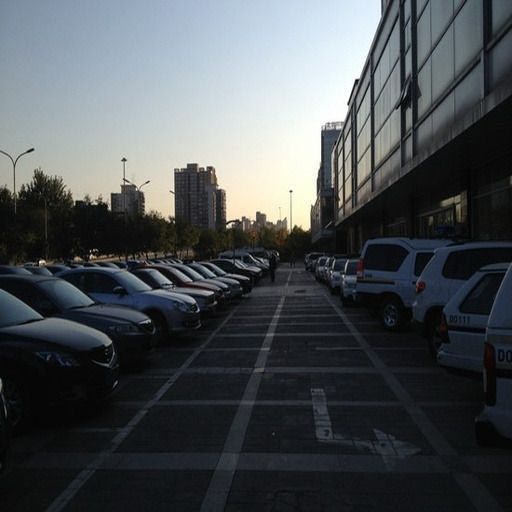} \\
  \end{tabular}
  \caption{
    Qualitative analysis of the LRT module operating directly on the degraded quantized and continuous latents. The prediction of the LRT remains hazy, indicating that the LRT cannot restore images without the VAR transformer.
  }
  \label{supfig:lrt_analysis}
\end{figure}

In this section, we analyze if the LRT module can perform restoration without relying on the VAR transformer. Toward this aim, we conduct two experiments:
\[
\hat{f}_{\text{cont}} = f^\text{deg}_{\text{quant}} + \text{LRT}(f^\text{deg}_{\text{quant}}, z),\vspace{-4pt}
\]

and 

\[
\hat{f}_{\text{cont}} = f^\text{deg}_{\text{cont}} + \text{LRT}(f^\text{deg}_{\text{cont}}, z),\vspace{-4pt}
\]

where $f^\text{deg}_\text{quant}$ and $f^\text{deg}_{\text{cont}}$ denote the quantized and continuous VAE latents obtained directly from the degraded input. Note that we are passing $z$ as input to the LRT since the LRT is trained to utilize $z$ to predict a meaningful residual. Zeroing out $z$ leads to significant artifacts as we are giving incorrect guidance to the LRT. Fig.~\ref{supfig:lrt_analysis} provides the results for these experiments. Despite having access to the restored latent's representation $z$, the LRT is unable to recover the clean latent (and so is the VAE decoder). This demonstrates that the LRT and the VAE decoder cannot independently perform restoration. These experiments reaffirm that restoration is carried out mainly by the VAR transformer, while the LRT and decoder serve to enhance fidelity by operating on VAR’s restored latent representation.

\section{Additional Quantitative Results}
\label{supsubsec: addn_quant}

\begin{wrapfigure}{r}{0.5\textwidth}
    \centering
    \includegraphics[width=1\linewidth]{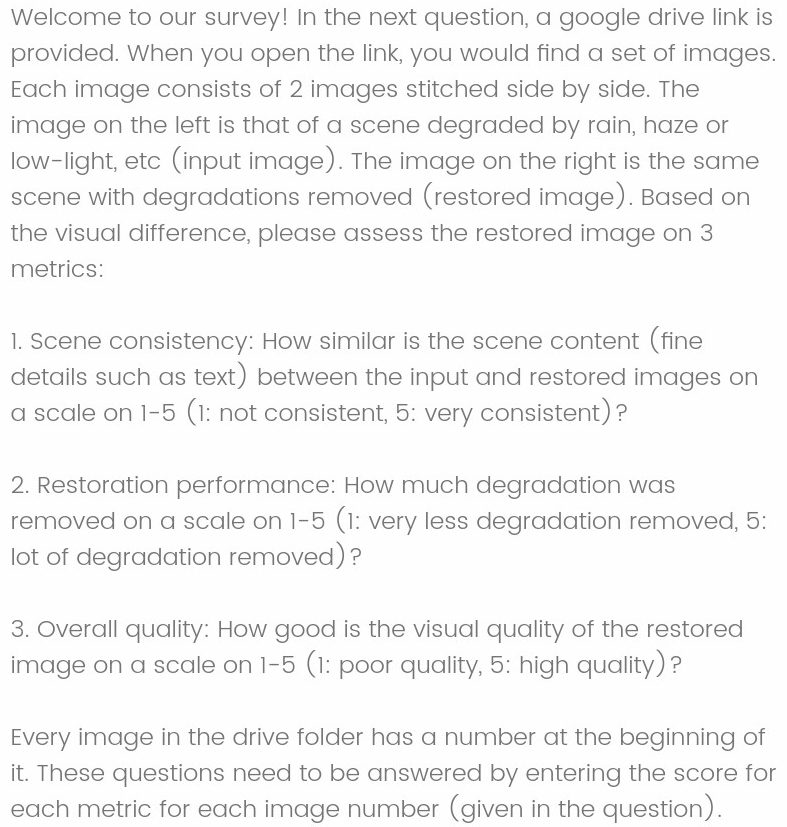}
    \caption{Screenshot of the user study instruction page.}
    \label{supfig:user_study_intro}
\end{wrapfigure}

In this section, we provide additional quantitative analysis of the results in Table~\ref{tab:within_dist}, Table~\ref{tab:generalization_metrics} and Table~\ref{tab:user_study}.


\subsection{User Study}
\label{supsubsec: user_study_extra}

We now provide rigorous statistical analysis for the user study, by computing paired differences between RestoreVAR and each method across all $36$ responses ($\Delta=\text{RestoreVAR} - \text{Method}$), and report $95\%$ confidence intervals (CI), paired t-tests, and Wilcoxon signed-rank tests. The results, summarized in Table \ref{suptab:user_study_stat}, show that RestoreVAR significantly outperforms all methods: every mean improvement $\Delta$ is positive and the $95\%$ CIs do not cross zero. Furthermore, the t-tests and Wilcoxon signed-rank tests report $p < 10^{-4}$. These results demonstrate that the advantages of RestoreVAR over non-generative methods are statistically significant across user study responses.

\begin{figure}[t]
    \centering
    \includegraphics[width=\linewidth]{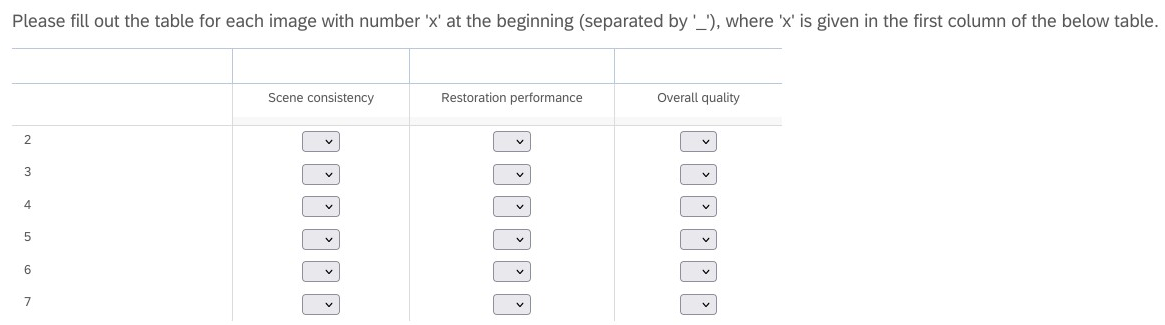}
    \caption{Screenshot of the user study data collection page.}
    \label{supfig:user_study_table}
\end{figure}

We also provide additional details for the user study methodology. We selected $50$ images for evaluation and collected $36$ independent responses. Participants evaluated each result on a five-point scale for (i) restoration performance, (ii) overall quality, and (iii) scene consistency. The survey was conducted using Qualtrics (Provo, UT). Screenshots of the survey are given in Figs.~\ref{supfig:user_study_intro} and~\ref{supfig:user_study_table}. Fig.~\ref{supfig:user_study_intro} shows the introduction page of the user study. The user would be provided with a randomly sampled google drive link consisting of restored outputs of different methods, where each method is assigned a number. The user needs to evaluate each of these images on the afore-mentioned criteria and enter the scores in the table shown in Fig.~\ref{supfig:user_study_table}.

\subsection{Generalization metrics}
\label{supsubsec: generalization_extra}

We additionally perform statistical significance analysis for our quantitative results in Table~\ref{tab:generalization_metrics}. Specifically, we compute per-image paired differences between RestoreVAR and each non-generative method on MUSIQ and CLIPIQA and apply dataset-weighted statistical tests to ensure that each dataset contributes equally to the global statistic. As shown in Table \ref{suptab:table2_stat_analysis}, RestoreVAR consistently achieves positive improvements with tight CIs and extremely small $p$ values. Once again, these results confirm that RestoreVAR achieves statistically significant gains in perceptual image-quality metrics over non-generative methods.

\subsection{More perceptual quality evaluations}
\label{supsubsec: percep_extra}
We provide no-reference metrics MUSIQ and CLIPIQA which gauge perceptual quality for the test sets in Table 1 where there is a significant gap between non-generative methods and RestoreVAR on pixel-level metrics. From the MUSIQ/CLIPIQA scores provided in Table~\ref{suptab:noref_within_dist}, it can be observed that the perceptual quality of RestoreVAR is on-par or even higher than those of non-generative methods, indicating perceptually comparable results. However, due to lower pixel-level fidelity than non-generative approaches, RestoreVAR scores lower on pixel-level metrics.




\begin{table}[t]
\centering
\small
\caption{Statistical analysis of user study results using $95\%$ confidence intervals (CIs), paired t-test and Wilcoxon signed-rank test. $\Delta$ denotes the mean paired difference between RestoreVAR and each method.}
\begin{tabular}{lcccc}
\toprule
Method & $\Delta$ & 95\% CI & $p$ (t-test) & $p$ (Wilcoxon) \\
\midrule
AutoDIR      & $0.694$ & $\pm 0.257$ & $6.49\times10^{-6}$  & $3.06\times10^{-5}$ \\
AWRaCLe      & $2.028$ & $\pm 0.245$ & $7.15\times10^{-18}$ & $2.91\times10^{-11}$ \\
PromptIR     & $2.250$ & $\pm 0.221$ & $1.12\times10^{-20}$ & $2.91\times10^{-11}$ \\
DCPT         & $1.944$ & $\pm 0.262$ & $2.15\times10^{-16}$ & $2.91\times10^{-11}$ \\
InstructIR   & $1.417$ & $\pm 0.293$ & $3.28\times10^{-11}$ & $1.66\times10^{-6}$ \\
\bottomrule
\end{tabular}
\label{suptab:user_study_stat}
\end{table}

\begin{table*}[t]
\centering
\small
\caption{Statistical analysis for results from Table~\ref{tab:generalization_metrics} using $95\%$ confidence intervals (CIs), paired t-test and Wilcoxon signed-rank test. $\Delta$ denotes the weighted mean paired difference between RestoreVAR and each method.}
\begin{tabular}{lcccccc}
\toprule
& \multicolumn{3}{c}{MUSIQ} & \multicolumn{3}{c}{CLIPIQA}\\
\cmidrule(lr){2-4} \cmidrule(lr){5-7}
Method & $\Delta$ & 95\% CI & $p$-value & $\Delta$ & 95\% CI & $p$-value\\
\midrule
AWRaCLe     & 5.084 & $\pm 0.554$ & $2.5\times10^{-55}$ & 0.0536 & $\pm 0.0080$ & $8.5\times10^{-34}$ \\
DCPT        & 5.206 & $\pm 0.613$ & $3.0\times10^{-49}$ & 0.0166 & $\pm 0.0079$ & $3.9\times10^{-5}$ \\
DFPIR       & 8.276 & $\pm 0.757$ & $2.5\times10^{-71}$ & 0.0733 & $\pm 0.0093$ & $3.6\times10^{-44}$ \\
InstructIR  & 6.984 & $\pm 0.902$ & $1.3\times10^{-42}$ & 0.0495 & $\pm 0.0065$ & $1.8\times10^{-41}$ \\
PromptIR    & 7.002 & $\pm 0.648$ & $3.3\times10^{-70}$ & 0.0395 & $\pm 0.0068$ & $8.4\times10^{-27}$ \\
\bottomrule
\end{tabular}%
\label{suptab:table2_stat_analysis}
\end{table*}

\begin{table*}[t]
  \centering
  \small
  \caption{Comparison of perceptual quality of RestoreVAR and non-generative methods on RESIDE, Snow100k, Rain13K, LOLv1 and GoPro datasets using MUSIQ and CLIPIQA metrics. }
  \vspace{-7pt}
  \setlength{\tabcolsep}{3pt}
  \resizebox{\textwidth}{!}{%
    \begin{tabular}{l cc cc cc cc cc cc}
      \toprule
      \multirow{2}{*}{\textbf{Method}} &
      \multicolumn{2}{c}{\textbf{GoPro}} &
      \multicolumn{2}{c}{\textbf{LOLv1}} &
      \multicolumn{2}{c}{\textbf{RESIDE}} &
      \multicolumn{2}{c}{\textbf{Rain13K}} &
      \multicolumn{2}{c}{\textbf{Snow100K}} &
      \multicolumn{2}{c}{\textbf{Average}} \\
      \cmidrule(lr){2-3}
      \cmidrule(lr){4-5}
      \cmidrule(lr){6-7}
      \cmidrule(lr){8-9}
      \cmidrule(lr){10-11}
      \cmidrule(lr){12-13}
        & MUSIQ$\uparrow$ & CLIPIQA$\uparrow$
        & MUSIQ$\uparrow$ & CLIPIQA$\uparrow$
        & MUSIQ$\uparrow$ & CLIPIQA$\uparrow$
        & MUSIQ$\uparrow$ & CLIPIQA$\uparrow$
        & MUSIQ$\uparrow$ & CLIPIQA$\uparrow$
        & MUSIQ$\uparrow$ & CLIPIQA$\uparrow$ \\
      \midrule
      PromptIR
        & 35.87 & 0.155  
        & 50.24 & 0.306  
        & 66.54 & 0.434  
        & 64.63 & 0.471  
        & 63.29 & 0.417  
        & 56.11 & 0.357  
        \\
      InstructIR
        & 53.06 & 0.251  
        & 67.55 & 0.364  
        & 67.59 & 0.449  
        & 66.51 & 0.504  
        & --    & --     
        & 63.68 & 0.392  
        \\
      AWRaCLe
      & 44.953 & 0.173
      & 63.198 & 0.459
      & 61.938 & 0.346
      & 64.683 & 0.598
      & 62.334 & 0.512
      & 59.421 & 0.418
        \\
      DCPT
        & 46.12 & 0.180  
        & 67.73 & 0.404  
        & 66.46 & 0.432  
        & 62.63 & 0.437  
        & --    & --     
        & 60.73 & 0.363  
        \\
      DFPIR
        & 48.03    & 0.190     
        & 66.77 & 0.409  
        & 66.55 & 0.423  
        & 62.28 & 0.432  
        & --    & --     
        & 60.91 & 0.364  
        \\
      \rowcolor{gray!25}
      \textbf{RestoreVAR}
        & 55.45 & 0.231  
        & 71.47 & 0.396  
        & 66.45 & 0.481  
        & 63.83 & 0.516  
        & 63.56 & 0.482  
        & 64.15 & 0.421  
        \\
      \bottomrule
    \end{tabular}%
  }
  \label{suptab:noref_within_dist}
  \vspace{-5pt}
\end{table*}

\section{Additional Architectural Details}
\label{supsec: arch_details}
We now provide additional architectural details for the RestoreVAR framework. We first describe the details for the RestoreVAR transformer, followed by the Latent Refiner Transformer (LRT).

\begin{figure}[h]
    \centering
    \includegraphics[width=0.95\linewidth]{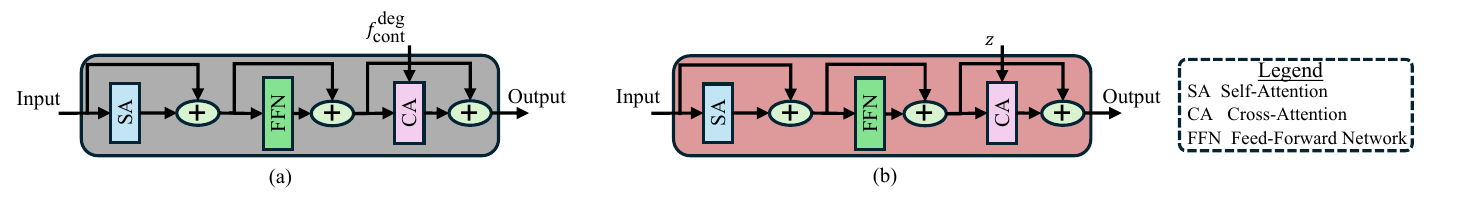}
    \vspace{-8pt}
    \caption{Illustration of a transformer block in (a) RestoreVAR transformer and (b) Latent Refiner Transformer.}
    \label{supfig:more_arch}
\end{figure}

\textbf{RestoreVAR Transformer. }We adopted the VAR model with a transformer depth of 16, i.e., the architecture consists of 16 transformer blocks. The structure of each block is illustrated in Fig.~\ref{supfig:more_arch}(a). The embedding dimension was set to $1024$, and the number of attention heads used was $16$. Furthermore, the transformer predicted discrete latents at the following spatial resolutions in the latent space:
$1 \times 1$, $2 \times 2$, $3 \times 3$, $4 \times 4$, $6 \times 6$,
$9 \times 9$, $13 \times 13$, $18 \times 18$, $24 \times 24$, and $32 \times 32$. The start-of-sequence (SOS) token is constructed by augmenting the class embedding with the mean value (along spatial dimensions) of the features obtained after a learnable projection applied on $f_{\text{cont}}^{\text{deg}}$. Specifically, 
\[
\text{SOS}= \text{class}_{\text{emb}} +  g_\text{sos} \times \text{Mean}(\text{Proj}(f_{\text{cont}}^{\text{deg}}), \text{SOS} \in \mathbb{R}^{1\times C}.
\]
Here, $\text{class}_{\text{emb}}$ is the class token embedding and $g_\text{sos}$ is initialized as $0$ for gradual incorporation of degradation conditioning. Other notations follow Sec. 3.

\textbf{Latent Refiner Transformer. }The LRT follows a similar structure for the blocks as the RestoreVAR transformer, as shown in Fig.~\ref{supfig:more_arch}(b). It was configured with a depth of $12$, six attention heads, and an embedding dimension of $384$.

\section{Runtime and Parameter Breakdown}
\label{supsec: runtime}
\begin{wraptable}{r}{0.45\textwidth}
  \centering
  \small
  \setlength{\tabcolsep}{1.5pt}
  \begin{tabular}{l c c c}
    \toprule
    \textbf{Component} 
      & \textbf{VAE} 
      & \textbf{Transformer} 
      & \textbf{Refiner} \\
    \midrule
    \textbf{Time (s)} 
      & 0.0086 
      & 0.1863 
      & 0.0061 \\
    \textbf{Parameters (M)}     
      & 108.95 
      & 273.98 
      & 22.97 \\
    \bottomrule
  \end{tabular}
  \vspace{-3pt}
  \caption{Compute time and parameter count breakdown for each component of RestoreVAR. VAE time includes both encoding and decoding.}
  \label{suptab:component_computation}
  \vspace{-6pt}
\end{wraptable}
In this section, we provide a breakdown of the runtime and parameter count for the following components of the RestoreVAR framework: the VAE, the RestoreVAR transformer, and the Latent Refiner Transformer (LRT). This analysis provides insights into the distribution of the computational cost across the pipeline. As shown in Table~\ref{suptab:component_computation}, the majority of inference time is taken by the autoregressive RestoreVAR transformer. 




\section{Additional Visual Results}
\label{supsec: visual_results}

We now present additional visualizations for some of the ablations discussed in Sec.~\ref{subsec: ablations}, along with more qualitative comparisons across methods.

\subsection{Continuous vs. Discrete Conditioning}
\label{supsubsec: contvsdisc}

\begin{figure}[h]
  \centering
  \small
  \setlength{\tabcolsep}{2pt}
  \begin{tabular}{@{}cccc@{}}
    \includegraphics[width=0.24\linewidth]{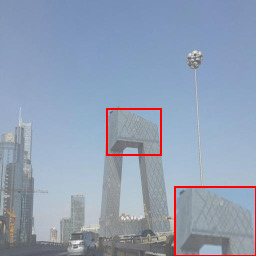} &
    \includegraphics[width=0.24\linewidth]{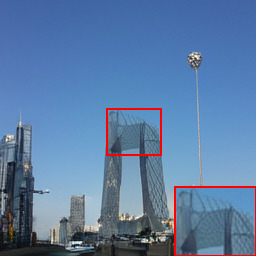} &
    \includegraphics[width=0.24\linewidth]{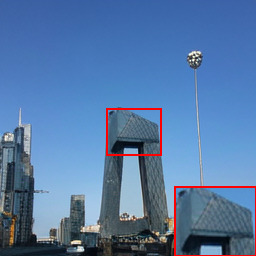} &
    \includegraphics[width=0.24\linewidth]{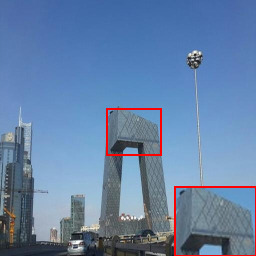} \\[4pt]
    \includegraphics[width=0.24\linewidth]{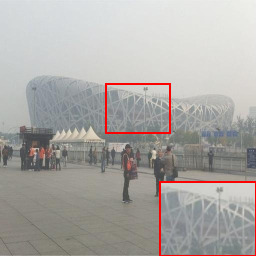} &
    \includegraphics[width=0.24\linewidth]{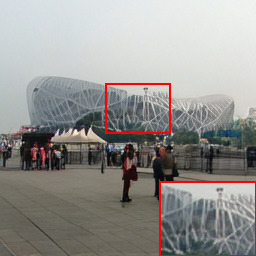} &
    \includegraphics[width=0.24\linewidth]{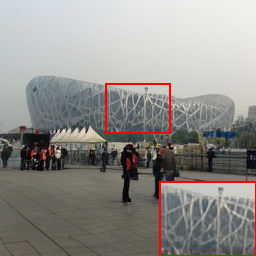} &
    \includegraphics[width=0.24\linewidth]{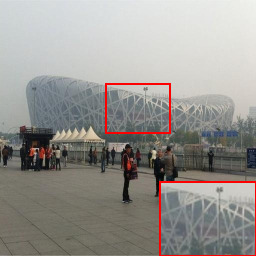} \\[2pt]
    \makebox[0.24\linewidth]{(a) Input} &
    \makebox[0.24\linewidth]{(b) Discrete} &
    \makebox[0.24\linewidth]{(c) Continuous} &
    \makebox[0.24\linewidth]{(d) GT} \\
  \end{tabular}
  \caption{Qualitative comparisons of RestoreVAR under discrete vs. continuous conditioning. RestoreVAR with discrete conditioning exhibits more hallucinations than the variant with continuous conditioning.}
  \label{supfig:contvsdisc}
\end{figure}

As shown in Sec.~\ref{subsec: ablations}, conditioning RestoreVAR on the continuous latent of the degraded image yields significantly better performance compared to using the quantized or discrete latent. Fig.~\ref{supfig:contvsdisc} further illustrates this using visual comparisons between the model trained with discrete and continuous conditioning. The model trained with discrete conditioning exhibits noticeably more hallucinations than the one trained with continuous conditioning.

\subsection{Qualitative comparisons of VAE decoders}
\label{supsubsec: vae_comp_qual}
\begin{wrapfigure}{t}{0.6\textwidth}
    \vspace{-20pt}
    \centering
    \begin{subfigure}[t]{0.24\linewidth}
        \includegraphics[width=\linewidth]{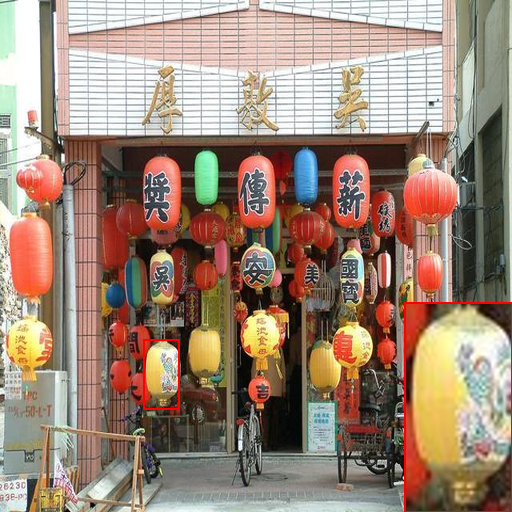}
        \caption*{(a) Input}
    \end{subfigure}\hspace{0.0001\linewidth}
    \begin{subfigure}[t]{0.24\linewidth}
        \includegraphics[width=\linewidth]{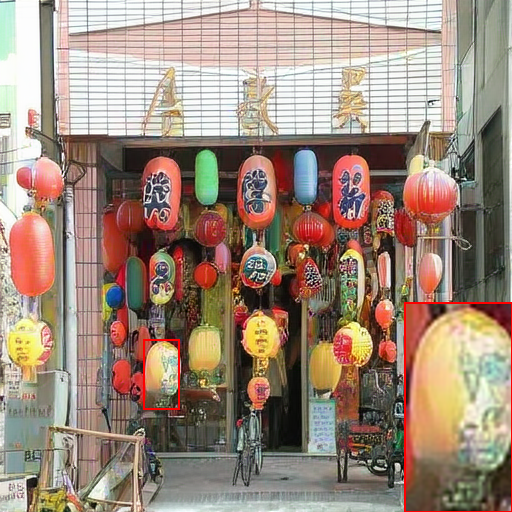}
        \caption*{(b) VAR}
    \end{subfigure}\hspace{0.0001\linewidth}
    \begin{subfigure}[t]{0.24\linewidth}
        \includegraphics[width=\linewidth]{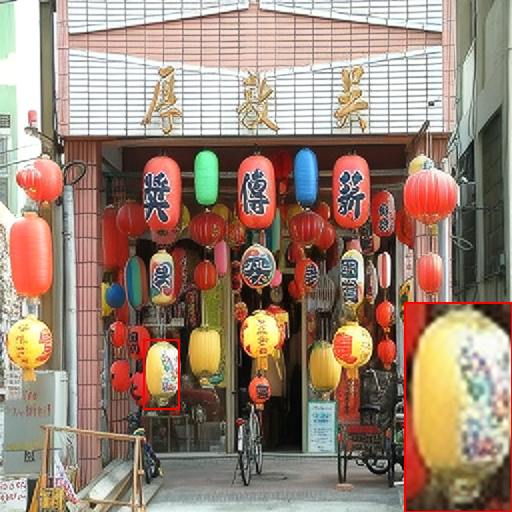}
        \caption*{(c) HART}
    \end{subfigure}\hspace{0.0001\linewidth}
    \begin{subfigure}[t]{0.24\linewidth}
        \includegraphics[width=\linewidth]{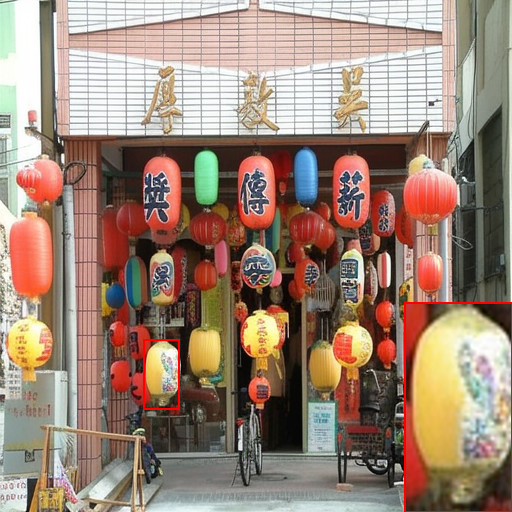}
        \caption*{(d) Ours}
    \end{subfigure}
    \vspace{-8pt}
    \caption{Qualitative comparisons of the input reconstructed using VAR~\citep{var}, HART~\citep{hart} and Our VAE decoder. Our result has minimal distortions.}
    \label{supfig:vae_artifacts}
    \vspace{-30pt}
\end{wrapfigure}
As mentioned in Sec.~\ref{subssubsec: lrm}, our fine-tuned VQ-VAE decoder achieves superior reconstruction performance compared to the decoders of VAR and HART. Fig.~\ref{supfig:vae_artifacts}, provides qualitative results to illustrate the same. Our decoder produces the best reconstruction.

\subsection{Visual Comparison of Refiner Variants}
\label{supsubsec: refiner}

\begin{figure}[t]
  \centering
  \setlength{\tabcolsep}{1pt}
  \begin{tabular}{@{}*{6}{c}@{}}
    \includegraphics[width=0.16\linewidth]{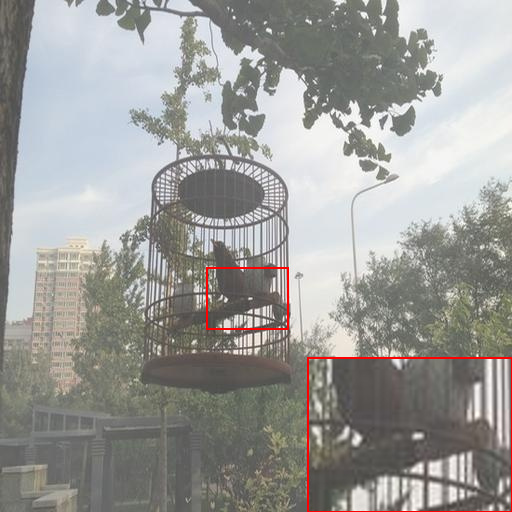} &
    \includegraphics[width=0.16\linewidth]{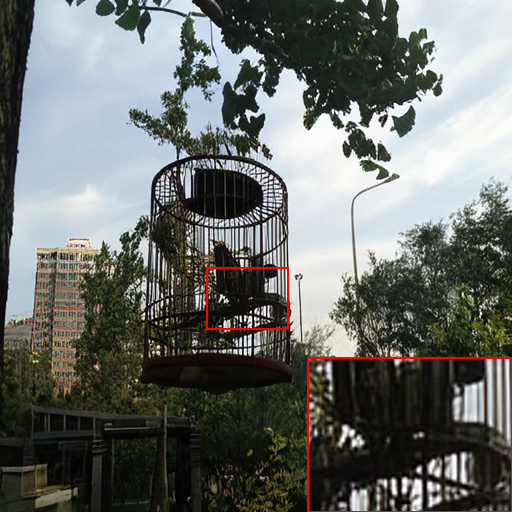} &
    \includegraphics[width=0.16\linewidth]{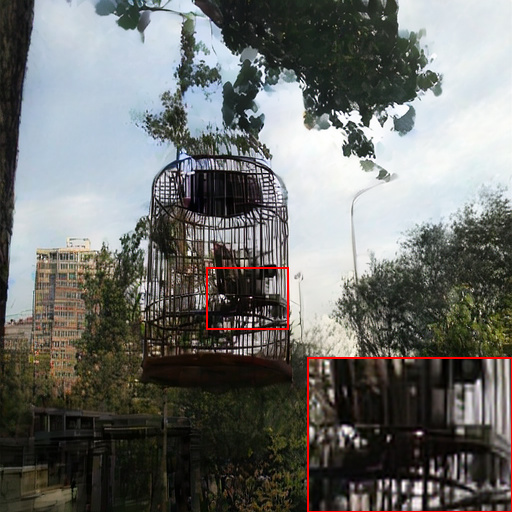} &
    \includegraphics[width=0.16\linewidth]{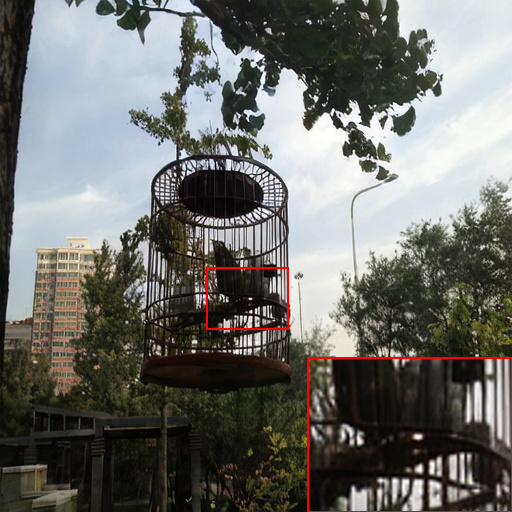} &
    \includegraphics[width=0.16\linewidth]{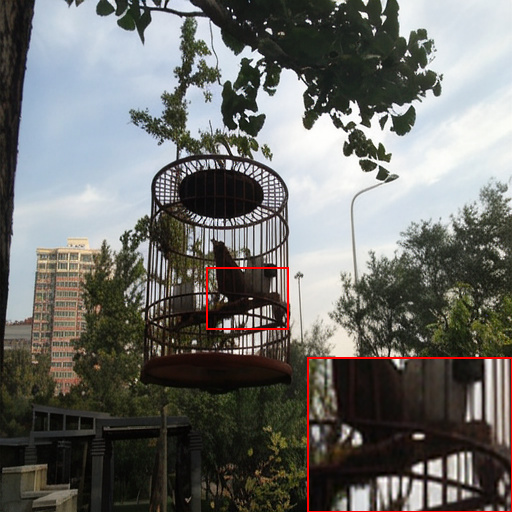} &
    \includegraphics[width=0.16\linewidth]{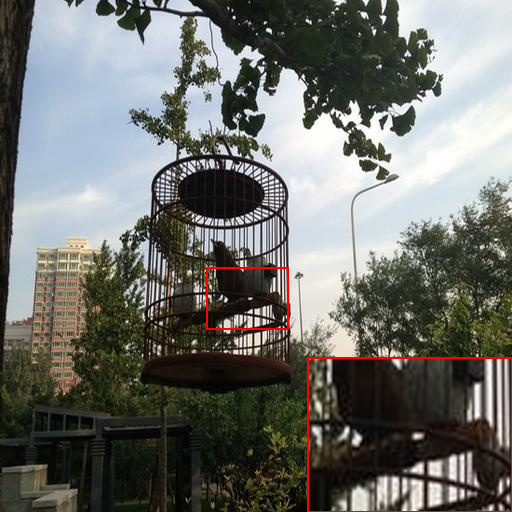} \\[4pt]
    \includegraphics[width=0.16\linewidth]{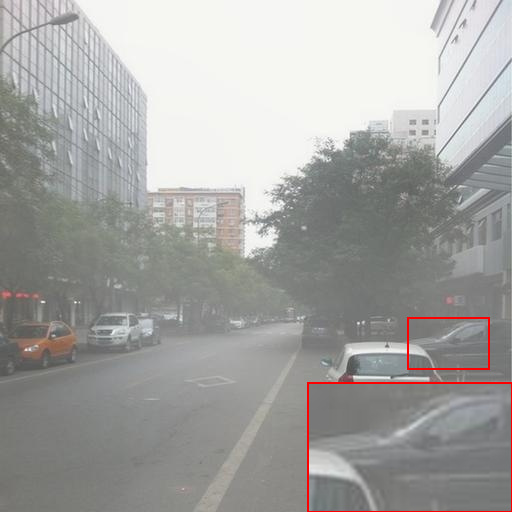} &
    \includegraphics[width=0.16\linewidth]{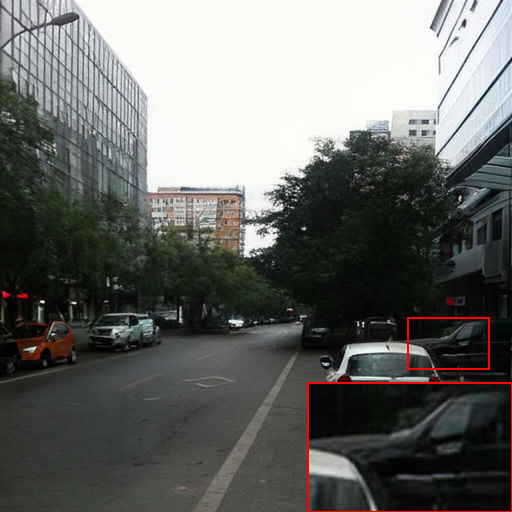} &
    \includegraphics[width=0.16\linewidth]{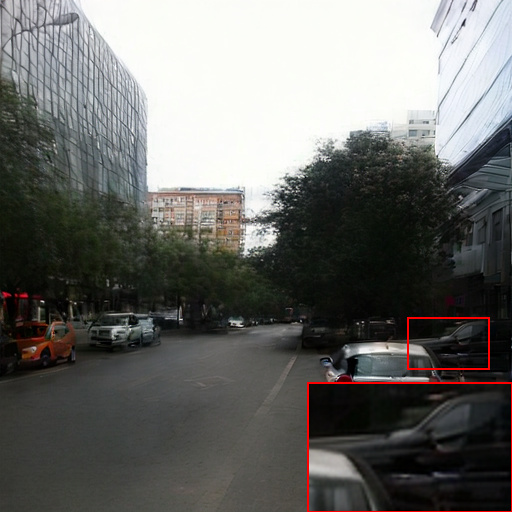} &
    \includegraphics[width=0.16\linewidth]{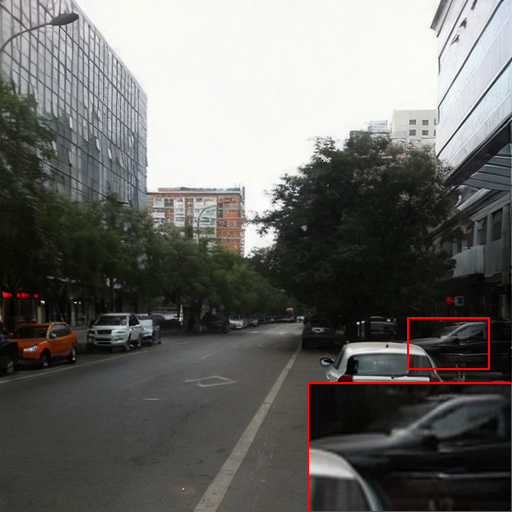} &
    \includegraphics[width=0.16\linewidth]{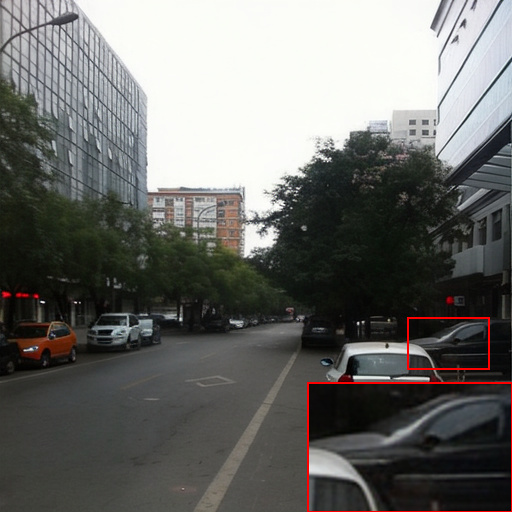} &
    \includegraphics[width=0.16\linewidth]{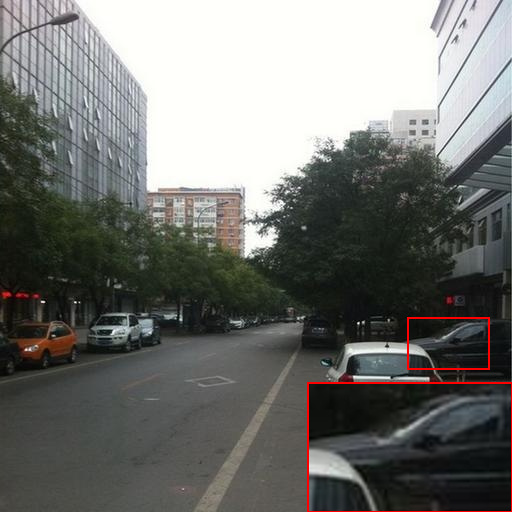} \\[2pt]
    \makebox[0.16\linewidth]{(a) Input} &
    \makebox[0.16\linewidth]{(b) No refiner} &
    \makebox[0.16\linewidth]{(c) w/o last block} &
    \makebox[0.16\linewidth]{(d) HART } &
    \makebox[0.16\linewidth]{(e) Our LRT} &
    \makebox[0.16\linewidth]{(f) GT} \\
  \end{tabular}
  \caption{Qualitative results for the ablation on latent refiner configurations. Our proposed LRT preserves fine details better than the other configurations.}
  \label{supfig:refiners}
\end{figure}

In Sec.~\ref{subsec: ablations}, we demonstrated that our proposed Latent Refiner Transformer (LRT) achieves the best performance compared to using no refiner, a refiner without last-block conditioning, and HART~\citep{hart}'s diffusion-based refiner. Quantitative results, reported in Table~\ref{tab:refiner_ablation}, included mean PSNR and SSIM scores on the RESIDE~\citep{reside} test set. Fig.~\ref{supfig:refiners} presents qualitative comparisons for these configurations. It can be observed that our LRT preserves fine details more effectively than the other variants.


\subsection{Additional Qualitative Comparisons}
\label{supsubsec: add_qual}

\begin{figure*}
    \centering
    \small
    \setlength{\tabcolsep}{1pt}
    \begin{tabular}{ccccccc}
    &Input&Diff-Plugin&AutoDIR&PixWizard&RestoreVAR&GT\\
          \rotatebox[origin=c]{90}{RESIDE\hspace{-43pt}}&\includegraphics[height=1.37cm, width=2.1cm]{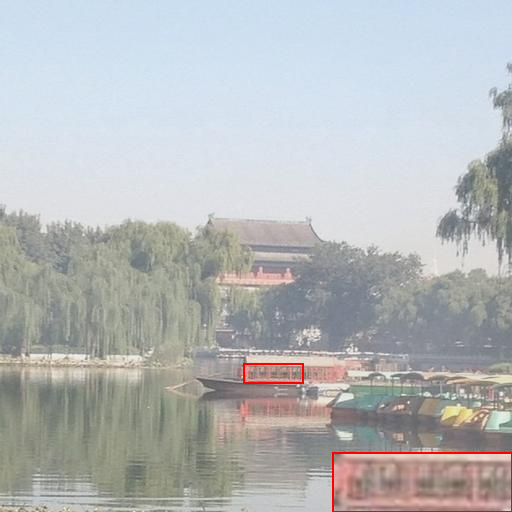}&\includegraphics[height=1.37cm, width=2.1cm]{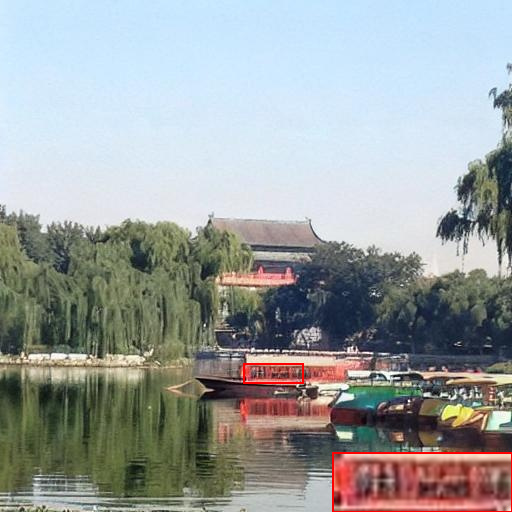}&\includegraphics[height=1.37cm, width=2.1cm]{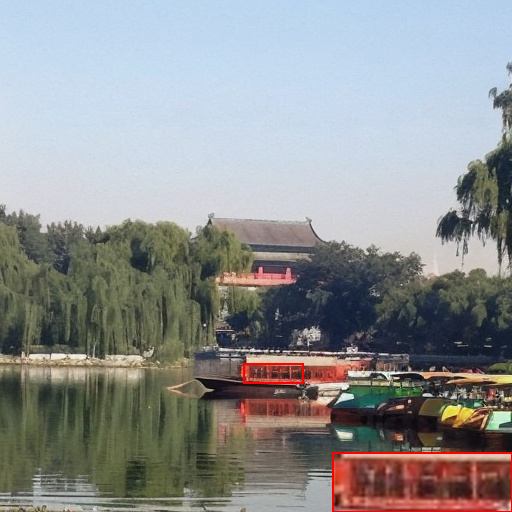}&\includegraphics[height=1.37cm, width=2.1cm]{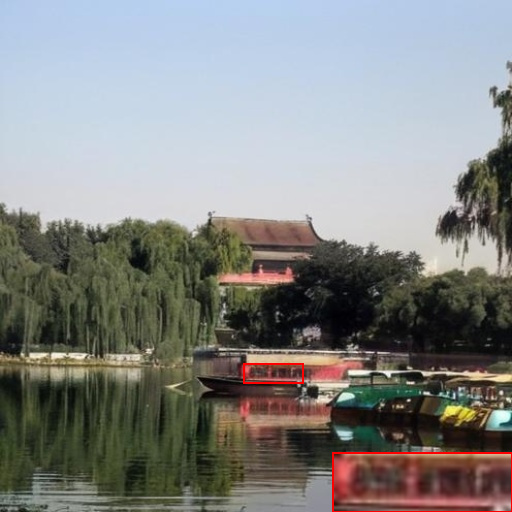}&\includegraphics[height=1.37cm, width=2.1cm]{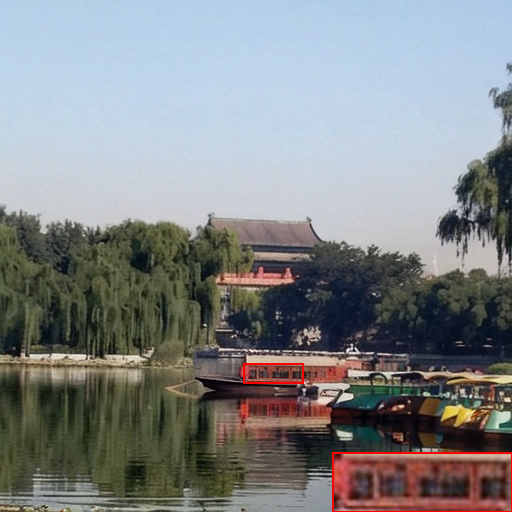}&\includegraphics[height=1.37cm, width=2.1cm]{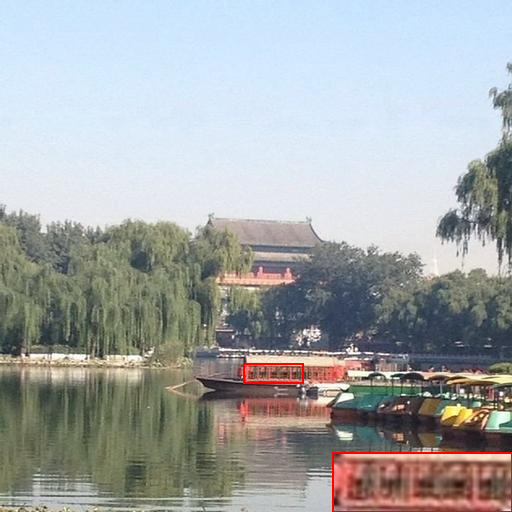}  \\
         
         \rotatebox[origin=c]{90}{Snow100k\hspace{-43pt}}&\includegraphics[height=1.37cm, width=2.1cm]{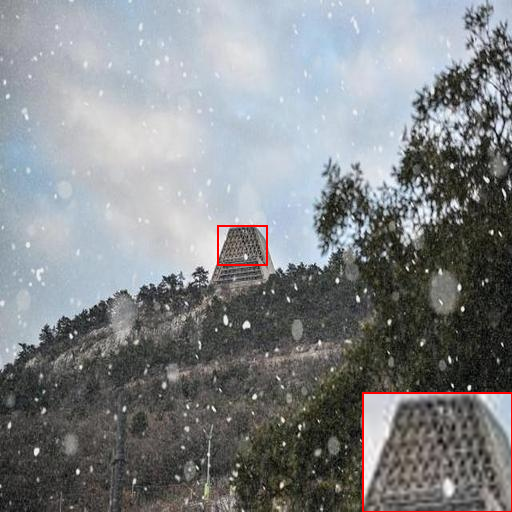}&\includegraphics[height=1.37cm, width=2.1cm]{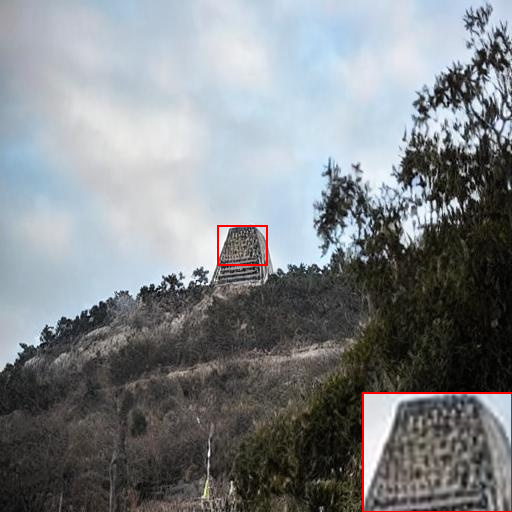}&\includegraphics[height=1.37cm, width=2.1cm]{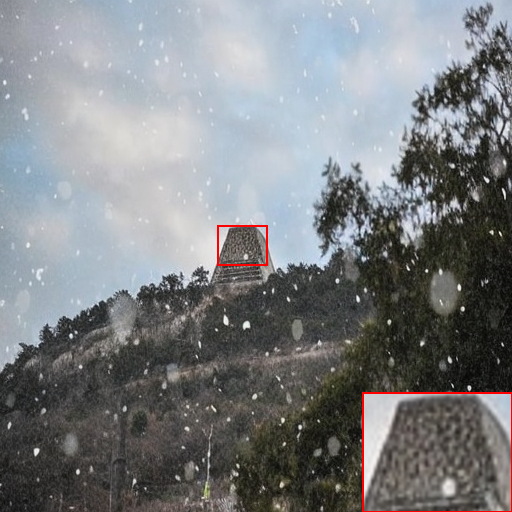}&\includegraphics[height=1.37cm, width=2.1cm]{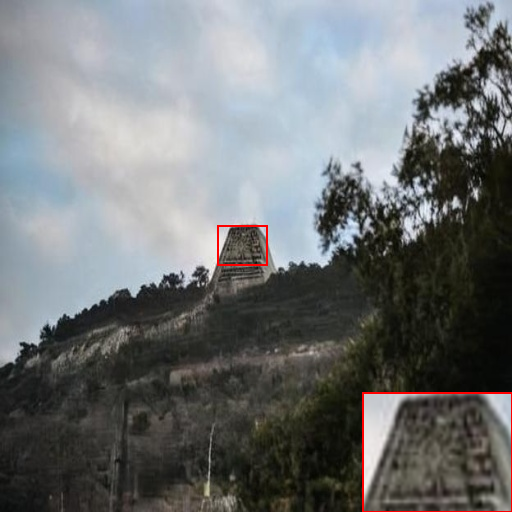}&\includegraphics[height=1.37cm, width=2.1cm]{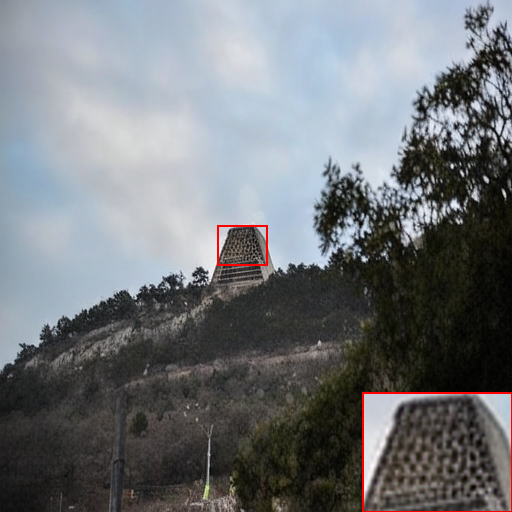}&\includegraphics[height=1.37cm, width=2.1cm]{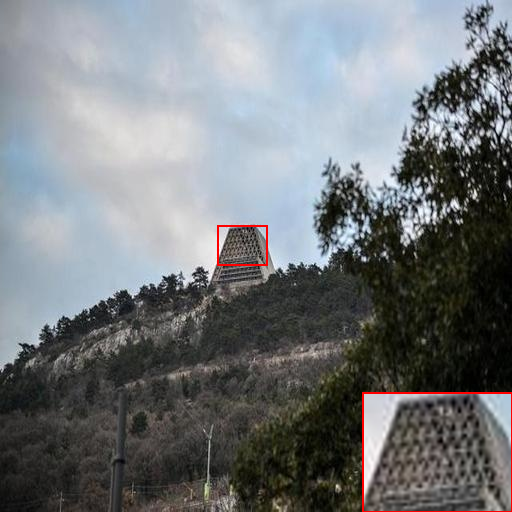}  \\

         \rotatebox[origin=c]{90}{Snow100k\hspace{-43pt}}&\includegraphics[height=1.37cm, width=2.1cm]{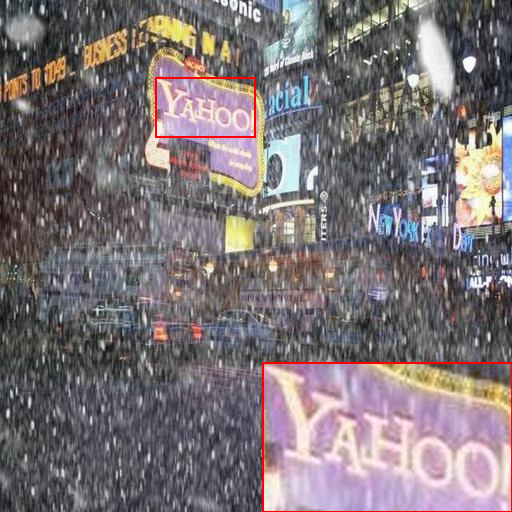}&\includegraphics[height=1.37cm, width=2.1cm]{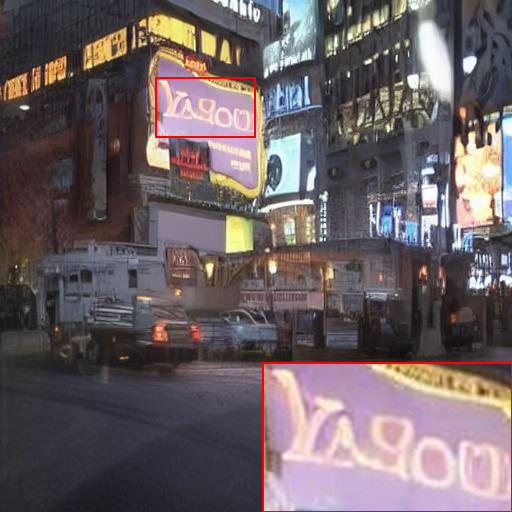}&\includegraphics[height=1.37cm, width=2.1cm]{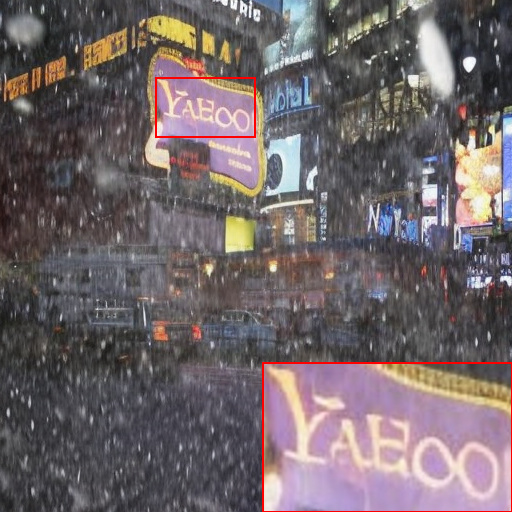}&\includegraphics[height=1.37cm, width=2.1cm]{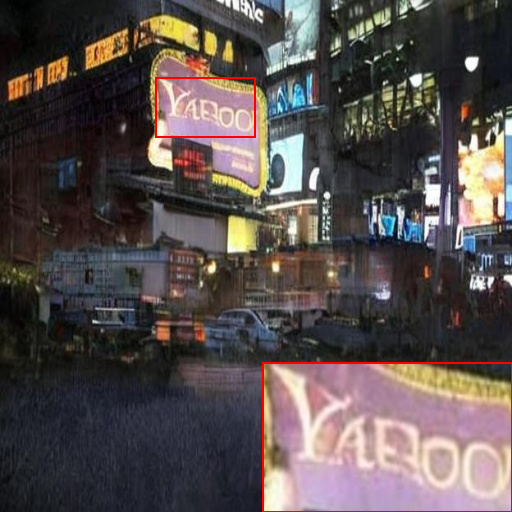}&\includegraphics[height=1.37cm, width=2.1cm]{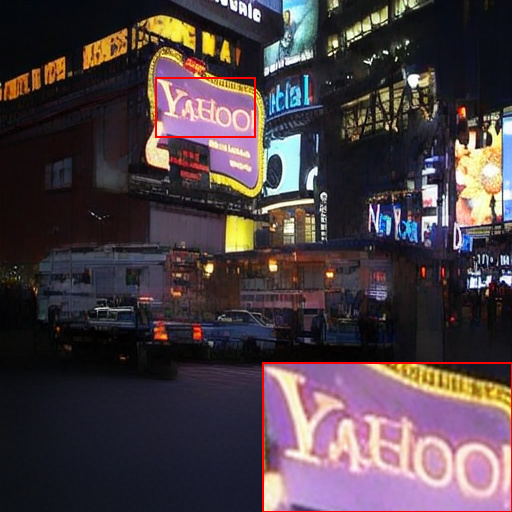}&\includegraphics[height=1.37cm, width=2.1cm]{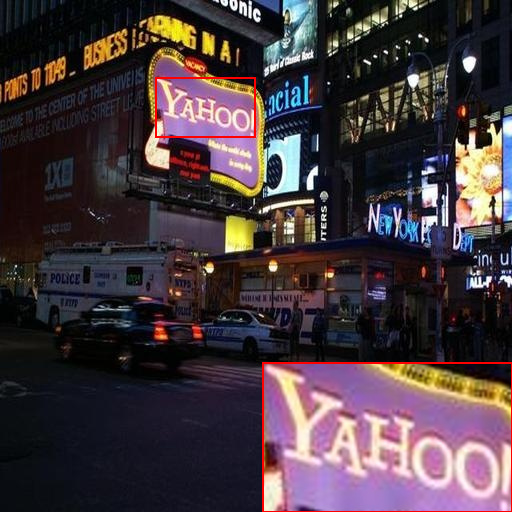}  \\

         \rotatebox[origin=c]{90}{Rain13K\hspace{-43pt}}&\includegraphics[height=1.37cm, width=2.1cm]{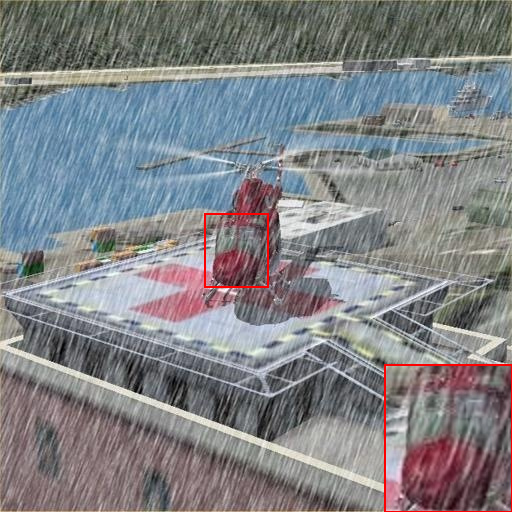}&\includegraphics[height=1.37cm, width=2.1cm]{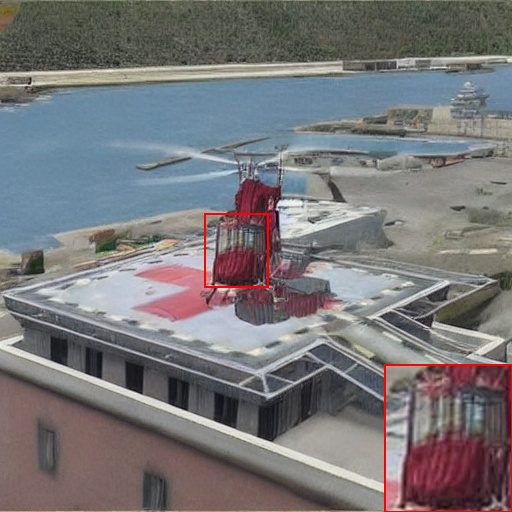}&\includegraphics[height=1.37cm, width=2.1cm]{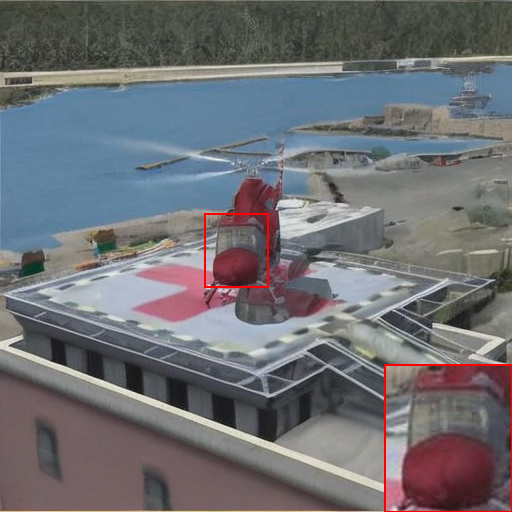}&\includegraphics[height=1.37cm, width=2.1cm]{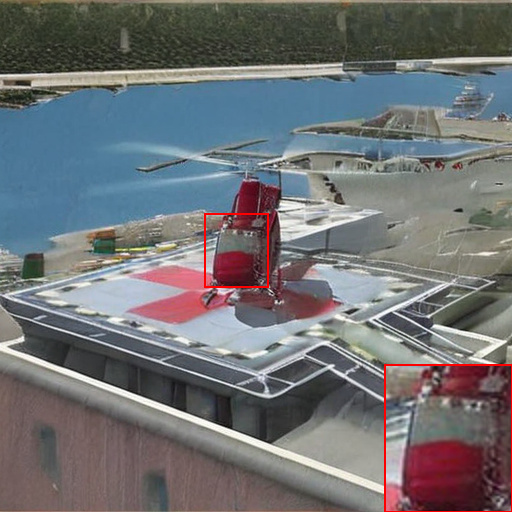}&\includegraphics[height=1.37cm, width=2.1cm]{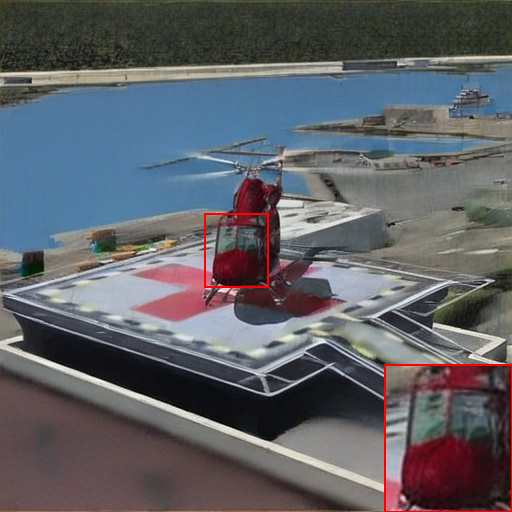}&\includegraphics[height=1.37cm, width=2.1cm]{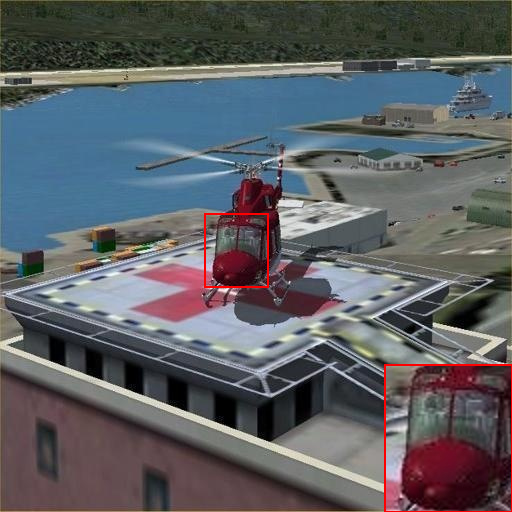}  \\

         \rotatebox[origin=c]{90}{LOLv1\hspace{-43pt}}&\includegraphics[height=1.37cm, width=2.1cm]{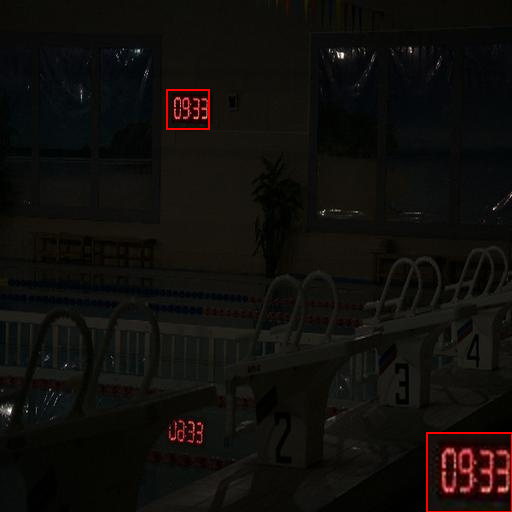}&\includegraphics[height=1.37cm, width=2.1cm]{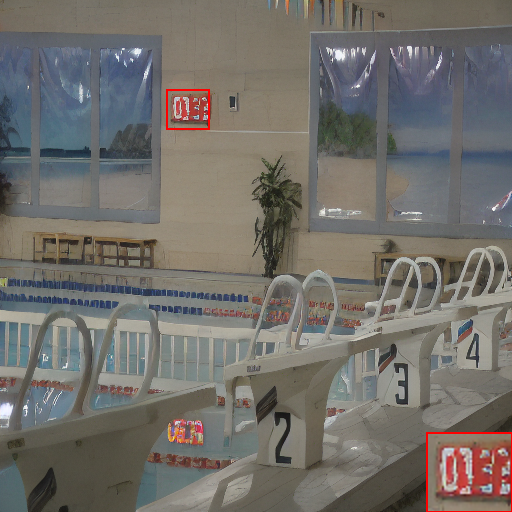}&\includegraphics[height=1.37cm, width=2.1cm]{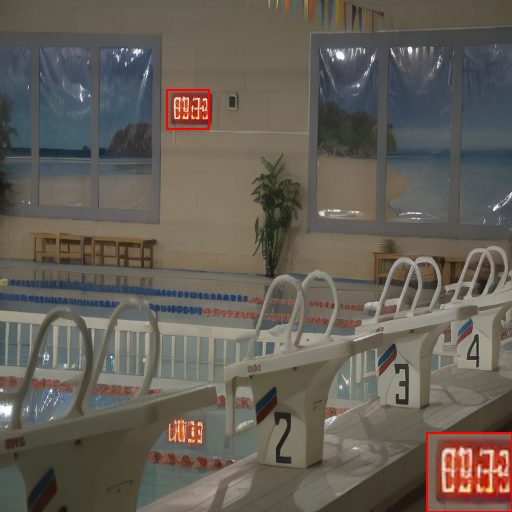}&\includegraphics[height=1.37cm, width=2.1cm]{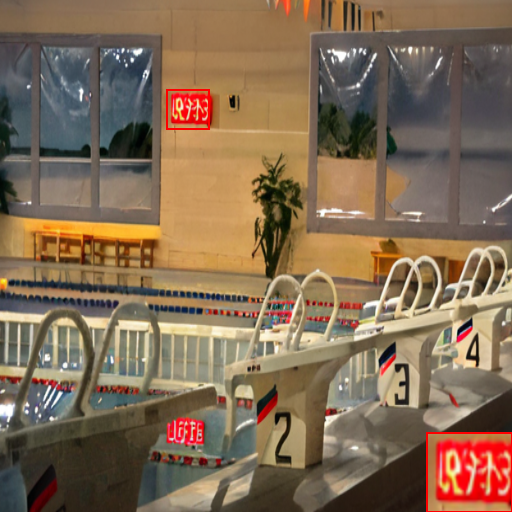}&\includegraphics[height=1.37cm, width=2.1cm]{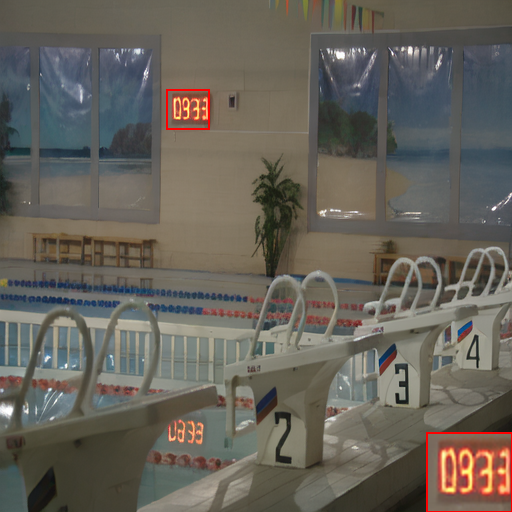}&\includegraphics[height=1.37cm, width=2.1cm]{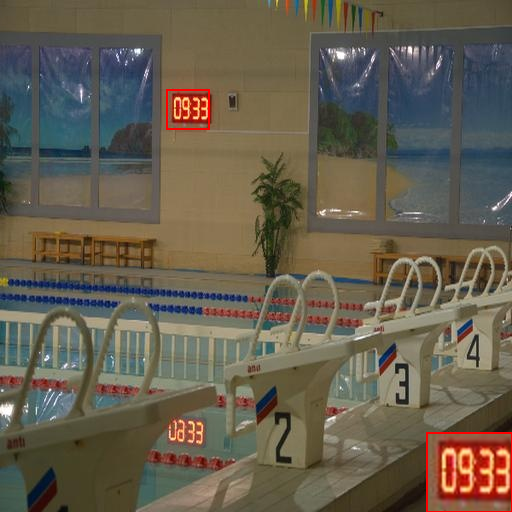}  \\

         \rotatebox[origin=c]{90}{LOLv1\hspace{-43pt}}&\includegraphics[height=1.37cm, width=2.1cm]{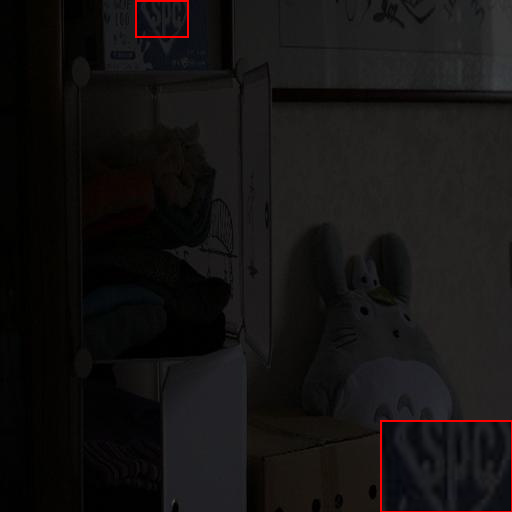}&\includegraphics[height=1.37cm, width=2.1cm]{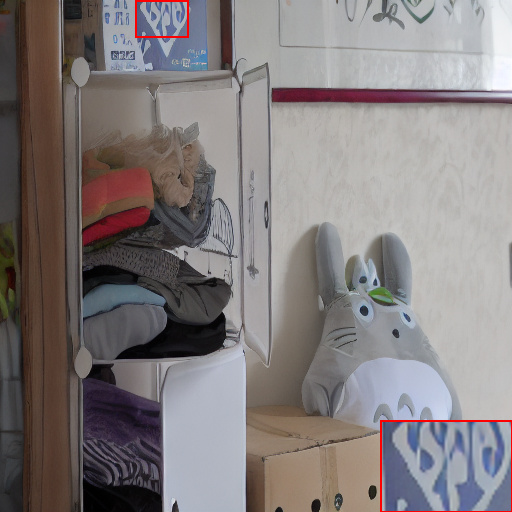}&\includegraphics[height=1.37cm, width=2.1cm]{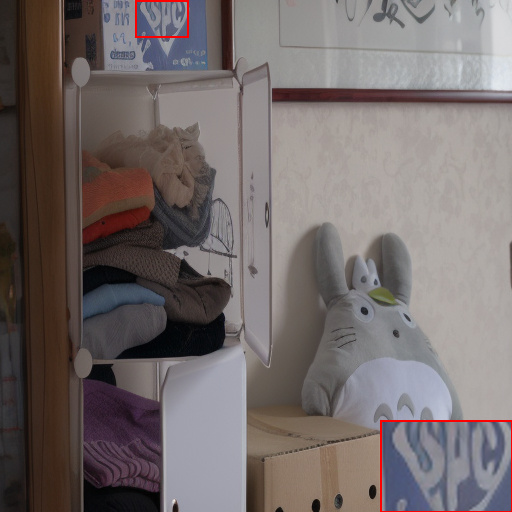}&\includegraphics[height=1.37cm, width=2.1cm]{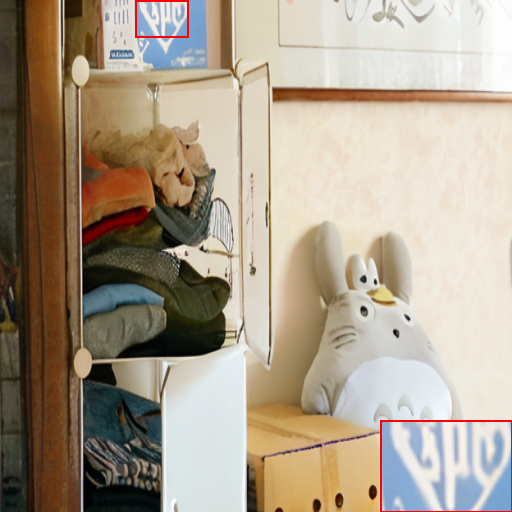}&\includegraphics[height=1.37cm, width=2.1cm]{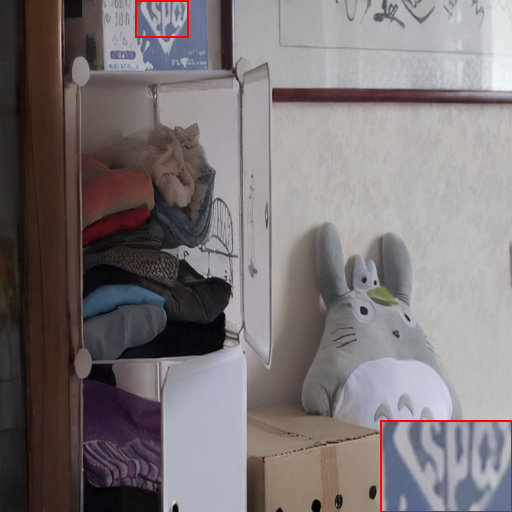}&\includegraphics[height=1.37cm, width=2.1cm]{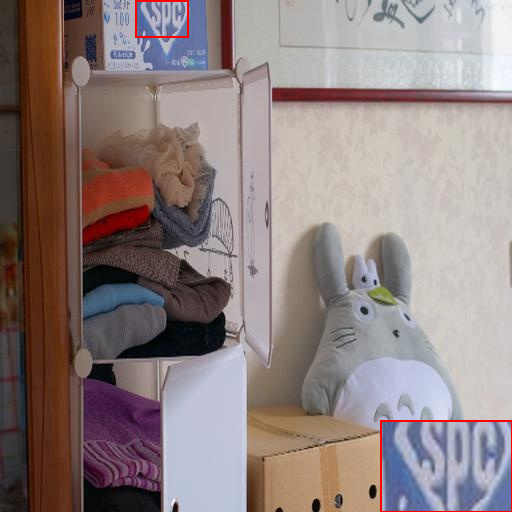}  \\

         \rotatebox[origin=c]{90}{GoPro\hspace{-43pt}}&\includegraphics[height=1.37cm, width=2.1cm]{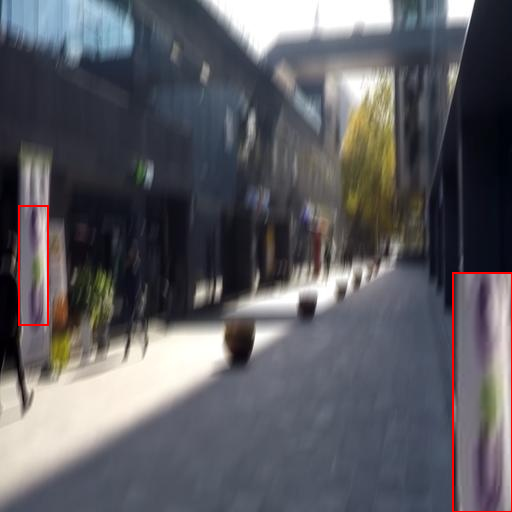}&\includegraphics[height=1.37cm, width=2.1cm]{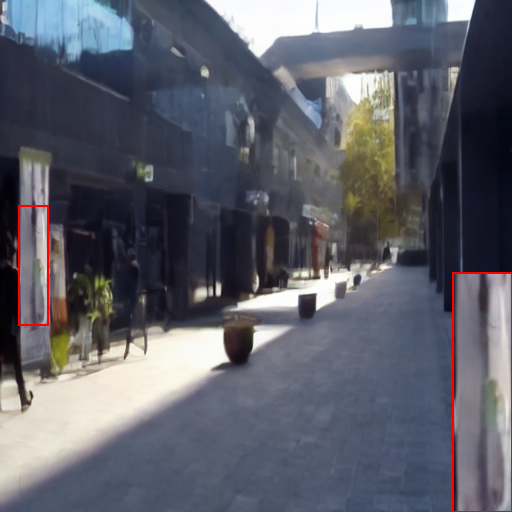}&\includegraphics[height=1.37cm, width=2.1cm]{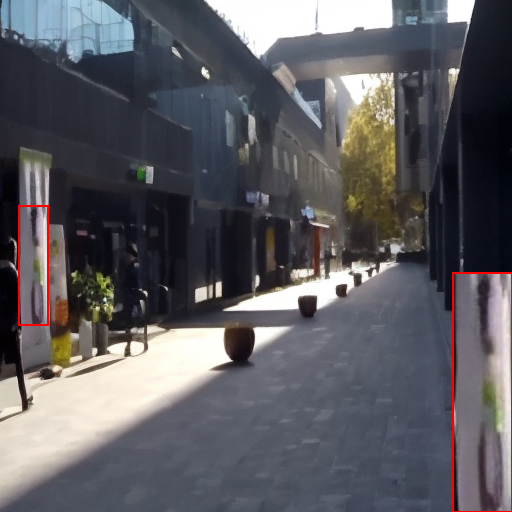}&\includegraphics[height=1.37cm, width=2.1cm]{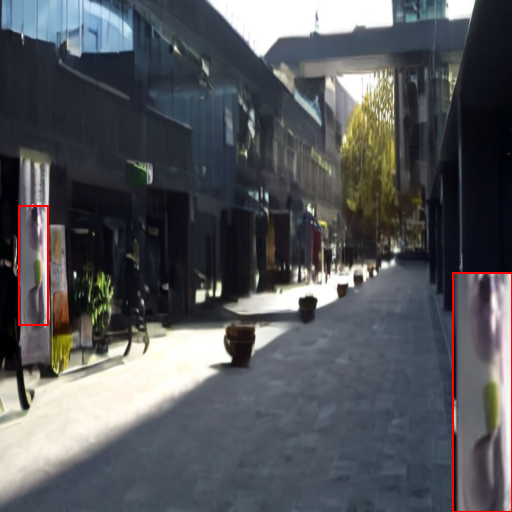}&\includegraphics[height=1.37cm, width=2.1cm]{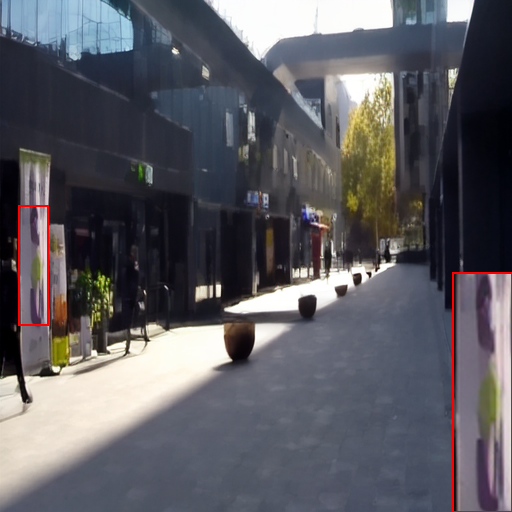}&\includegraphics[height=1.37cm, width=2.1cm]{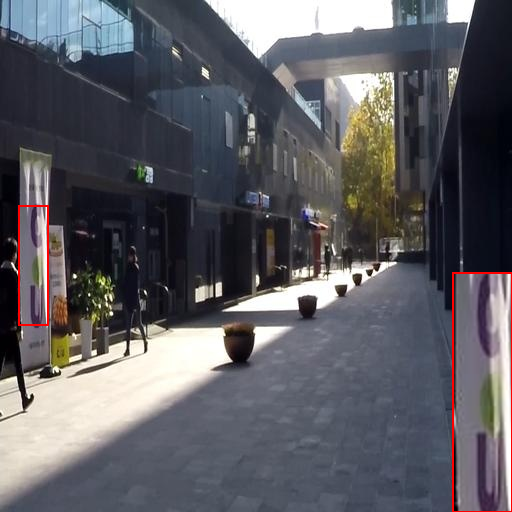}
    \end{tabular}
    \vspace{-5pt}
    \caption{Additional qualitative comparisons of RestoreVAR with LDM-based AiOR approaches. RestoreVAR consistently preserves fine details more effectively than the LDM-based methods.}
    \label{supfig: main_qual}
    \vspace{-12pt}
\end{figure*}

In this section, we provide additional qualitative comparisons of RestoreVAR with state-of-the-art LDM-based and non-generative all-in-one image restoration (AiOR) methods. Fig.~\ref{supfig: main_qual} presents results from RestoreVAR alongside Diff-Plugin\citep{diffplugin}, AutoDIR~\citep{autodir}, and PixWizard~\citep{pixwizard} on the RESIDE~\citep{reside}, Snow100k~\citep{snow100k}, Rain13K~\citep{mprnet}, LOLv1~\citep{lolv1}, and GoPro~\citep{gopro} datasets. RestoreVAR consistently produces outputs that are more semantically aligned with the ground truth (see zoomed-in patches).

\begin{figure*}[t]
    \centering
    \small
    \setlength{\tabcolsep}{1pt}
    \begin{tabular}{c cccccc}
        & LHP & REVIDE & TOLED & POLED & CDD (H $+$ R) \\
        
        \rotatebox[origin=c]{90}{Input\hspace{-33.5pt}} &
        \includegraphics[height=1.37cm, width=1.88cm]{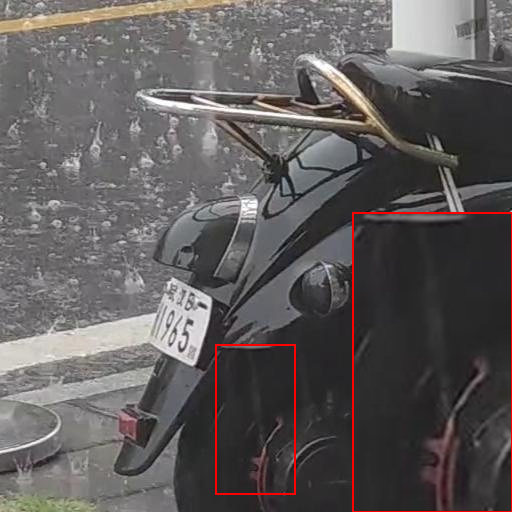} &
        \includegraphics[height=1.37cm, width=1.88cm]{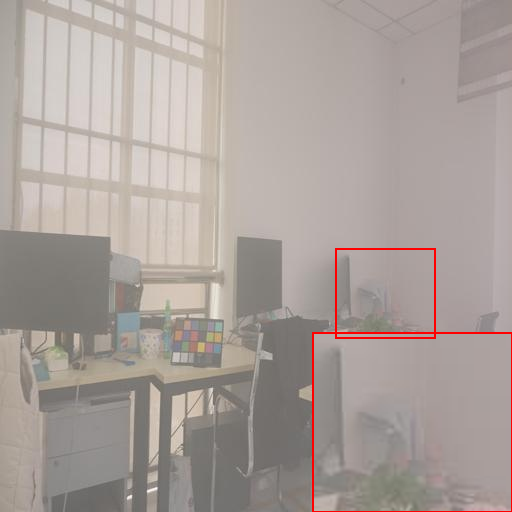} &
        \includegraphics[height=1.37cm, width=1.88cm]{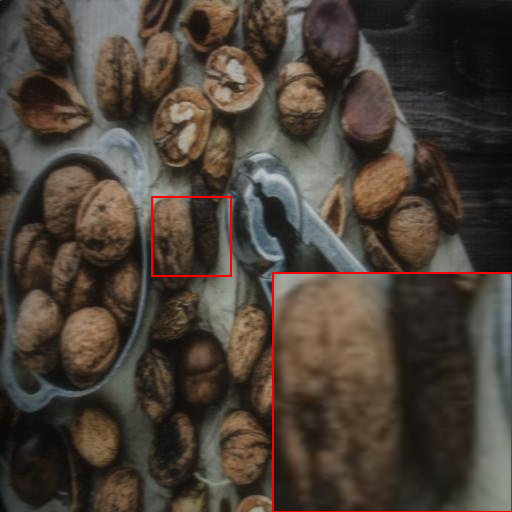} &
        \includegraphics[height=1.37cm, width=1.88cm]{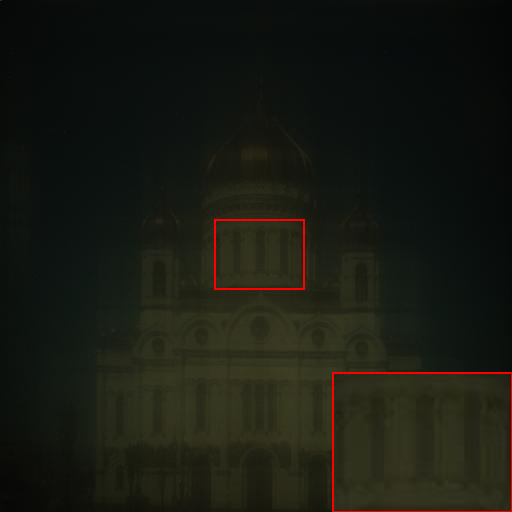} &
        \includegraphics[height=1.37cm, width=1.88cm]{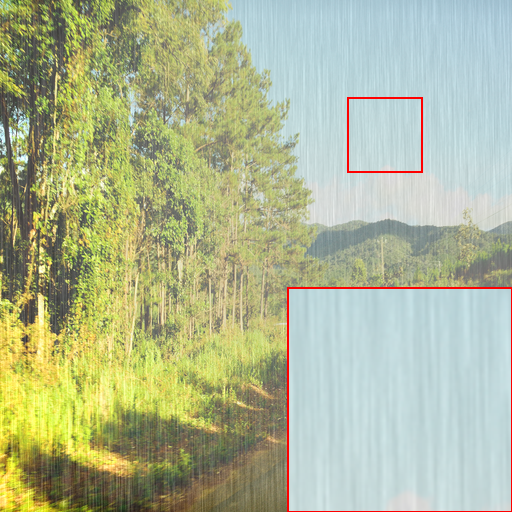}\\
        
        \rotatebox[origin=c]{90}{PromptIR\hspace{-33.5pt}} &
        \includegraphics[height=1.37cm, width=1.88cm]{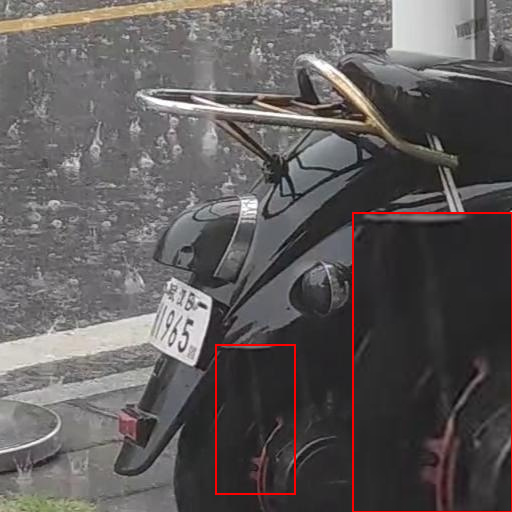} &
        \includegraphics[height=1.37cm, width=1.88cm]{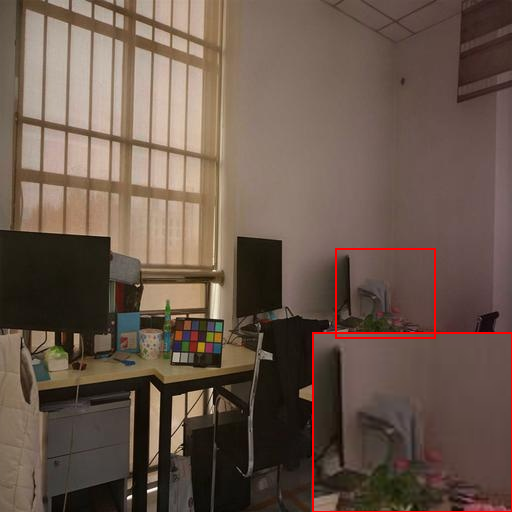} &
        \includegraphics[height=1.37cm, width=1.88cm]{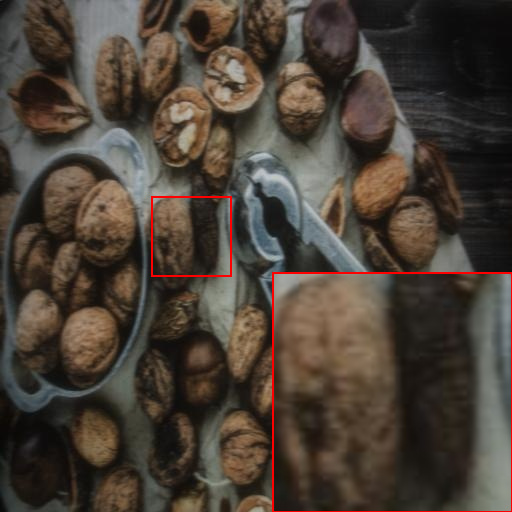} &
        \includegraphics[height=1.37cm, width=1.88cm]{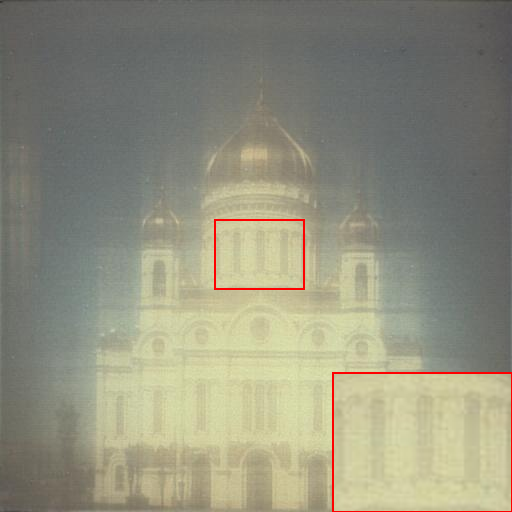} &
        \includegraphics[height=1.37cm, width=1.88cm]{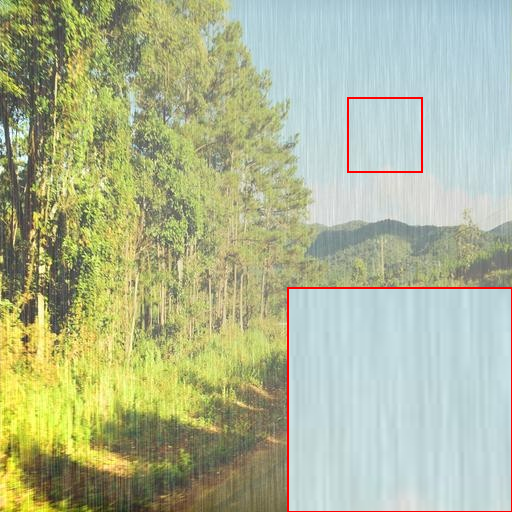}\\
        
        \rotatebox[origin=c]{90}{InstructIR\hspace{-33.5pt}} &
        \includegraphics[height=1.37cm, width=1.88cm]{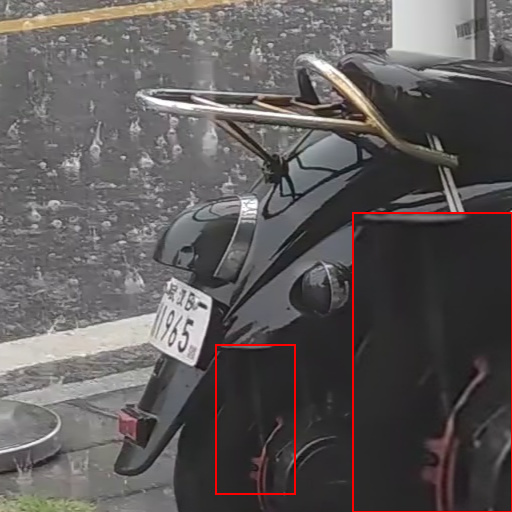} &
        \includegraphics[height=1.37cm, width=1.88cm]{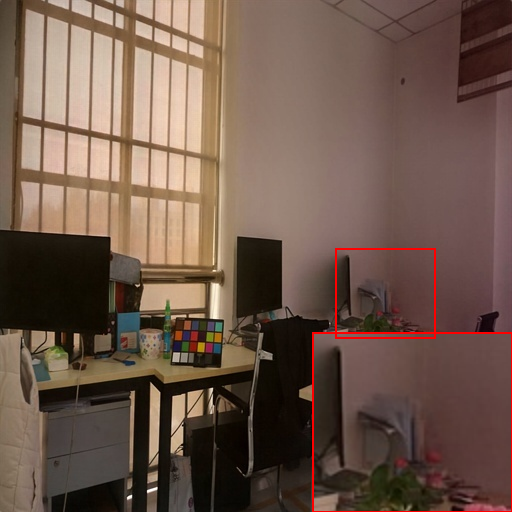} &
        \includegraphics[height=1.37cm, width=1.88cm]{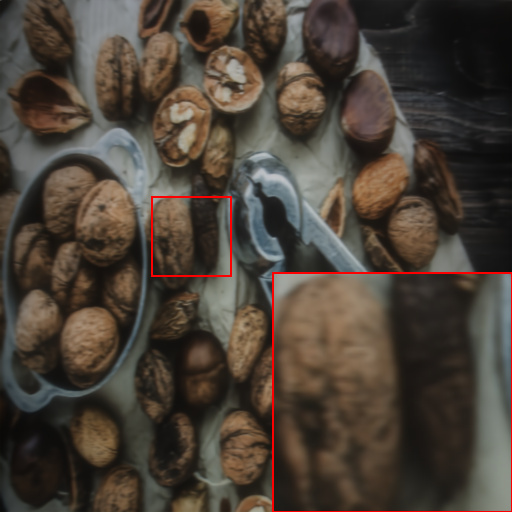} &
        \includegraphics[height=1.37cm, width=1.88cm]{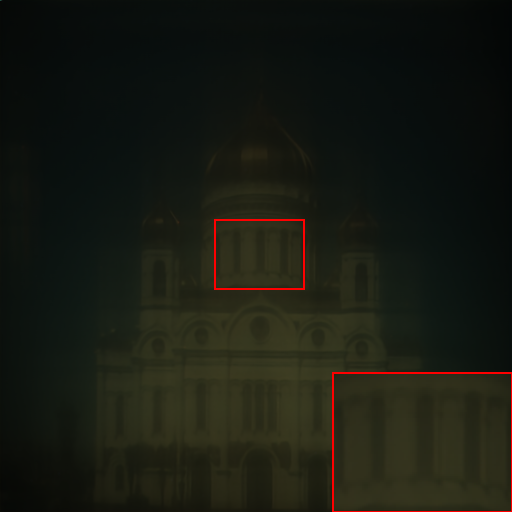} &
        \includegraphics[height=1.37cm, width=1.88cm]{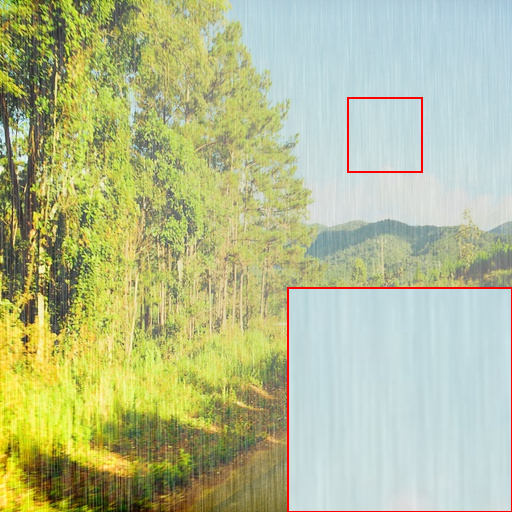}\\
        
        \rotatebox[origin=c]{90}{AWRaCLe\hspace{-33.5pt}} &
        \includegraphics[height=1.37cm, width=1.88cm]{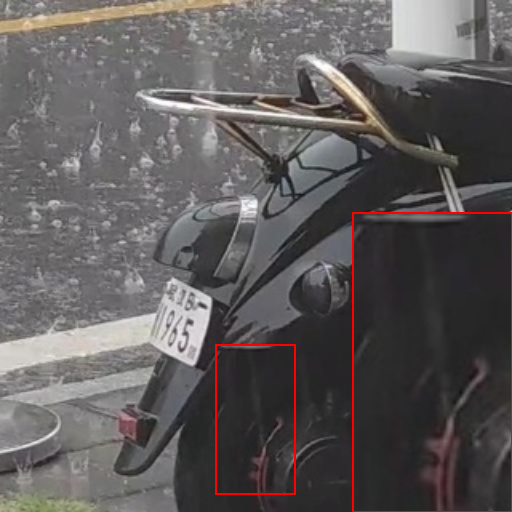} &
        \includegraphics[height=1.37cm, width=1.88cm]{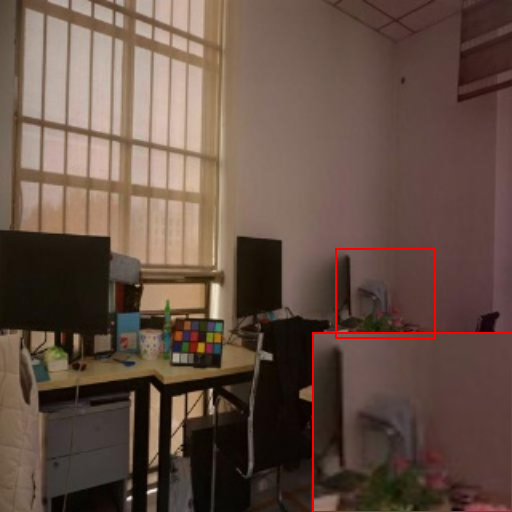} &
        \includegraphics[height=1.37cm, width=1.88cm]{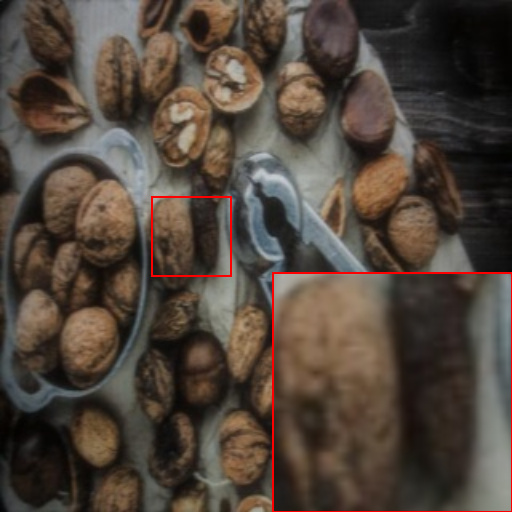} &
        \includegraphics[height=1.37cm, width=1.88cm]{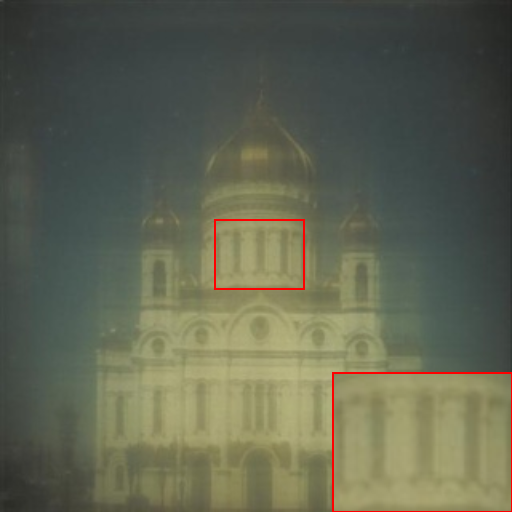} &
        \includegraphics[height=1.37cm, width=1.88cm]{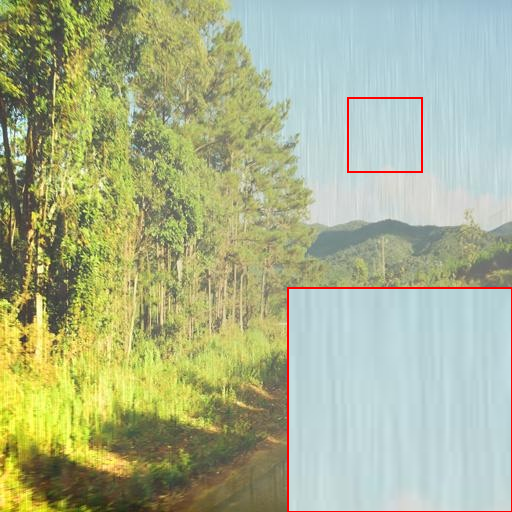}\\
        
        \rotatebox[origin=c]{90}{DCPT\hspace{-33.5pt}} &
        \includegraphics[height=1.37cm, width=1.88cm]{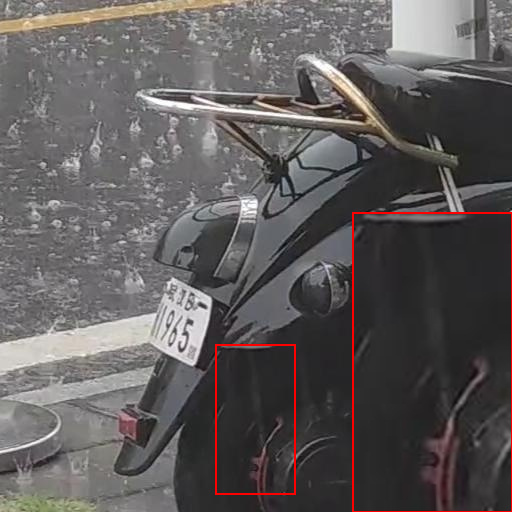} &
        \includegraphics[height=1.37cm, width=1.88cm]{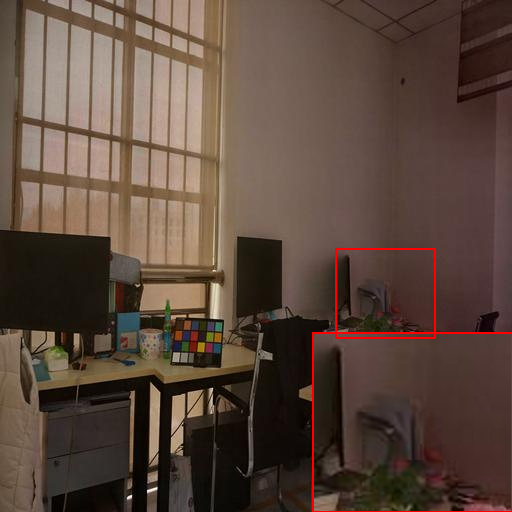} &
        \includegraphics[height=1.37cm, width=1.88cm]{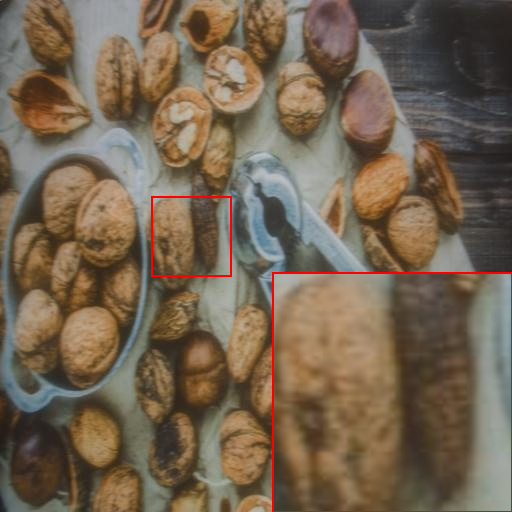} &
        \includegraphics[height=1.37cm, width=1.88cm]{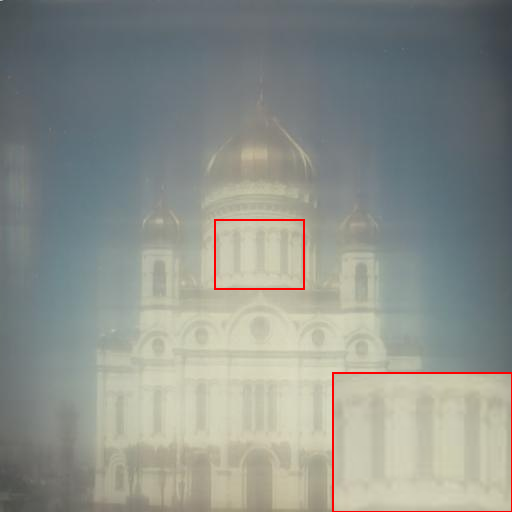} &
        \includegraphics[height=1.37cm, width=1.88cm]{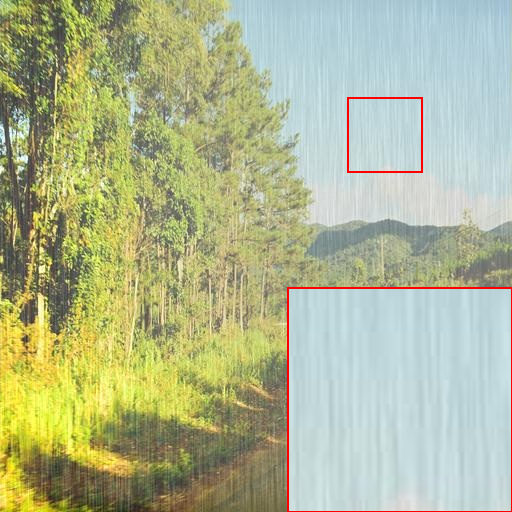}\\

        \rotatebox[origin=c]{90}{DFPIR\hspace{-33.5pt}} &
        \includegraphics[height=1.37cm, width=1.88cm]{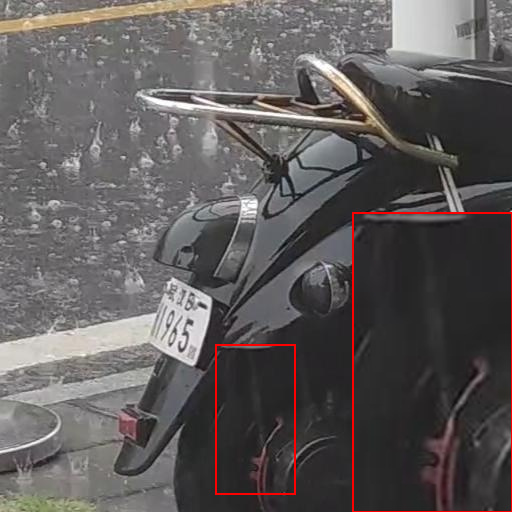} &
        \includegraphics[height=1.37cm, width=1.88cm]{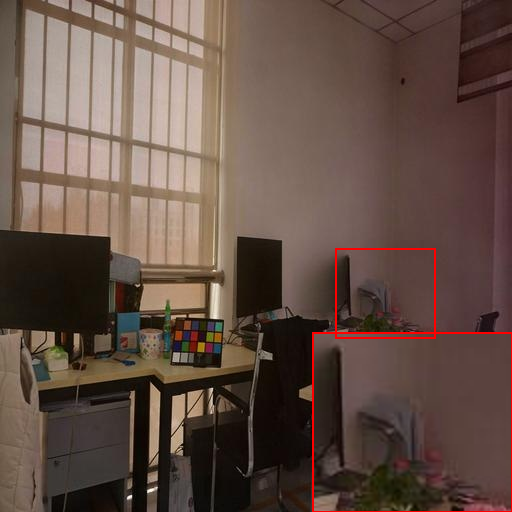} &
        \includegraphics[height=1.37cm, width=1.88cm]{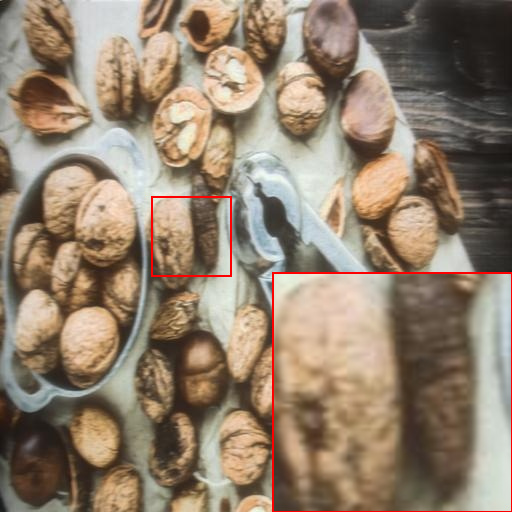} &
        \includegraphics[height=1.37cm, width=1.88cm]{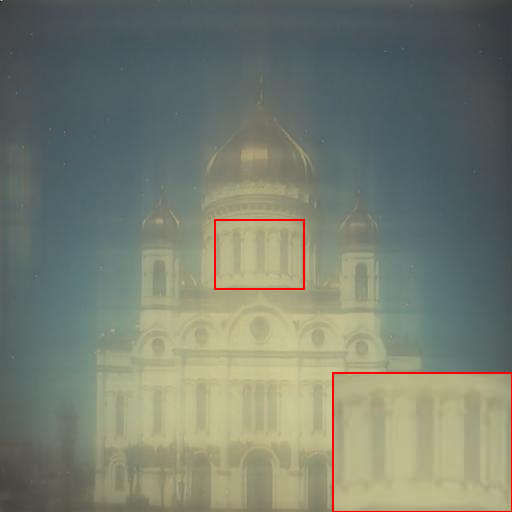} &
        \includegraphics[height=1.37cm, width=1.88cm]{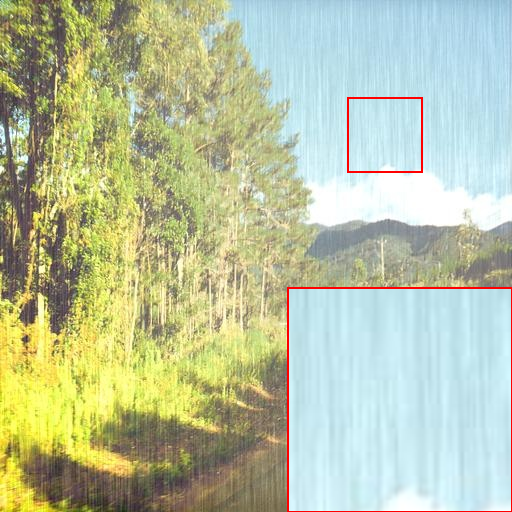}\\
        
        \rotatebox[origin=c]{90}{RestoreVAR\hspace{-33.5pt}} &
        \includegraphics[height=1.37cm, width=1.88cm]{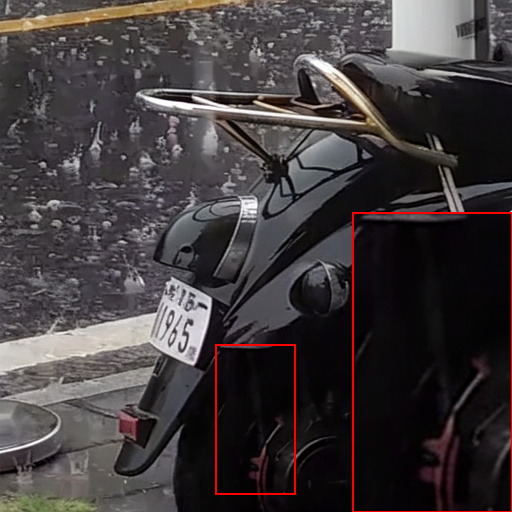} &
        \includegraphics[height=1.37cm, width=1.88cm]{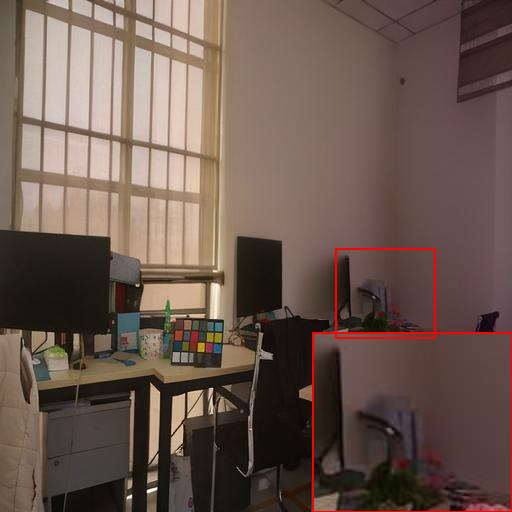} &
        \includegraphics[height=1.37cm, width=1.88cm]{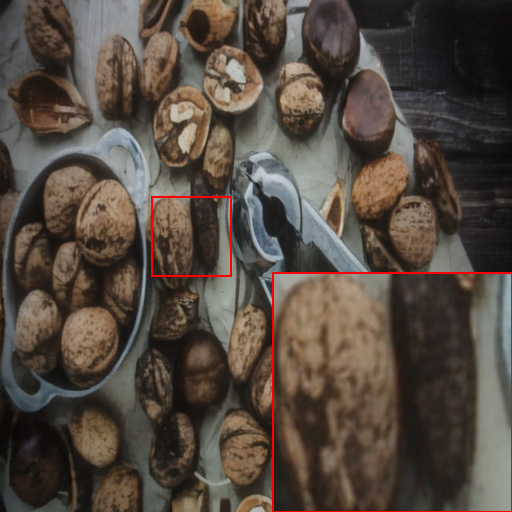} &
        \includegraphics[height=1.37cm, width=1.88cm]{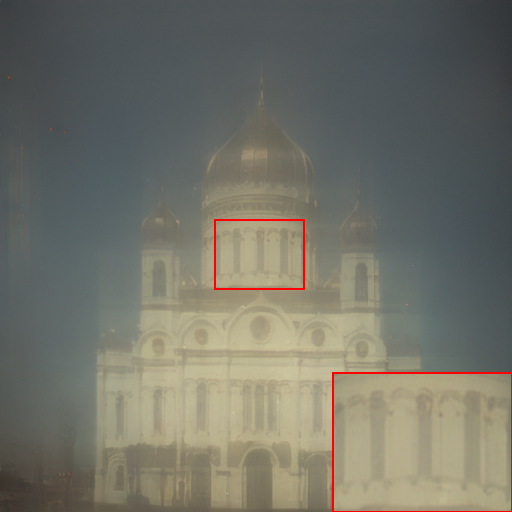} &
        \includegraphics[height=1.37cm, width=1.88cm]{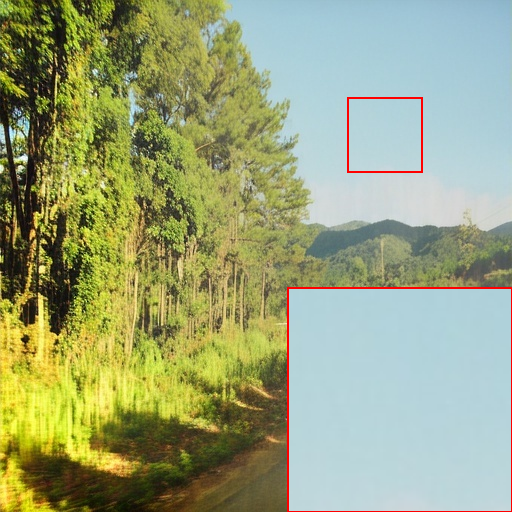} \\
        
        \rotatebox[origin=c]{90}{GT\hspace{-33.5pt}} &
        \includegraphics[height=1.37cm, width=1.88cm]{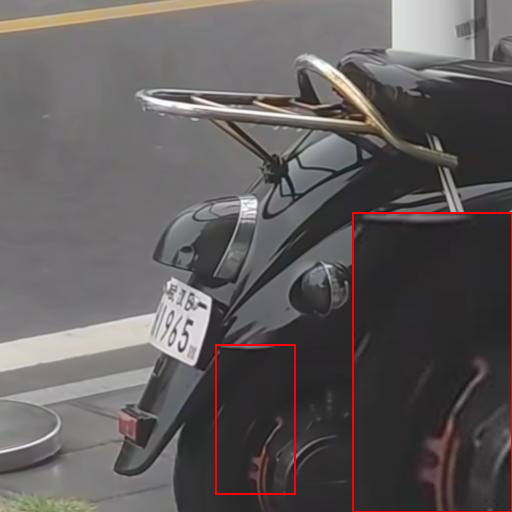} &
        \includegraphics[height=1.37cm, width=1.88cm]{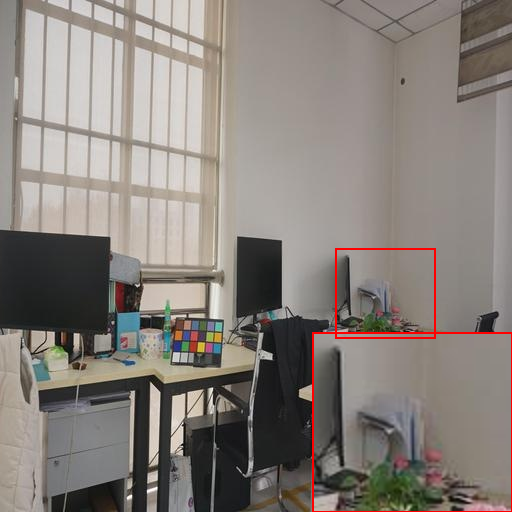} &
        \includegraphics[height=1.37cm, width=1.88cm]{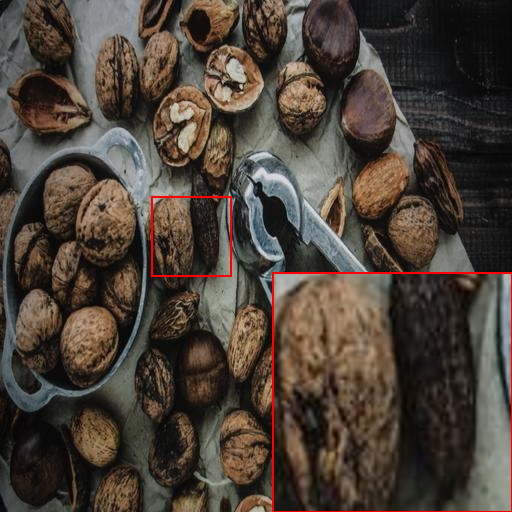} &
        \includegraphics[height=1.37cm, width=1.88cm]{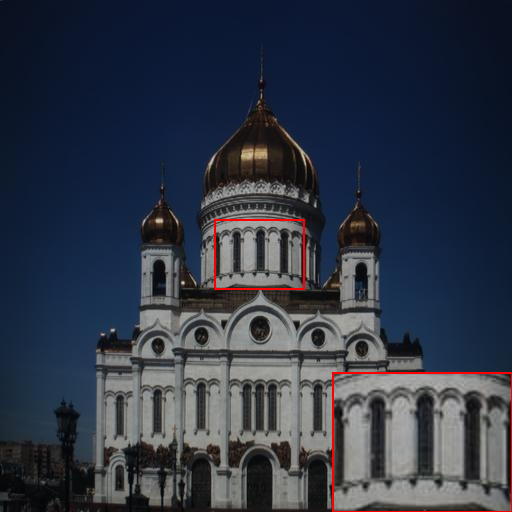}&\includegraphics[height=1.37cm, width=1.88cm]{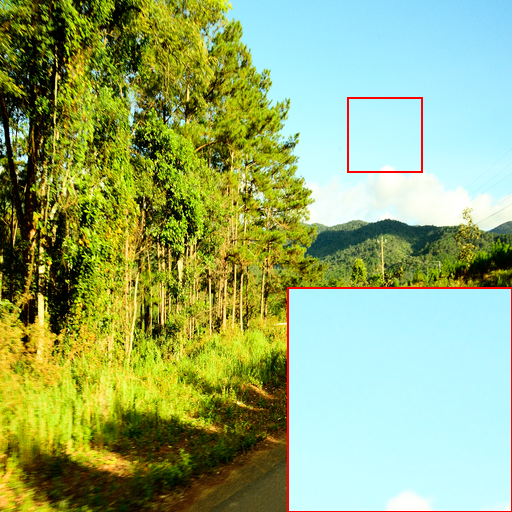} \\

    \end{tabular}
    \vspace{-5pt}
    \caption{Additional qualitative comparisons of RestoreVAR with non-generative methods on real-world, unseen and mixed degradations. RestoreVAR achieves better results, highlighting its superior generalization.}
    \label{supfig: ood_qual}
    \vspace{-8pt}
\end{figure*}

Fig.~\ref{supfig: ood_qual} provides comparisons with non-generative methods—PromptIR\citep{promptir}, InstructIR~\citep{instructir}, AWRaCLe~\citep{awracle}, DCPT~\citep{dcpt} and DFPIR~\citep{dfpir}—for real-world, mixed and unseen degradation generalization on LHP~\citep{lhprain}, REVIDE~\citep{revide}, TOLED~\citep{toledpoled}, POLED~\citep{toledpoled}, CDD~\citep{cdd} and LOLBlur~\citep{lolblur} datasets. RestoreVAR generates sharper, more realistic outputs with fewer artifacts than non-generative models. For instance, for the TOLED and POLED cases, RestoreVAR outputs are visibly sharper than non-generative methods. Similarly, the results of RestoreVAR are superior in the case of mixed degradations.

\begin{figure*}[t]
    \centering
    \small
    \setlength{\tabcolsep}{1pt}
    \begin{tabular}{cccccccc}
        & Input & PromptIR & InstructIR & AWRaCLe & DCPT & DFPIR & RestoreVAR \\

        \rotatebox[origin=c]{90}{LOLBlur\hspace{-33pt}}&\includegraphics[height=1.37cm, width=1.88cm]{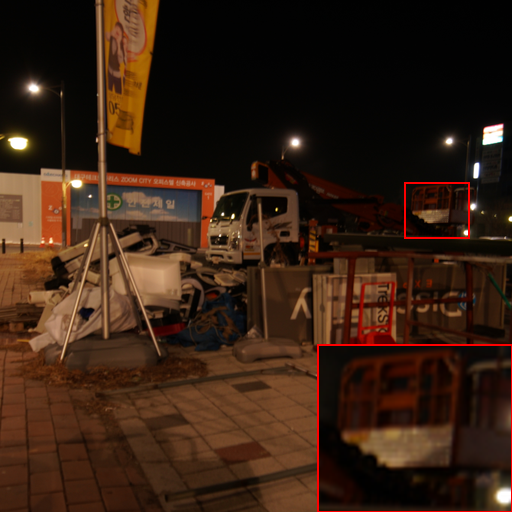}&\includegraphics[height=1.37cm, width=1.88cm]{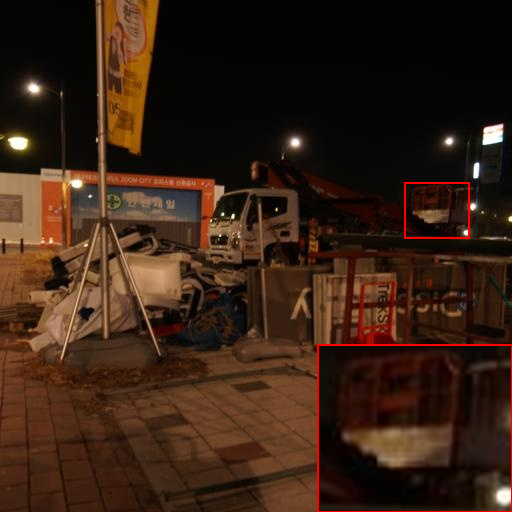}&\includegraphics[height=1.37cm, width=1.88cm]{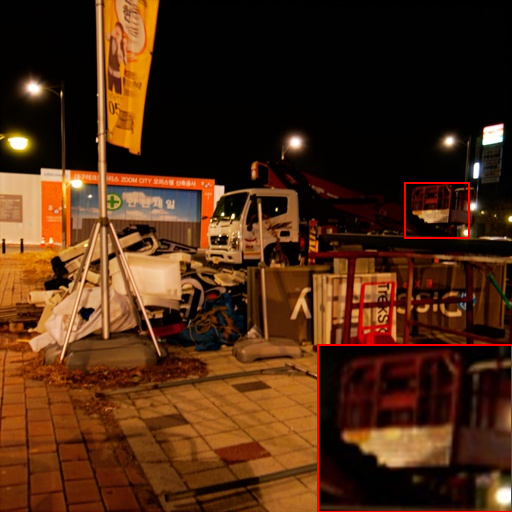}&\includegraphics[height=1.37cm, width=1.88cm]{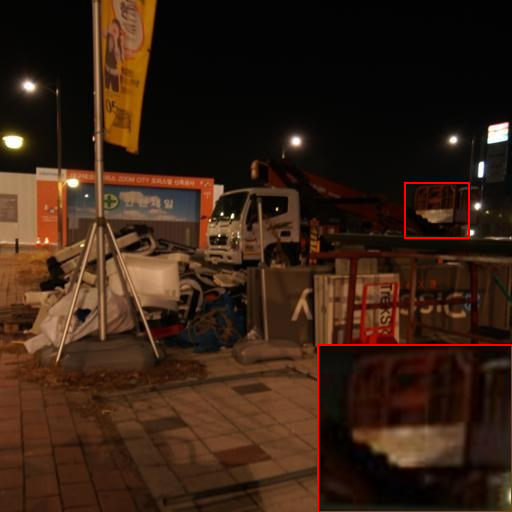}&\includegraphics[height=1.37cm, width=1.88cm]{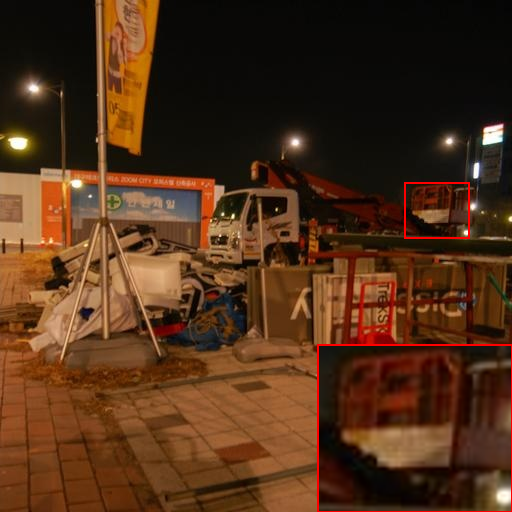}&\includegraphics[height=1.37cm, width=1.88cm]{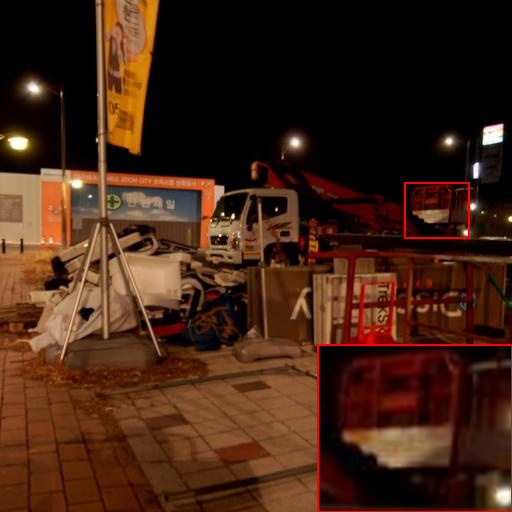}&\includegraphics[height=1.37cm, width=1.88cm]{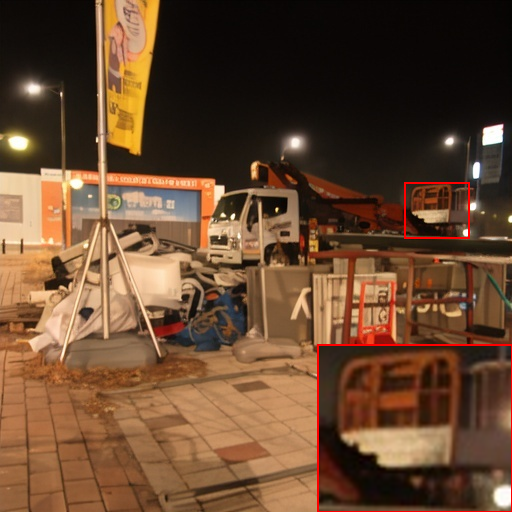}\\

        \rotatebox[origin=c]{90}{LOLBlur\hspace{-33pt}} &
        \includegraphics[height=1.37cm, width=1.88cm]{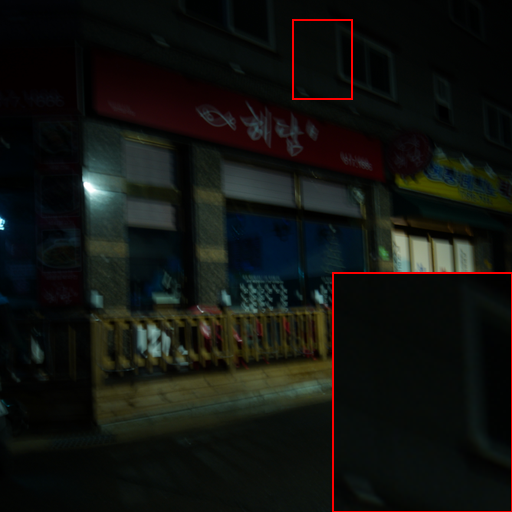} &
        \includegraphics[height=1.37cm, width=1.88cm]{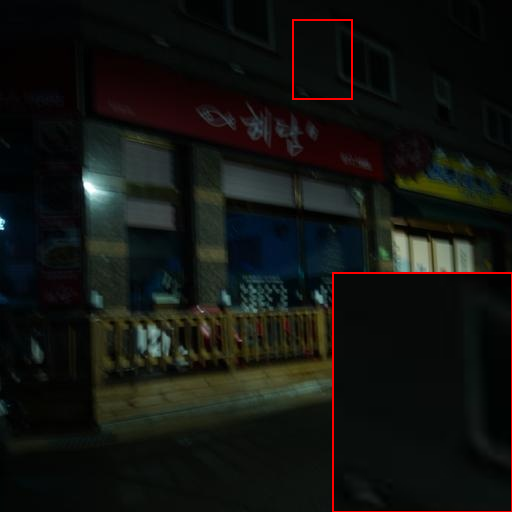} &
        \includegraphics[height=1.37cm, width=1.88cm]{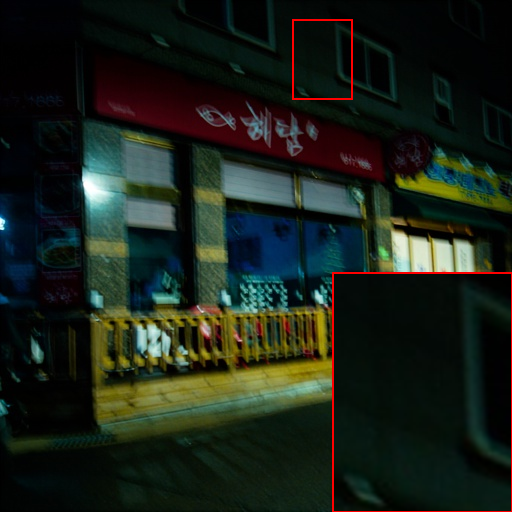} &
        \includegraphics[height=1.37cm, width=1.88cm]{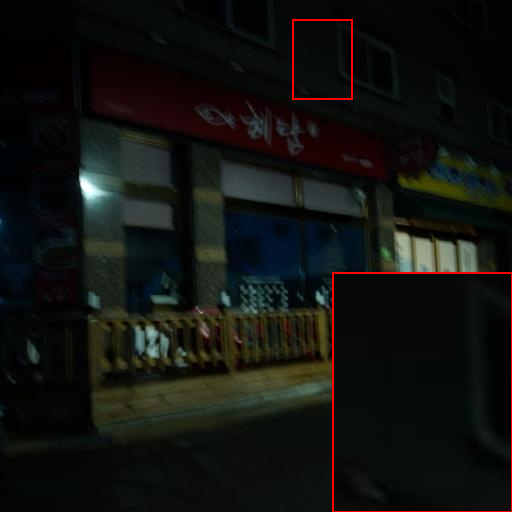} &
        \includegraphics[height=1.37cm, width=1.88cm]{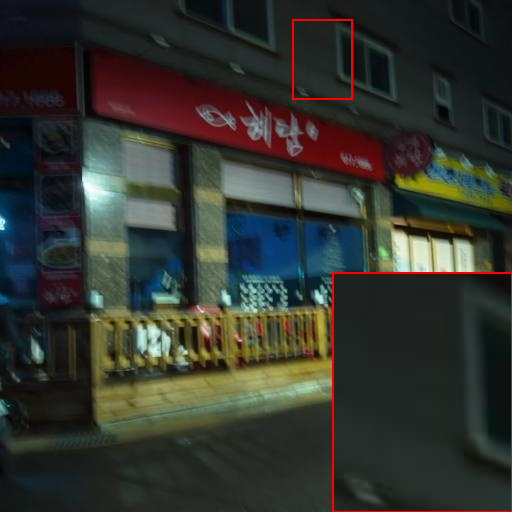} &
        \includegraphics[height=1.37cm, width=1.88cm]{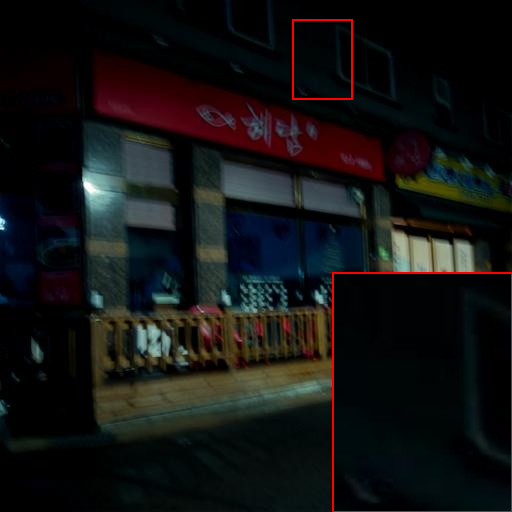} &
        \includegraphics[height=1.37cm, width=1.88cm]{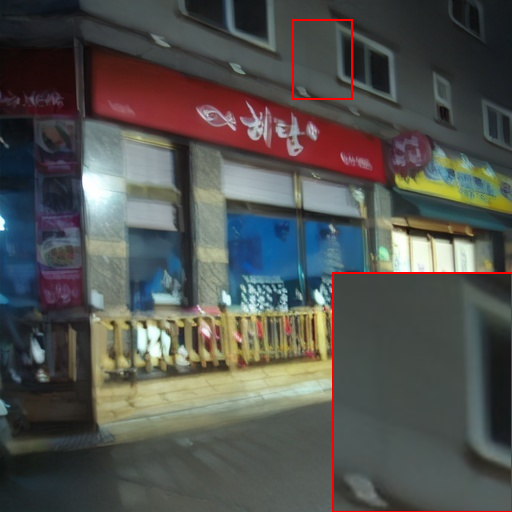} \\


    \end{tabular}
    \vspace{-5pt}
    \caption{Qualitative comparisons with non-generative methods on samples from the real mixed-degradation dataset LOLBlur.}
    \label{supfig:lolblur}
    \vspace{-8pt}
\end{figure*}

\section{More details about AutoDIR comparison}
\label{supsec: autodir_comp}

\begin{figure}
    \centering
    \small
    \setlength{\tabcolsep}{1pt}
    \begin{tabular}{cccc}
        Input&$I_{\text{sd}}$ (from LDM)&$I_{\text{sd}}=0$&GT\\

         \includegraphics[height=2cm, width=2.5cm]{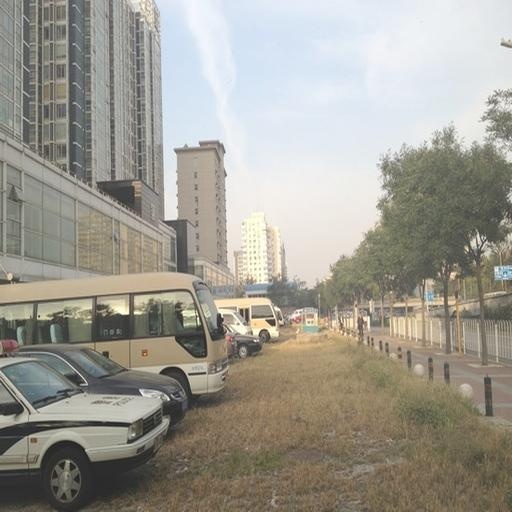}&\includegraphics[height=2cm, width=2.5cm]{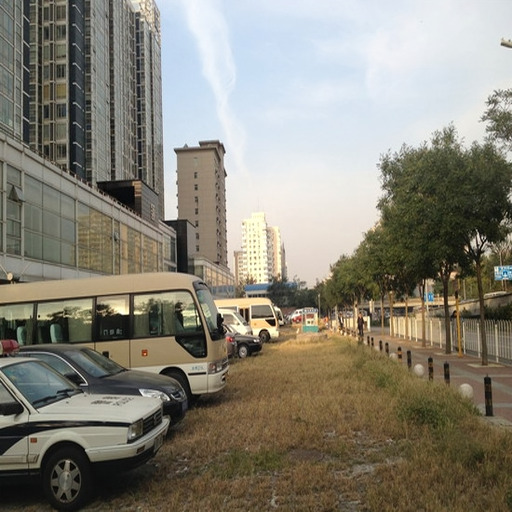}&\includegraphics[height=2cm, width=2.5cm]{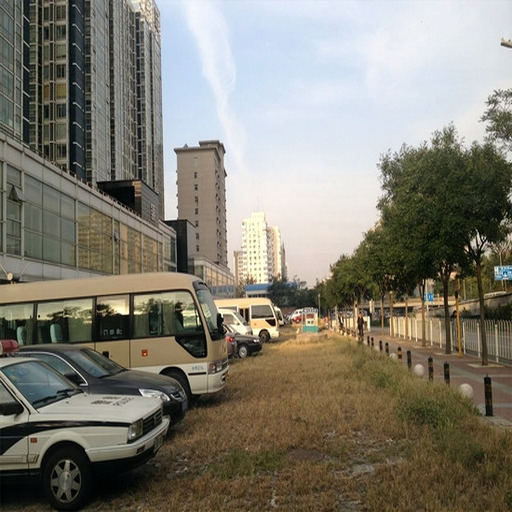}&\includegraphics[height=2cm, width=2.5cm]{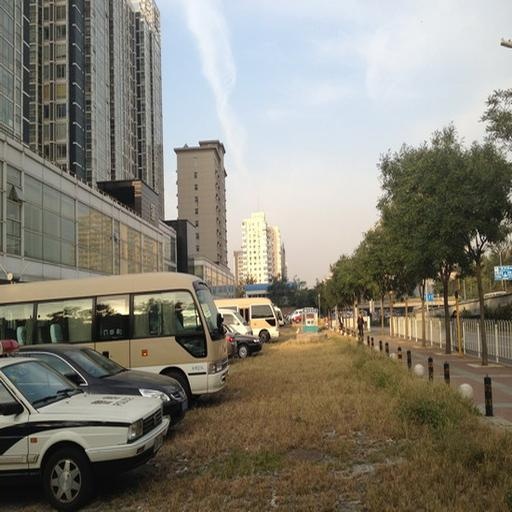}  \\

         \includegraphics[height=2cm, width=2.5cm]{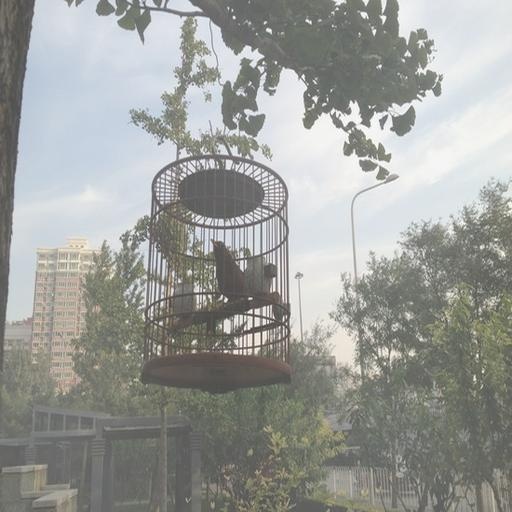}&\includegraphics[height=2cm, width=2.5cm]{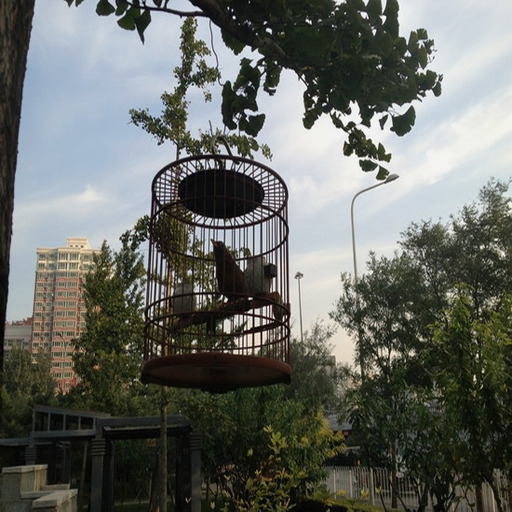}&\includegraphics[height=2cm, width=2.5cm]{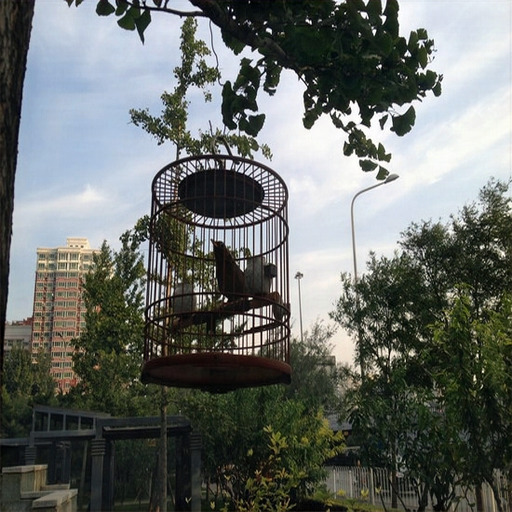}&\includegraphics[height=2cm, width=2.5cm]{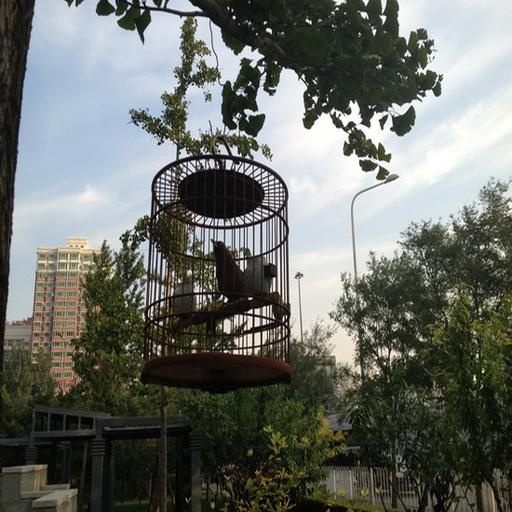}  \\

        \includegraphics[height=2cm, width=2.5cm]{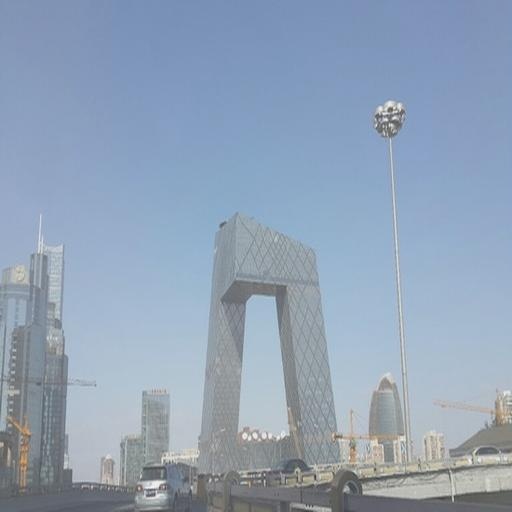}&\includegraphics[height=2cm, width=2.5cm]{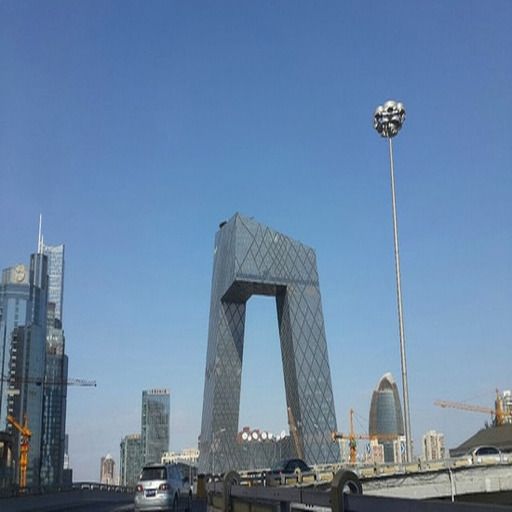}&\includegraphics[height=2cm, width=2.5cm]{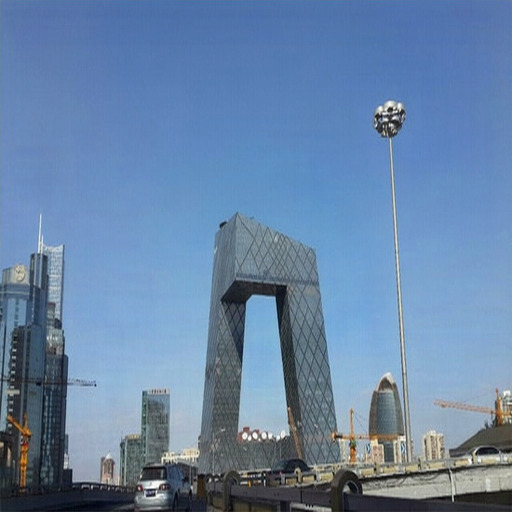}&\includegraphics[height=2cm, width=2.5cm]{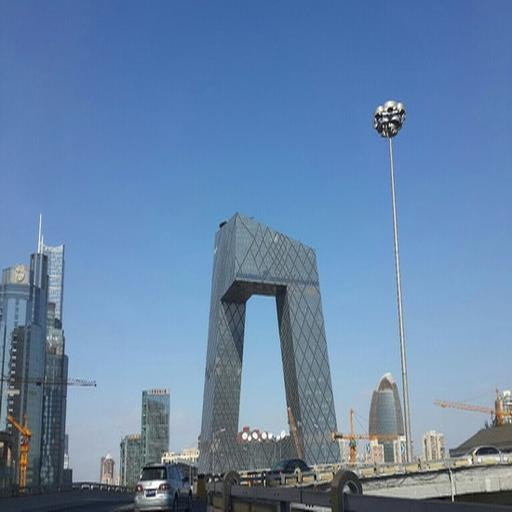}  \\
         
    \end{tabular}
    \caption{Illustration of the behavior of AutoDIR’s~\citep{autodir} structure correction module (SCM). The second column shows outputs when the SCM is applied to the LDM output $I_{\text{sd}}$, while the third column shows results when $I_{\text{sd}}$ is set to zero. Despite no structural information (third column), the SCM still restores the image, indicating that it functions as a separate non-generative restoration model.}
    \label{supfig:autodircomp}
\end{figure}

As mentioned in the main paper, comparisons with AutoDIR~\citep{autodir} were conducted without its structure correction module (SCM). AutoDIR consists of a latent diffusion model (LDM) for initial restoration, followed by an SCM which is a non-generative post-processing network. The intuition behind this approach is that the SCM predicts a residual based on the degraded input image and the restored output of the LDM, to correct the VAE-induced distortions. In short, 
\[
I_{\text{result}} = I_{\text{sd}} + \mathcal{F} \left( \left[ I_{\text{sd}}, I_{\text{deg}} \right] \right),
\]
where $I_{\text{sd}}$ is the restored output from the LDM, $I_{\text{deg}}$ is the original degraded input image, and $\mathcal{F}(\cdot)$ denotes the SCM which operates on the concatenated inputs $[I_{\text{sd}}, I_{\text{deg}}]$. However, we found that instead of slightly modulating the structural details in $I_{\text{sd}}$, the SCM behaves like a separate non-generative restoration model which directly restores $I_{\text{deg}}$. To show this, we evaluated AutoDIR with the SCM on the RESIDE~\citep{reside} dataset for two cases: (1) using $I_{\text{sd}}$ as the actual LDM output and (2) setting $I_{\text{sd}}=0$, effectively removing any structural information from the LDM. If the SCM were functioning as a corrective module, performance in the second case should deteriorate significantly. However, we found that the SCM was able to independently restore the degraded input in the second case, as shown in Fig.~\ref{supfig:autodircomp}. This suggests that the SCM largely ignores the LDM output and instead performs direct restoration on $I_{\text{deg}}$, thereby behaving as a non-generative restoration network. Therefore, to ensure a fair comparison with other generative models, we evaluated only the LDM output of AutoDIR.


\section{Limitations and Scope for Future Work}
\label{supsec: limitations}

Despite the strengths of RestoreVAR, there remains scope for improvement. First, its performance is inherently constrained by the latent refiner transformer (LRT) and the VAE decoder. While the LRT significantly improves results over using no refiner, it does not reach the upper bound set by directly decoding from ground-truth continuous latents. Exploring improved VQVAE and refiner architectures could help address this. Another promising direction is to employ our non-generative LRT in fully generative VAR models, given its strong performance for AiOR. Finally, future work can investigate how the performance of RestoreVAR scales with larger VAR backbones.
\begin{wrapfigure}{r}{0.5\textwidth}
    \centering
    \small
    \setlength{\tabcolsep}{1pt}
    \begin{tabular}{cccc}
    &Input&RestoreVAR&GT\\
          \rotatebox[origin=c]{90}{SID\hspace{-31pt}}&\includegraphics[height=1.37cm, width=2.1cm]{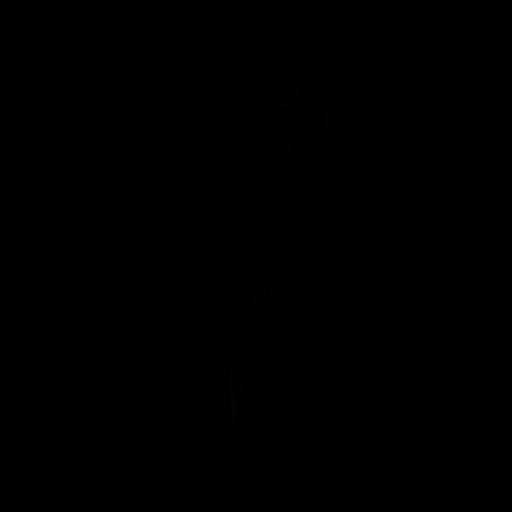}
          &\includegraphics[height=1.37cm, width=2.1cm]{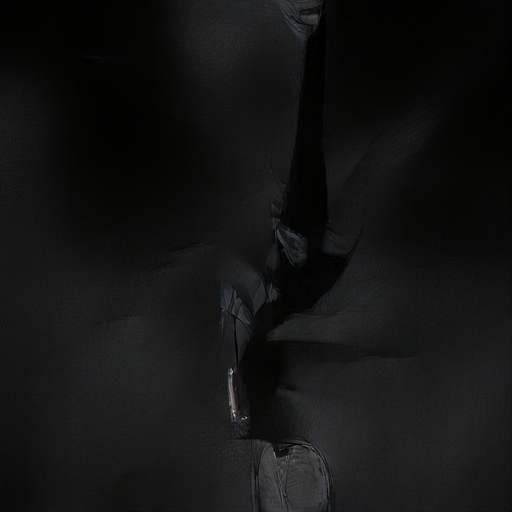}&
          \includegraphics[height=1.37cm, width=2.1cm]{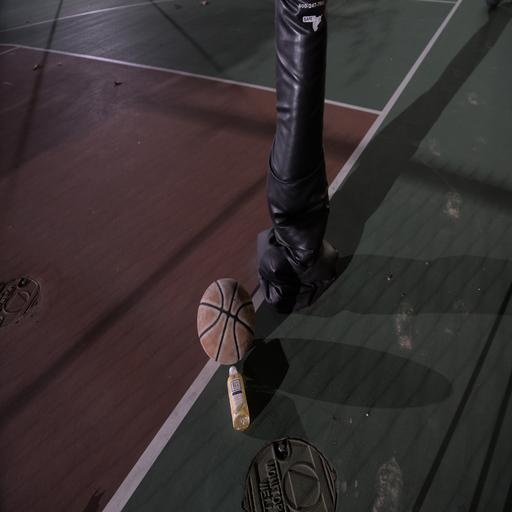}
    \end{tabular}
    \vspace{-5pt}
    \caption{RestoreVAR does not perform well when the input degradation is very severe.}
    \label{supfig: failure}
    \vspace{-12pt}
\end{wrapfigure}
Additionally, we found that RestoreVAR does not perform well when the input degradation is extremely severe. Fig.~\ref{supfig: failure} shows the result for a very dark low-light sample from the SID~\citep{sid} dataset. The restored output looks inferior to the GT due to the very severe nature of the degradation. 



\section{LLM Usage}
\label{supsec: llm_usage}

LLM was used only for polishing writing in parts of the main paper and supplementary. 

\end{document}